\documentclass[10pt,twocolumn,letterpaper]{article}

\usepackage{cvpr}              
\usepackage[accsupp]{axessibility} 
\usepackage{graphicx}
\usepackage{amsmath}
\usepackage{amssymb}
\usepackage{booktabs}

\usepackage[pagebackref,breaklinks,colorlinks]{hyperref}

\usepackage[capitalize]{cleveref}
\crefname{section}{Sec.}{Secs.}
\Crefname{section}{Section}{Sections}
\Crefname{table}{Table}{Tables}
\crefname{table}{Tab.}{Tabs.}

\def\cvprPaperID{1557} 
\def\confName{CVPR}
\def\confYear{2022}

\begin{document}

\title{Shape from Polarization for Complex Scenes in the Wild}

\author{Chenyang Lei\thanks{Joint first authors} $^{1}$ \quad Chenyang Qi\footnotemark[1] $^1$ \quad  Jiaxin Xie\footnotemark[1] $^1$ \quad  Na Fan$^1$ \quad  Vladlen Koltun$^2$ \quad  Qifeng Chen$^1$\\
$^1$HKUST  \qquad $^2$Apple\\
}

\maketitle

\begin{abstract}
We present a new data-driven approach with physics-based priors to scene-level normal estimation from a single polarization image. Existing shape from polarization (SfP) works mainly focus on estimating the normal of a single object rather than complex scenes in the wild. A key barrier to high-quality scene-level SfP is the lack of real-world SfP data in complex scenes. Hence, we contribute the first real-world scene-level SfP dataset with paired input polarization images and ground-truth normal maps. Then we propose a learning-based framework with a multi-head self-attention module and viewing encoding, which is designed to handle increasing polarization ambiguities caused by complex materials and non-orthographic projection in scene-level SfP. Our trained model can be generalized to far-field outdoor scenes as the relationship between polarized light and surface normals is not affected by distance. Experimental results demonstrate that our approach significantly outperforms existing SfP models on two datasets. Our dataset and source code will be publicly available at \url{https://github.com/ChenyangLEI/sfp-wild}.

\end{abstract}

\begin{figure}[h!]
\centering

\begin{tabular}{@{}c@{\hspace{0.5mm}}c@{\hspace{0.5mm}}c@{\hspace{0.5mm}}c@{\hspace{0.5mm}}c@{}}

\includegraphics[width=0.245\linewidth]{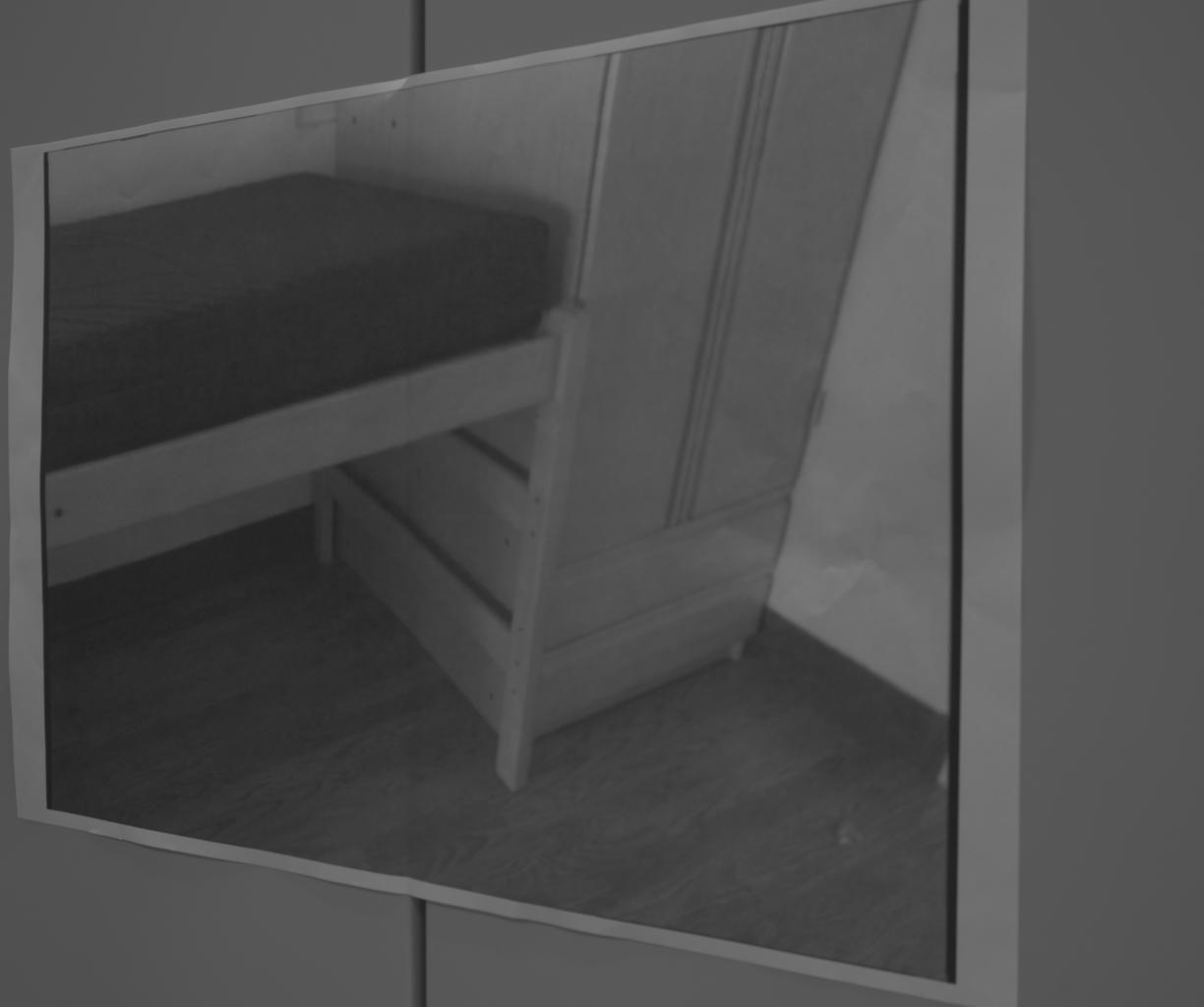}&
\includegraphics[width=0.245\linewidth]{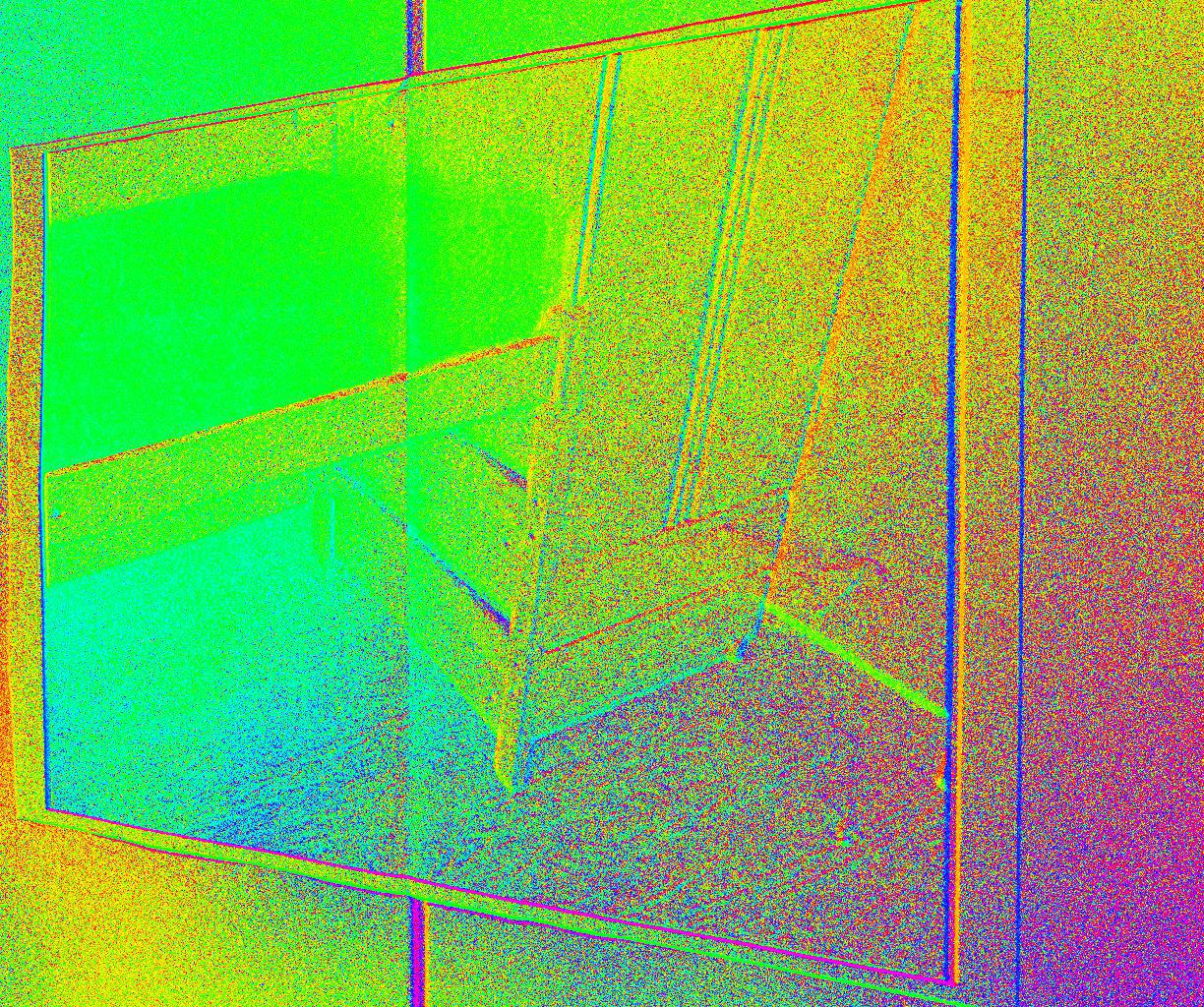}&
\includegraphics[width=0.245\linewidth]{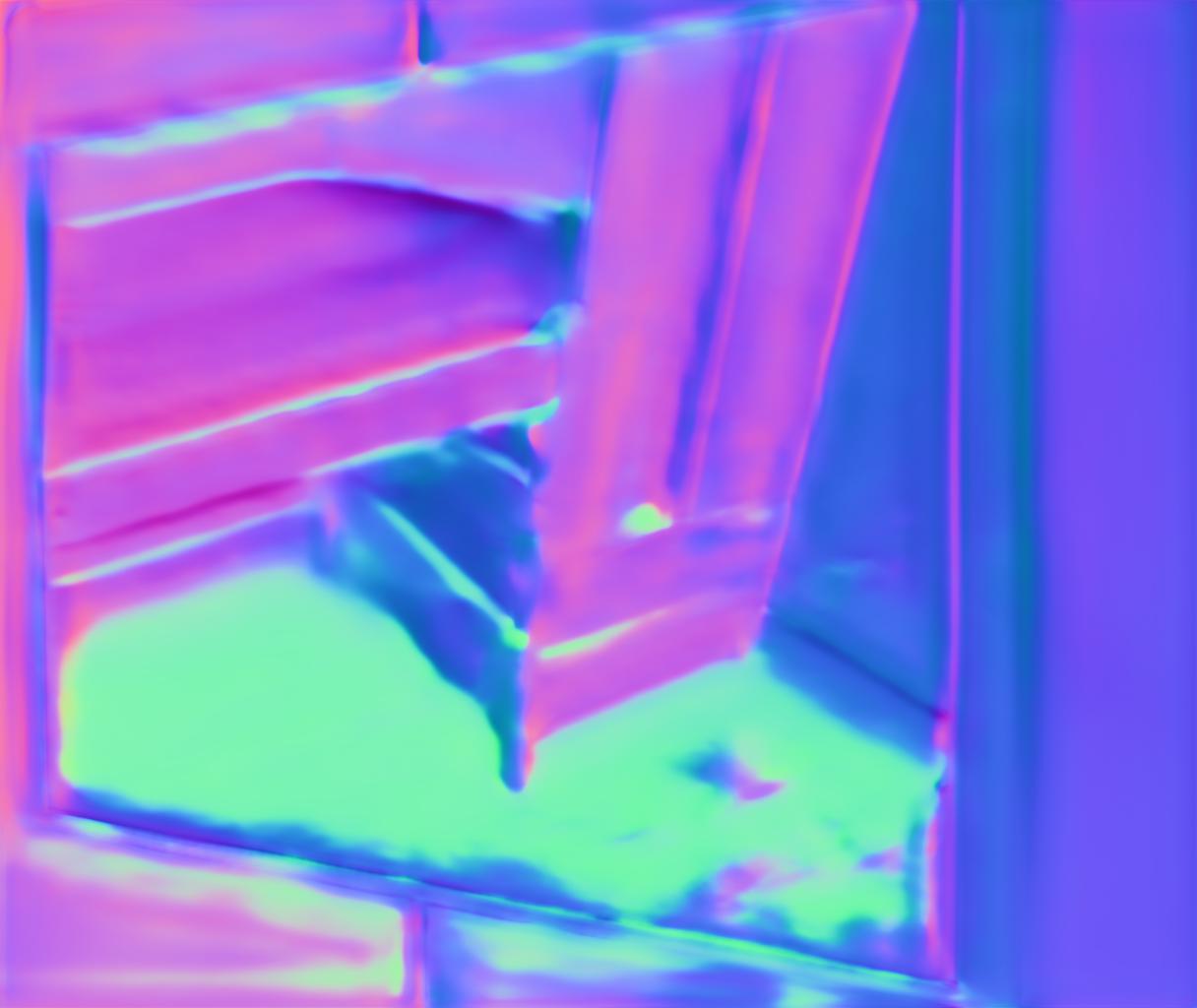}&
\includegraphics[width=0.245\linewidth]{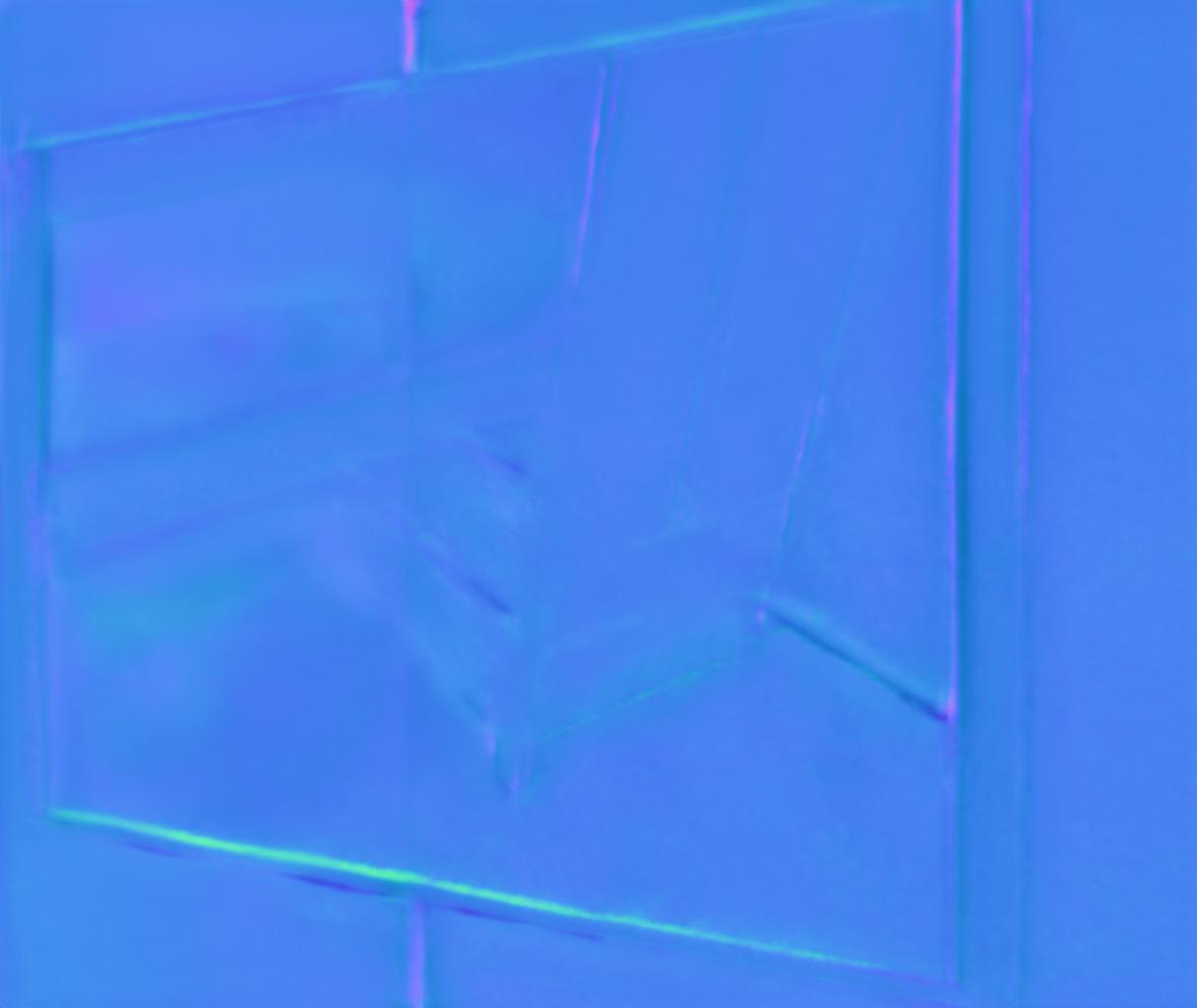}\\

\includegraphics[width=0.245\linewidth]{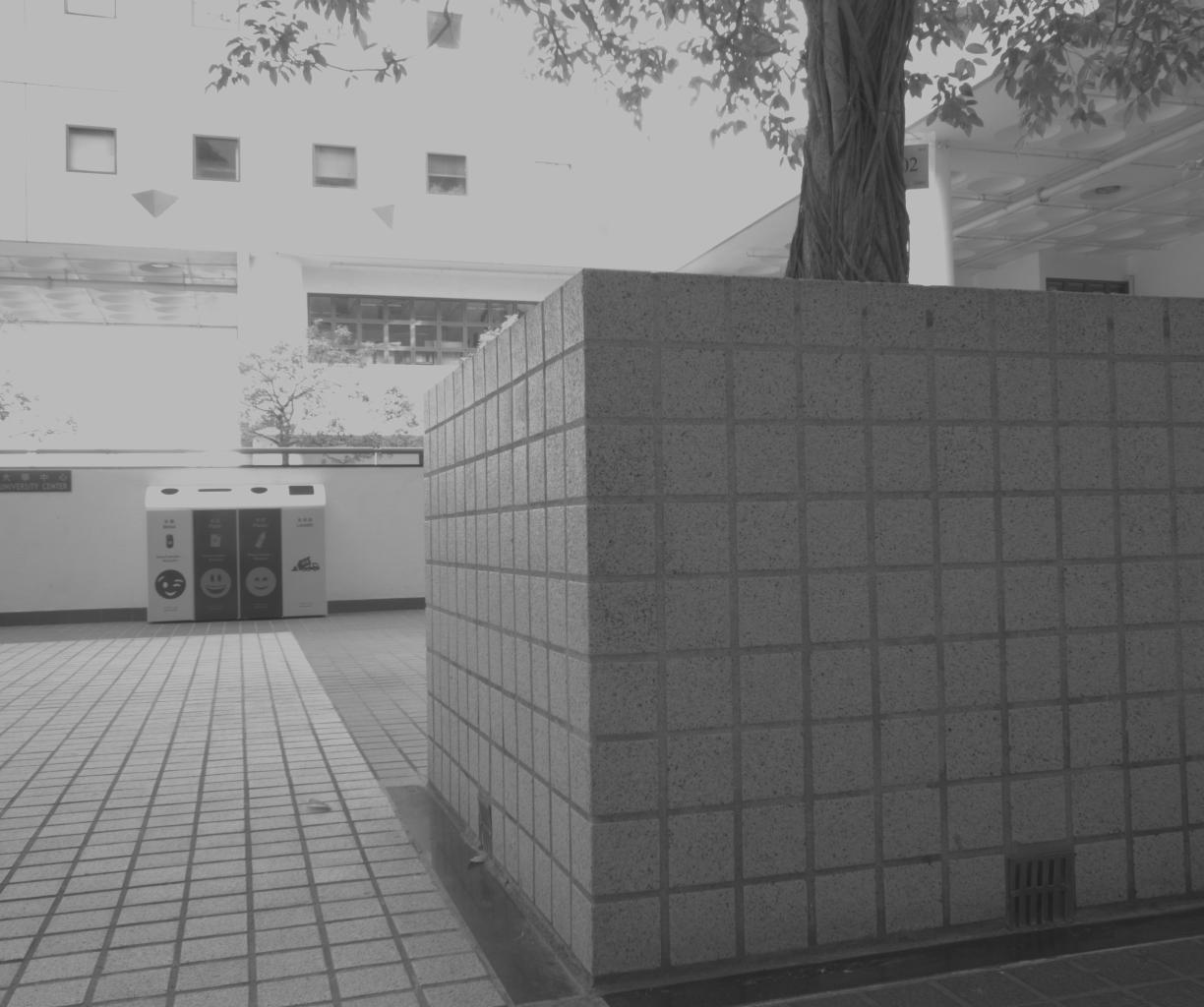}&
\includegraphics[width=0.245\linewidth]{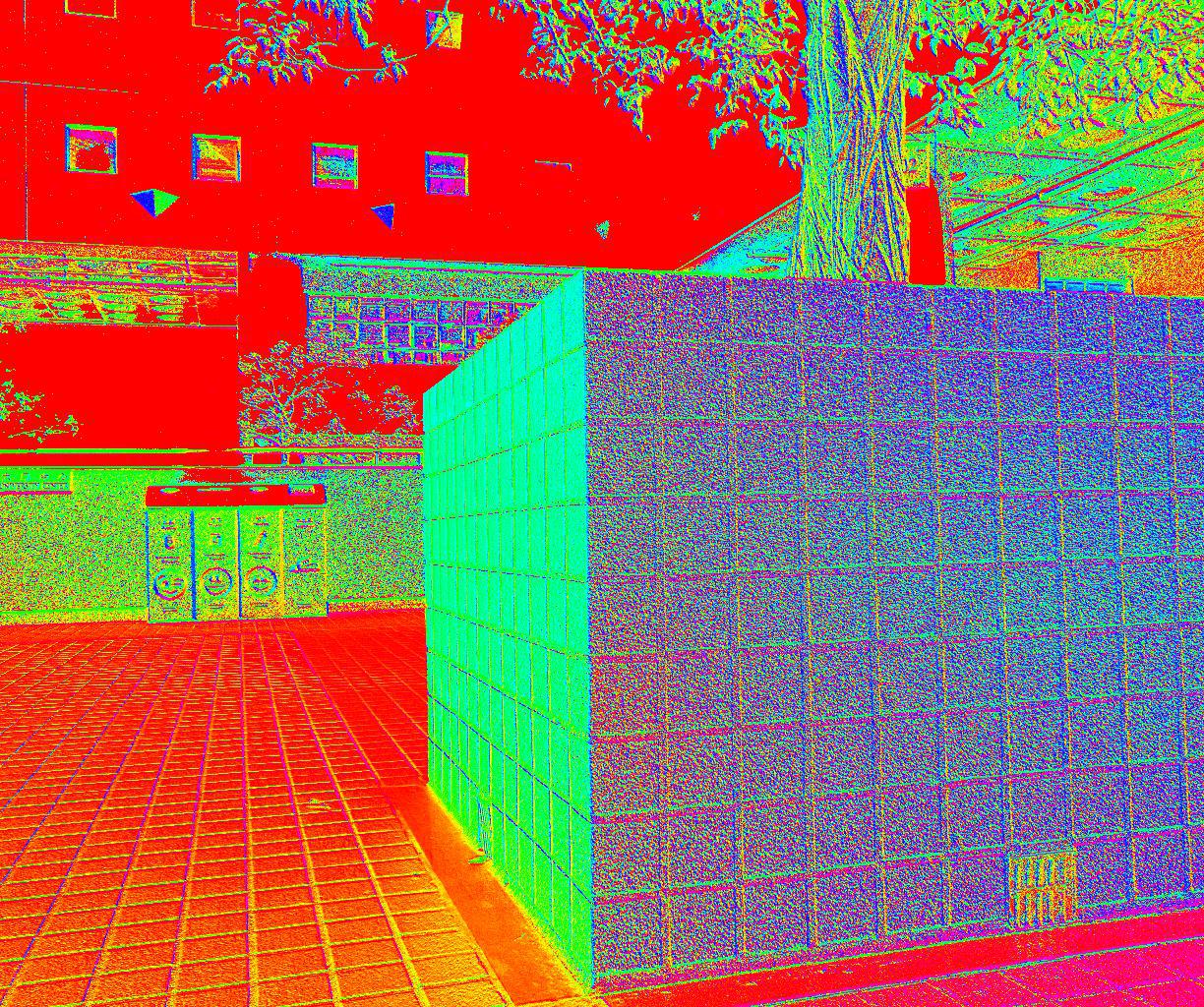}&
\includegraphics[width=0.245\linewidth]{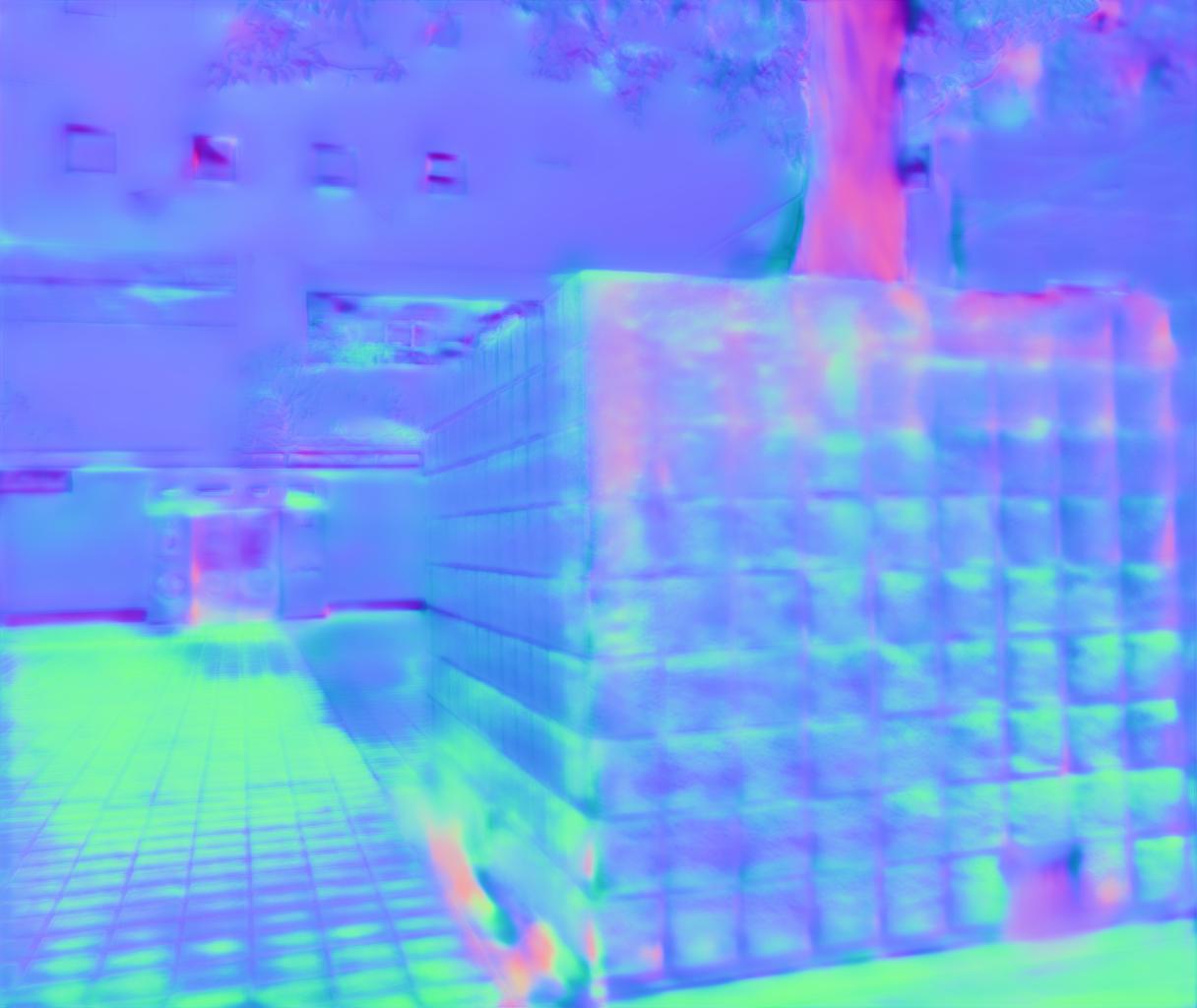}&
\includegraphics[width=0.245\linewidth]{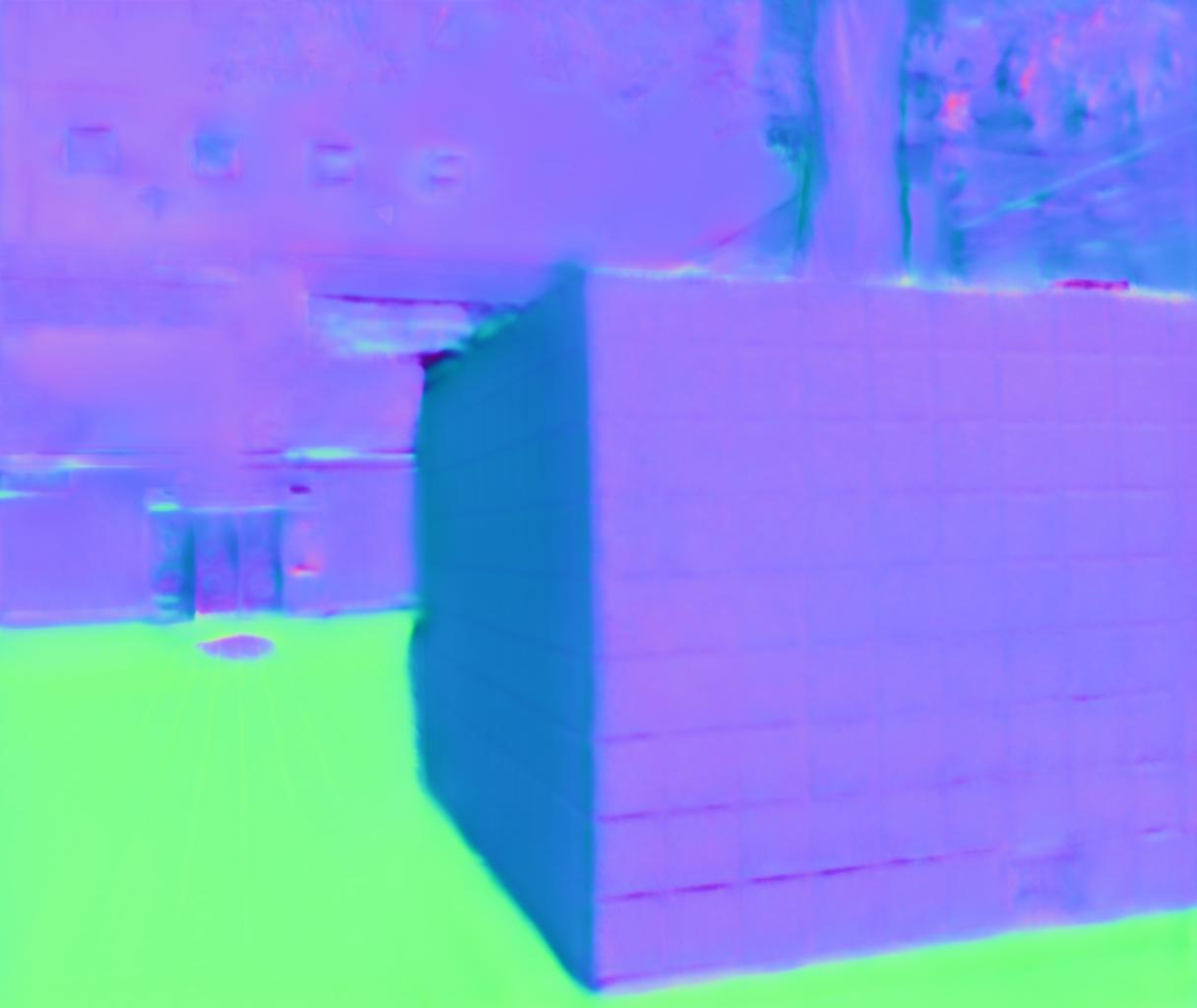}\\

\includegraphics[width=0.245\linewidth]{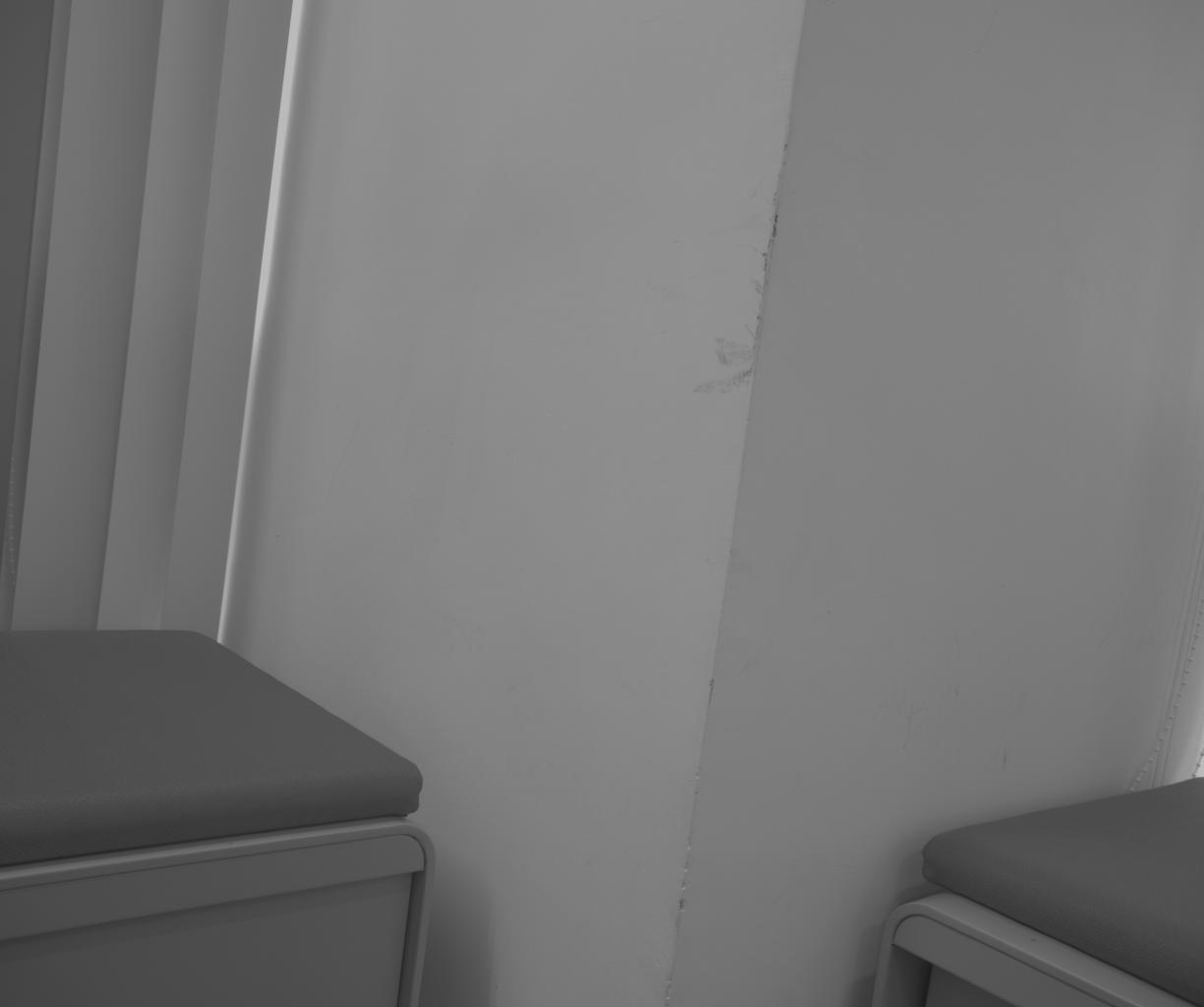}&
\includegraphics[width=0.245\linewidth]{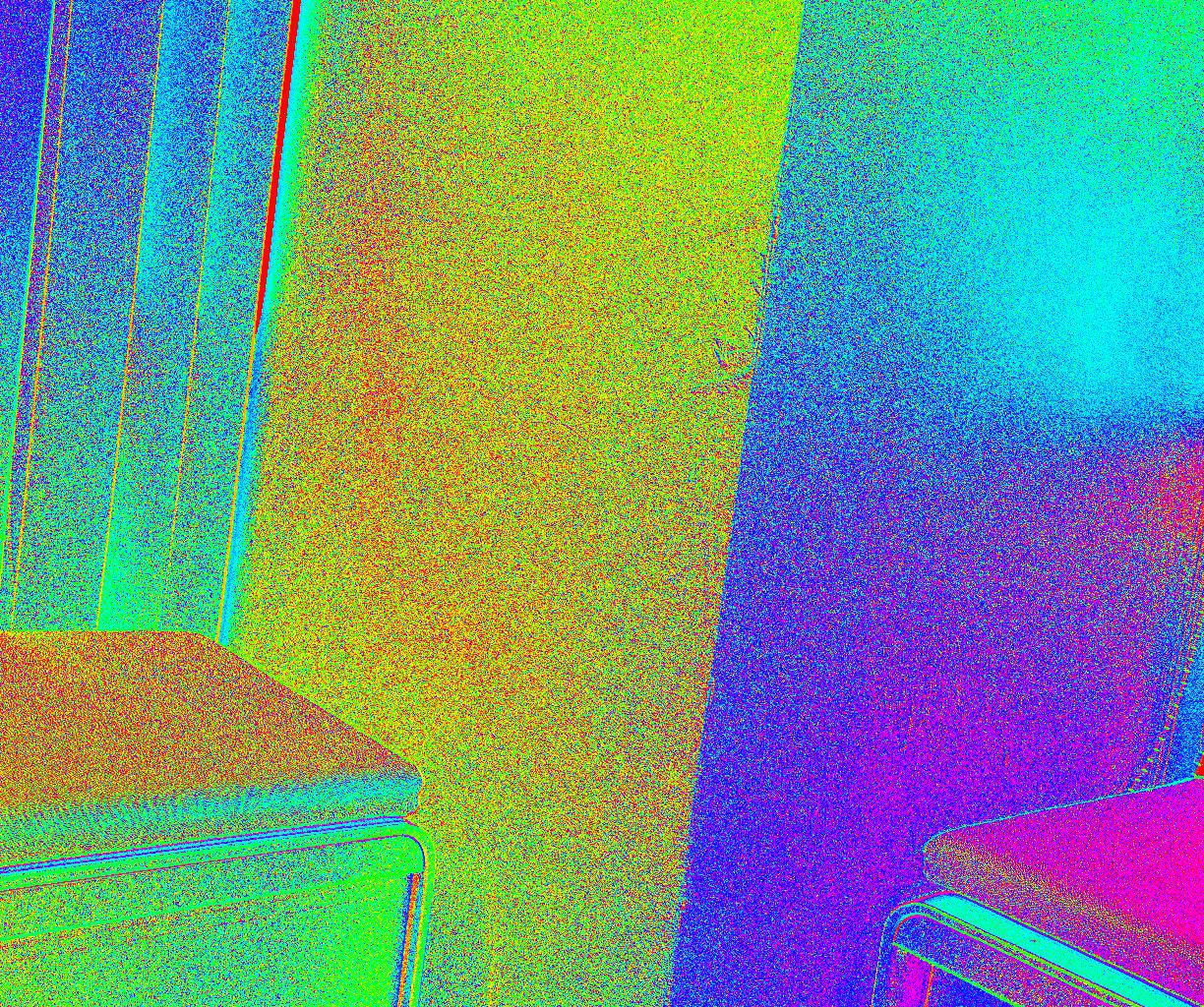}&
\includegraphics[width=0.245\linewidth]{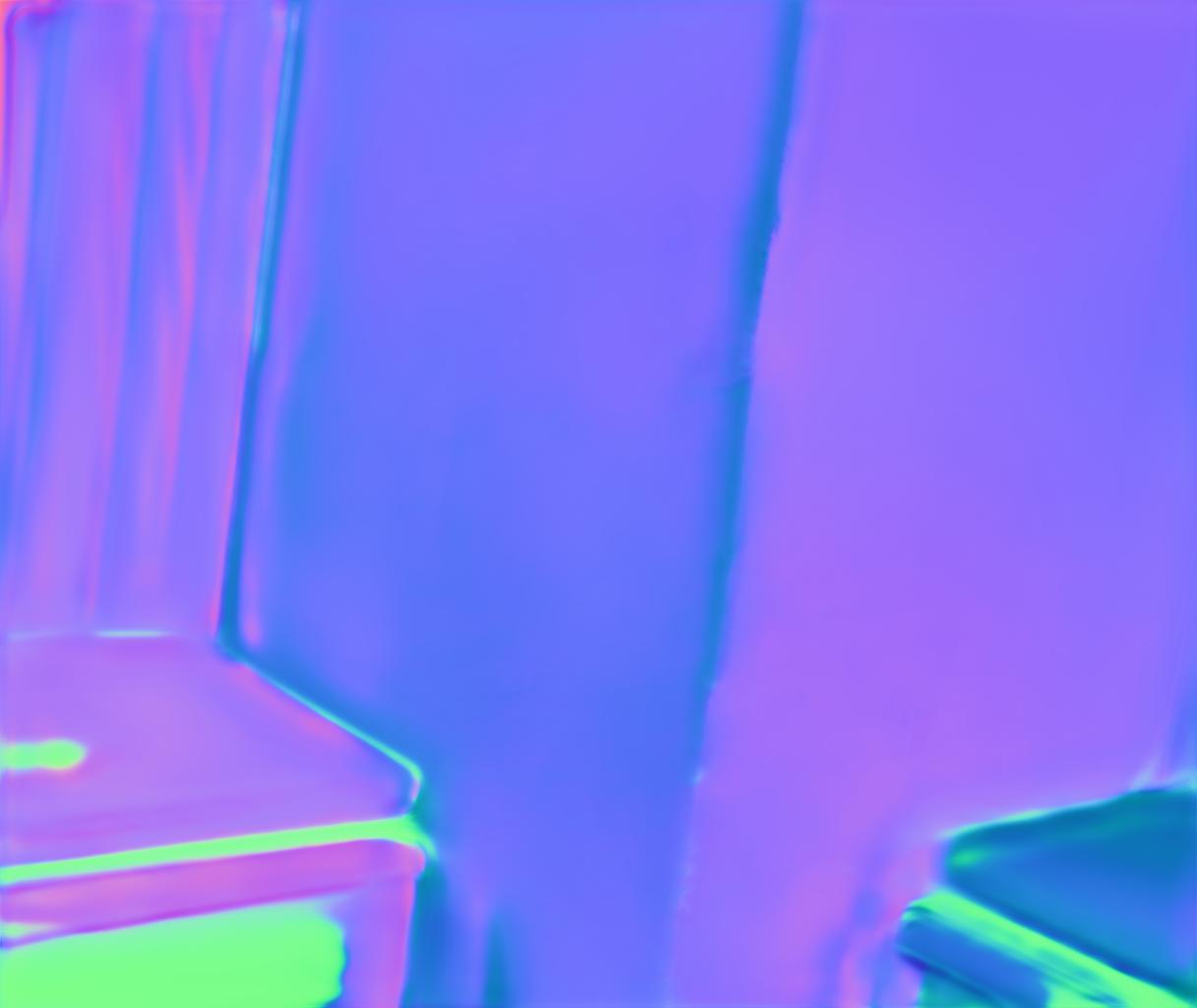}&
\includegraphics[width=0.245\linewidth]{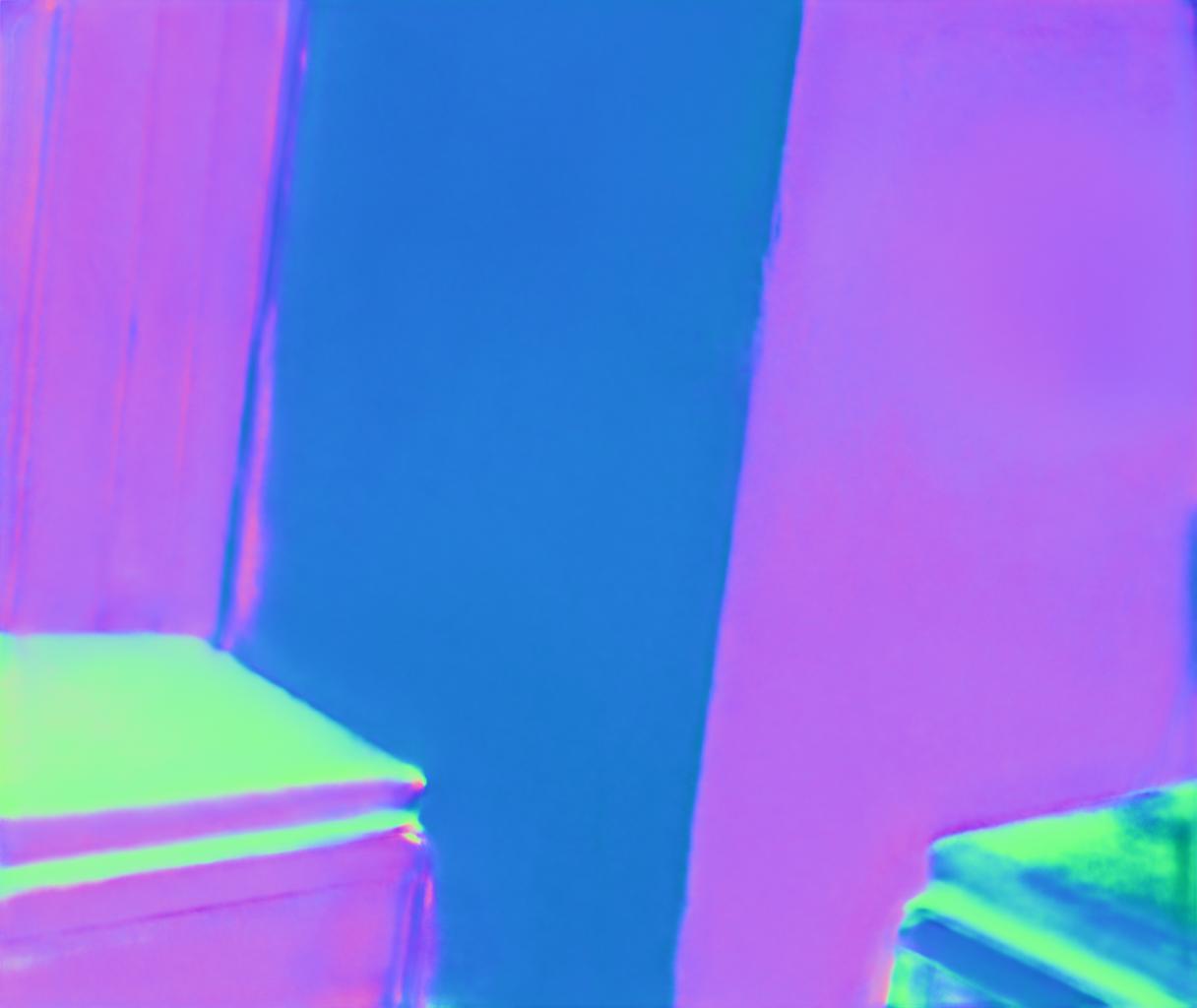}\\
\small{Input $I_{un}$}& \small{Input $\phi$} & \small{Without pol.} & \small{With pol.} \\

\end{tabular}

\caption{Our method can estimate dense scene-level surface normals from a single polarization image. Polarization can provide effective cues for obtaining more accurate results. In the first row, polarization provides geometry cues for our model so that it is not fooled by objects in the printed image on a wall. In the second and third rows, polarization provides guidance for planes with different surface normals even when their materials are quite similar. $I_{un}$: unpolarized image; $\phi$: angle of polarization. 
}
\label{fig:teaser}
\end{figure}

\section{Introduction}
Accurate surface normal estimation in the wild can provide valuable information about a scene's geometry and can be used in various computer vision tasks, including segmentation~\cite{fan2020sne}, 3D reconstruction~\cite{kazhdan2006poisson}, and many others~\cite{li2020inverse,Huang_2019_ICCV_framenet}. Therefore, normal estimation is an important task studied for a long time. However, estimating high-quality normals in the wild is still an open problem. Various techniques such as photometric stereo~\cite{DBLP:chen2020PS, chen2019selfps} can produce high-frequency normals, but most of them only provide short-range object-level normal maps. Active depth sensors can be another approach to obtaining normals from depth maps, but the corresponding depth maps are often sparse (LiDAR) or noisy (time-of-flight, structured light) so they can not estimate normals reliably. Also, the depth range of active sensors is limited.




In this work, we are interested in estimating surface normal from a single polarization image for complex scenes in the wild. Since the polarization of light changes differently when the light interacts with the surfaces of different shapes and materials (governed by the Fresnel equations~\cite{collett2005field}), the polarization images can provide dense surface orientation cues from the polarized light perceived at each pixel. Also, compared with the active sensors and object-level normal estimation techniques (e.g., photometric stereo), the polarization camera is a passive sensor that is not constrained to a specific depth range. Thus polarization images are promising data sources for accurate normal estimation in the wild.

However, estimating normals from a polarization image for complex scenes (scene-level SfP) is challenging. To the best of our knowledge, no existing SfP work focuses on complex scenes, and several challenges are yet to be solved. Firstly, polarization contains ambiguities from unknown information such as object materials and reflection types~\cite{collett2005field}. Object-level SfP methods approach these ambiguities by utilizing various cues (e.g., shading~\cite{smith2019height}) or making restrictive assumptions (e.g., known albedo~\cite{mahmoud2012direct}), which are unfeasible for multiple-object scenes because of the variabilities of material properties and complexities of reflections. Secondly, while some works~\cite{ba2020deep,kondo2020accurate} demonstrate the potential of combining convolutional neural networks and polarization cues in estimating normals for unknown materials, there are only object-level~\cite{ba2020deep} or synthetic data~\cite{kondo2020accurate} for training, which are not sufficient for scene-level SfP. Finally, scene-level SfP brings up another challenge. The viewing direction influences the measured polarization information. Previous object-level SfP approaches ignore the impact of viewing direction since they assume orthographic projection by placing objects at the center of an image, which does not hold for scene-level SfP.

To solve the challenge of lacking real-world scene-level polarization data, we construct the first real-world scene-level SfP dataset that contains diverse complex scenes. Building such a new dataset is necessary because the existing DeepSfP dataset~\cite{ba2020deep} only contains a single object per image and the dataset by Kondo et al.~\cite{kondo2020accurate} is synthetic and not publicly available.  


Due to the challenges of scene-level SfP, the performances of previous learning-based SfP works~\cite{ba2020deep,kondo2020accurate} are not satisfactory when they are trained on our scene-level data. To improve the performance of scene-level SfP in the wild, we adopt three novel designs in our model. First, we introduce multi-head self-attention~\cite{vaswani2017attention} in a convolutional neural network (CNN) for SfP. Multi-head self-attention utilizes the global context of an image, which helps the CNN resolve the local ambiguities in polarization cues. Second, to handle non-orthographic projection for scene-level SfP, the neural network must be aware of the viewing direction of each pixel since the convolution operation is translation invariant. We thus propose a simple but critical technique that improves the performance of SfP methods on scene-level data: providing per-pixel viewing encoding to the neural network. Finally, as an additional contribution, we design a novel polarization representation, which is effective and considerably more efficient than the representations in prior work~\cite{ba2020deep}.

We compare our approach with various state-of-the-art methods. Experimental results show that our model can generate a high-quality normal map from a single polarization image (Fig.~\ref{fig:teaser}) and can generalize beyond the depth range of the training data. In summary, our contributions are as follows.

\begin{figure*}[t]
\centering

\begin{tabular}{@{}c@{\hspace{1mm}}c@{\hspace{1mm}}c@{\hspace{1mm}}c@{\hspace{1mm}}c@{\hspace{1mm}}c@{\hspace{1mm}}c@{}}
\includegraphics[width=0.160\linewidth]{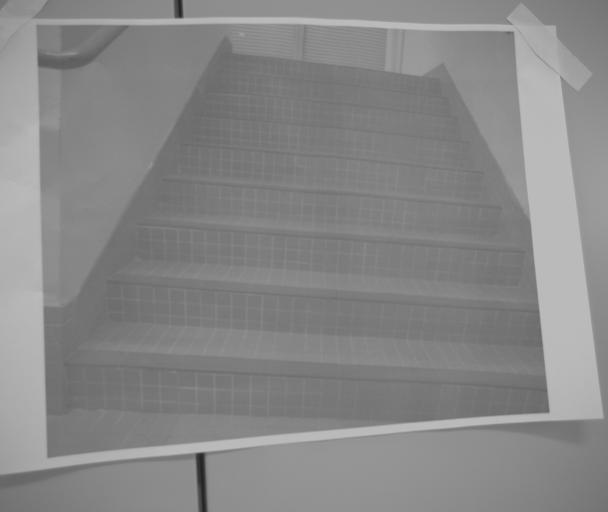}&
\includegraphics[width=0.160\linewidth]{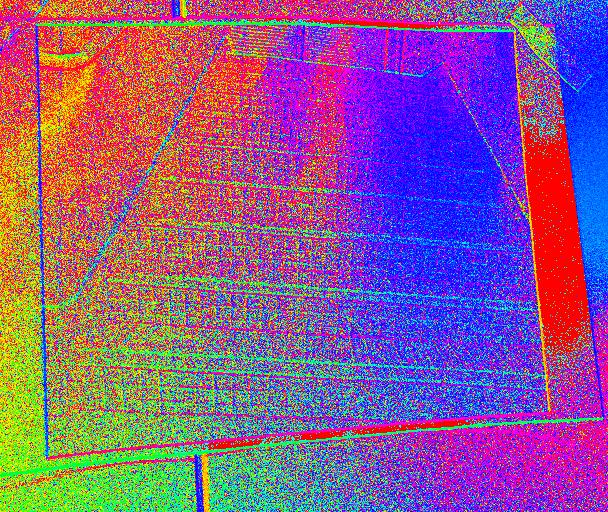}&
\includegraphics[width=0.160\linewidth]{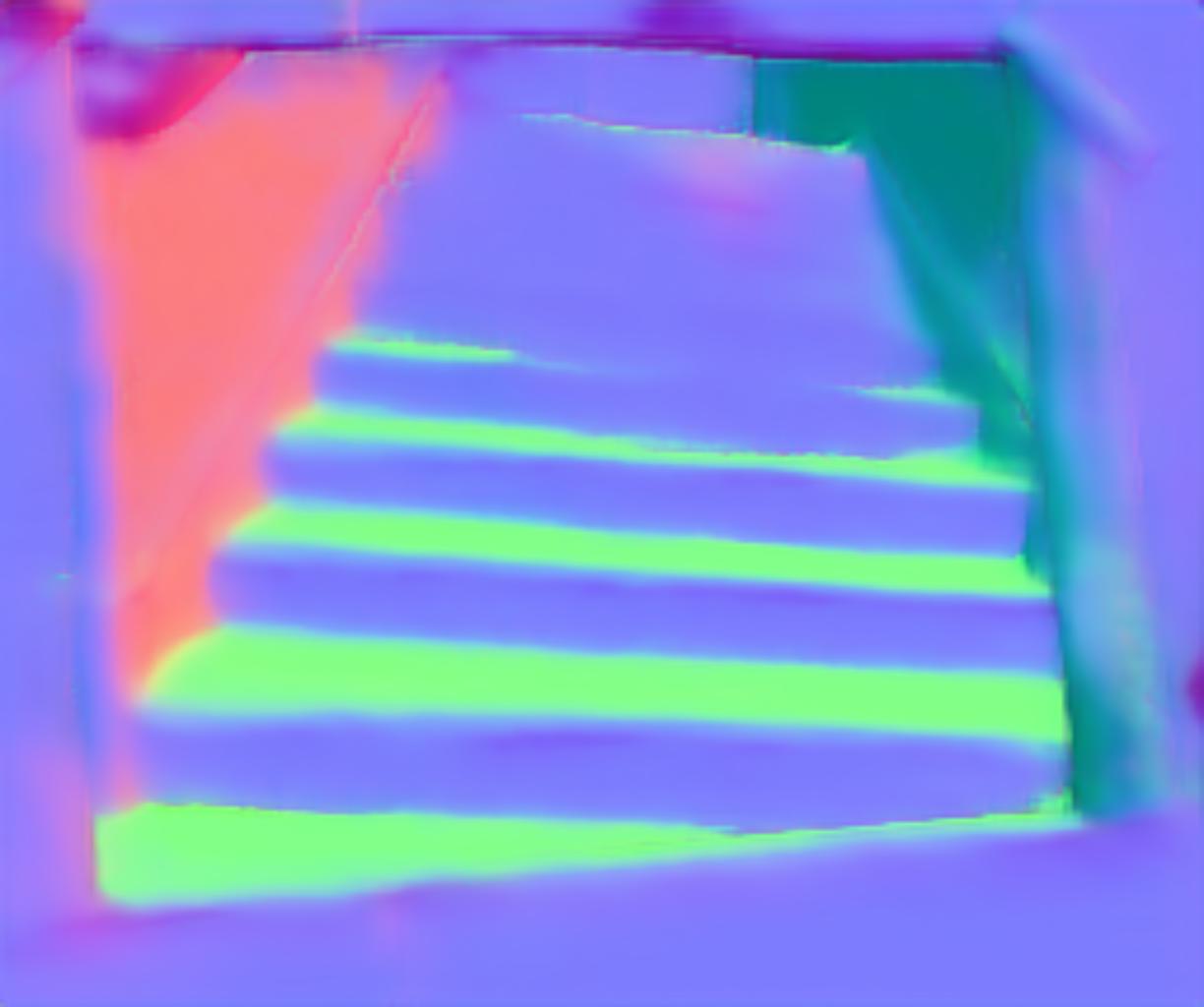}&
\includegraphics[width=0.160\linewidth]{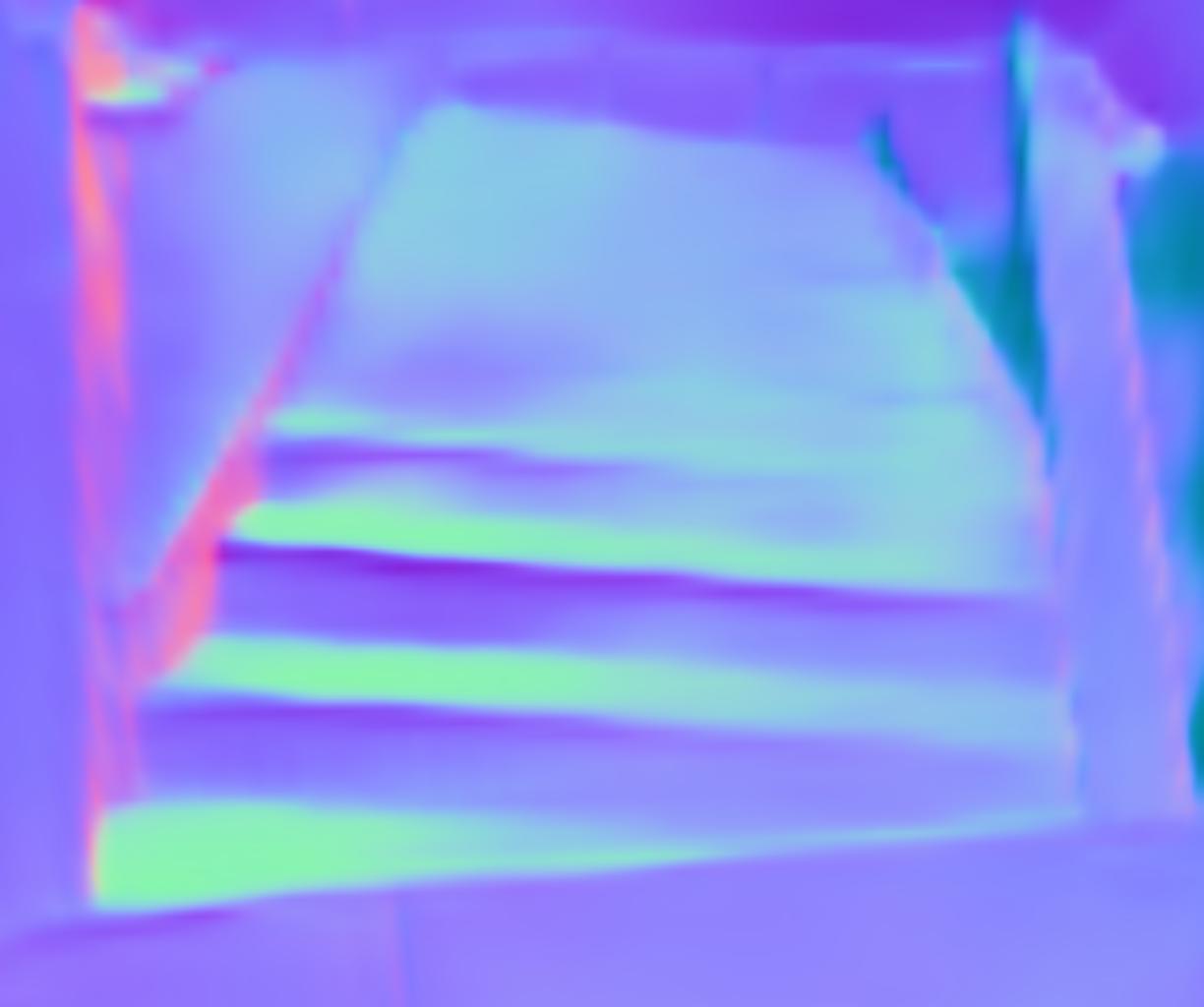}&
\includegraphics[width=0.160\linewidth]{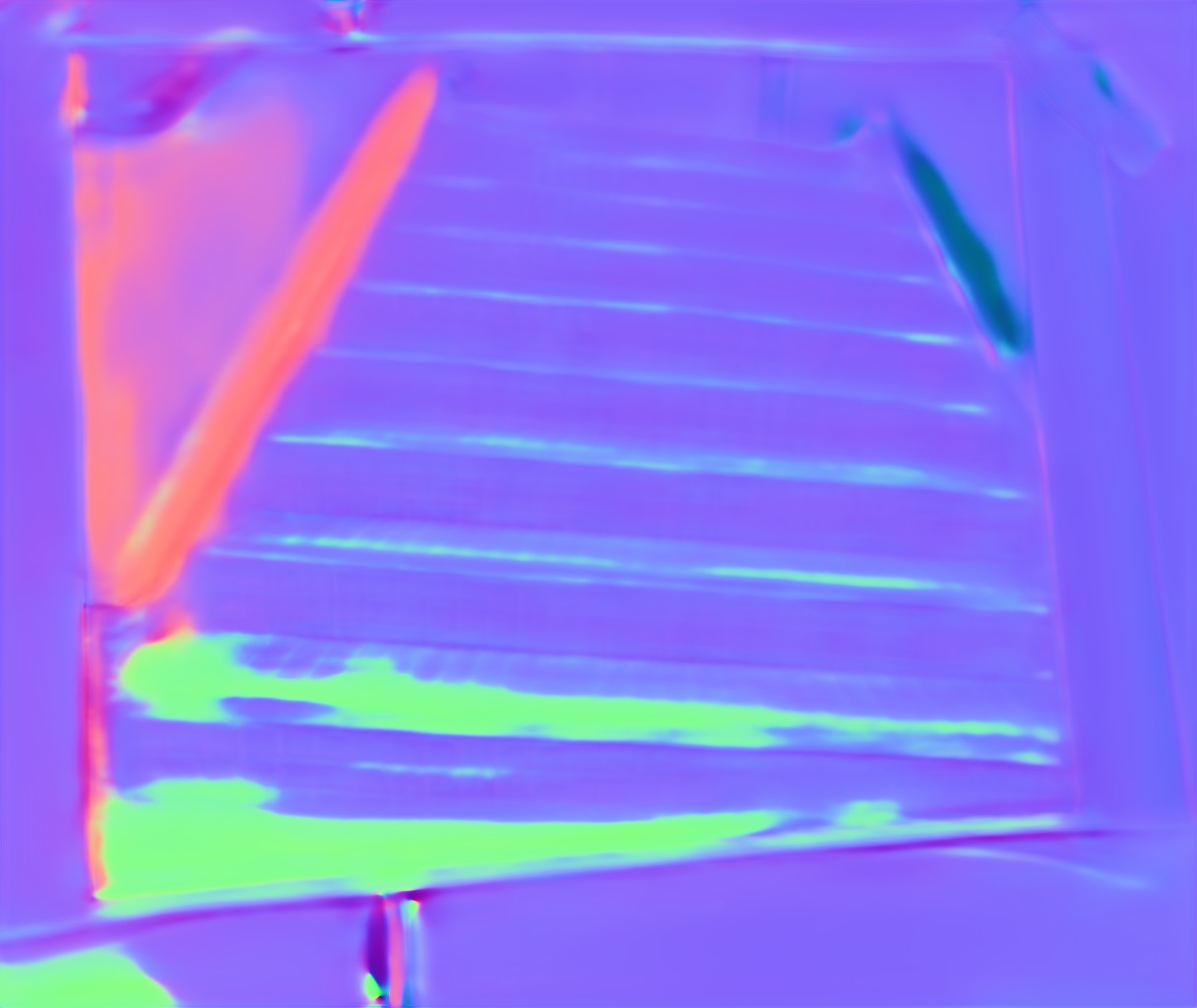}&
\includegraphics[width=0.160\linewidth]{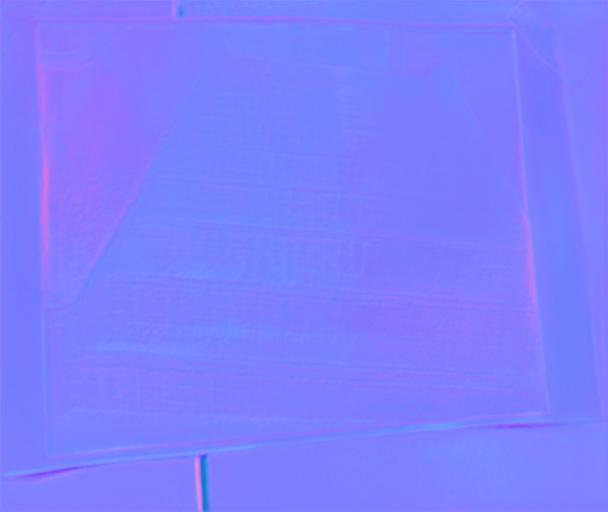}
\\

\small{Polarization image}& \small{Polarization angle} & XTC~\cite{Zamir2020Robust} & Do et al.~\cite{Do2020SurfaceNormal}& \small{Ours w/o polarization} & \small{Ours w/ polarization} \\ 

\end{tabular}

\caption{\textbf{The importance of polarization.} Note that polarization conveys the underlying physical shape while RGB-based methods~\cite{Do2020SurfaceNormal,Zamir2020Robust} are distracted by the semantics in a printed picture attached to a wall.}
\label{fig:percep_physics}
\end{figure*}

\begin{itemize}
    \item We construct the first real-world SfP dataset containing paired input polarization images and ground-truth normal maps in complex scenes. 
    
    \item Our proposed shape-from-polarization approach is the first one trained on complex real-world scene-level data and also the best-performing one for normal estimation from polarization in the wild.

    \item Technically, we introduce three novel designs to scene-level SfP: viewing encoding that can handle the challenge of non-orthographic projection in scene-level SfP, a dedicated network architecture that adopts multi-head self-attention for SfP, and a practical polarization representation that is effective and efficient. 
    

    
\end{itemize}

\section{Related Work}

\noindent \textbf{Shape from polarization.}
The polarization of light changes when the light interacts with a surface, which can be described by the Fresnel equations using the geometry and materials of objects~\cite{collett2005field}. Shape from polarization (SfP) works~\cite{rahmann2001reconstruction,atkinson2006recovery,miyazaki2003polarization} utilize this effect to estimate the surface normal of objects. Since the polarization state is affected by various factors simultaneously, early SfP methods usually enforce assumptions of reflection types and materials to constrain the problem. For example, Rahmann et al.~\cite{rahmann2001reconstruction} assume pure specular reflection and some works~\cite{atkinson2017polarisation_photometric,miyazaki2003polarization} assume pure diffuse reflection. 

Various cues and techniques have been explored to resolve the ambiguities in this problem. Atkinson et al.~\cite{atkinson2007shape_twoviews} use shading from two views. Baek et al.~\cite{baek2018simultaneous} perform joint optimization of appearance, normals, and refractive index. A coarse depth map from a depth sensor~\cite{kadambi2015polarized}, two-view stereo~\cite{zhu2019depth,fukao2021polarimetric}, reciprocal image pairs~\cite{ding2021polarimetric}, or multi-view stereo~\cite{cui2017polarimetric, Miyazaki2016SurfaceNE}, can also be served to disambiguate the problem. 
For single-view SfP, some methods combine photometric stereo~\cite{atkinson2017polarisation_photometric} or shading information~\cite{mahmoud2012direct,smith2019height} with SfP. Also, some works try to solve this problem under specific illumination conditions (e.g., front-flash illumination~\cite{deschaintre2021deep} and sunlight under the clear sky~\cite{ichikawa2021shape}). Unlike these works, our approach aims to estimate surface normal in the wild without specific assumptions or additional tools.

Deep learning is proven effective in solving the ambiguities of object-level SfP. Ba et al.~\cite{ba2020deep} collect a real-world object-level dataset and train a CNN to obtain normals from polarization, significantly outperforming physics-based SfP. Instead of collecting real-world data, Kondo et al.~\cite{kondo2020accurate} create a synthetic dataset of polarization images with a new polarimetric BRDF model. However, these approaches have not studied complex scenes in the wild due to the lack of real-world scene-level data. To address this issue, we propose the first real-world scene-level SfP dataset. Besides, we also notice existing frameworks~\cite{ba2020deep,kondo2020accurate} cannot achieve satisfactory results on our dataset due to the challenges that emerge in scene-level SfP, and we propose effective solutions to these challenges. 

\noindent \textbf{Surface normals from an RGB image.}
%
Even though RGB data does not directly contain geometry cues for objects such as polarization data, estimating the surface normal from a single RGB image is feasible, especially with the advent of deep learning. The related works~\cite{DBLP:pixelnet, Bansal16Marr, Li2015depth,Wang2015deep3d, zhang2016physically} train a neural network using a large amount of RGB-surface normal paired data, including real-world indoor dataset~\cite{silberman2012indoor_nyu} or synthetic dataset~\cite{dai2017scannet}. However, without the guidance of physics-based cues, these learning-based approaches mainly rely on semantic cues in the image, which leads to performance degradation when they are applied to data out of the training distribution (e.g., from indoor to outdoor data~\cite{chen2017surface} and from gravity-aligned to tilted images~\cite{Do2020SurfaceNormal}).

Some approaches attempt to utilize the relevance between surface normals and other information (e.g., depth, semantic information and shading). It has been shown that better surface normal estimation can be achieved by simultaneously estimating geometric information, such as depth~\cite{qi2018geonet,qi2020geonet++}, local principal axes~\cite{Huang_2019_ICCV_framenet}, Manhattan label map~\cite{Wang2020VPLNet}, planes and edges~\cite{Wang2016surge}. Eigen and Fergus~\cite{eigen2015predicting} and Zhang et al.~\cite{zhang2019pattern} jointly predict depth, normals, and semantics, exploiting affinity between these three modalities. Zamir et al.~\cite{Zamir2020Robust} consider the consistency of normals and other attributes, such as shading, depth, occlusion, and curvature. Surface normal estimation is also an essential element in inverse rendering, which aims to recover normals, reflectance, and illumination from one image~\cite{DBLP:journals/pami/BarronM15SIRFS,li2020inverse,yu2019inverserendernet} or multiple images~\cite{KimTO2016Multi-view,ZhaoMO2020Polarimetric}. However, according to our experiments, these approaches mainly depend on the semantic information of images for normal estimation. As a comparison, our method can better recover the physical geometry with the polarization cues, as shown in Fig.~\ref{fig:percep_physics}.

\begin{figure*}[t]
\centering
\begin{tabular}{@{}c@{\hspace{1mm}}c@{\hspace{1mm}}c@{\hspace{1mm}}c@{\hspace{1mm}}c@{\hspace{1mm}}c@{}}

\includegraphics[width=0.159\linewidth,height=0.131\linewidth]{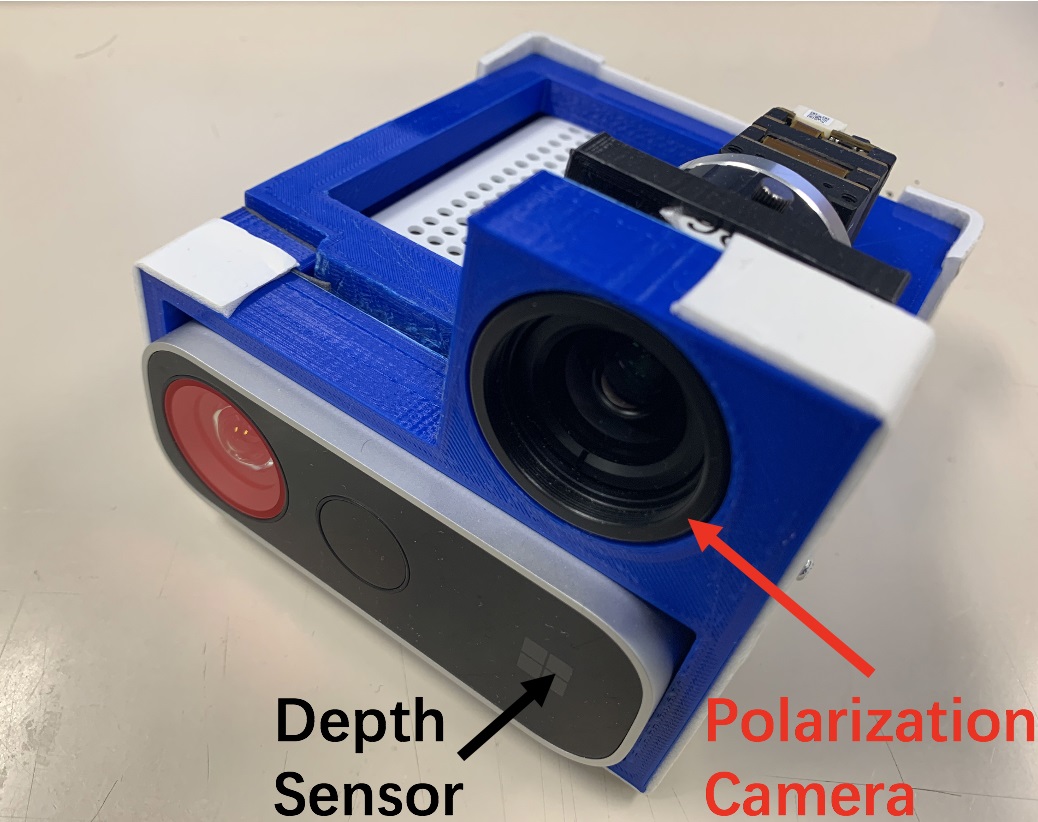}&
\includegraphics[width=0.159\linewidth,height=0.131\linewidth]{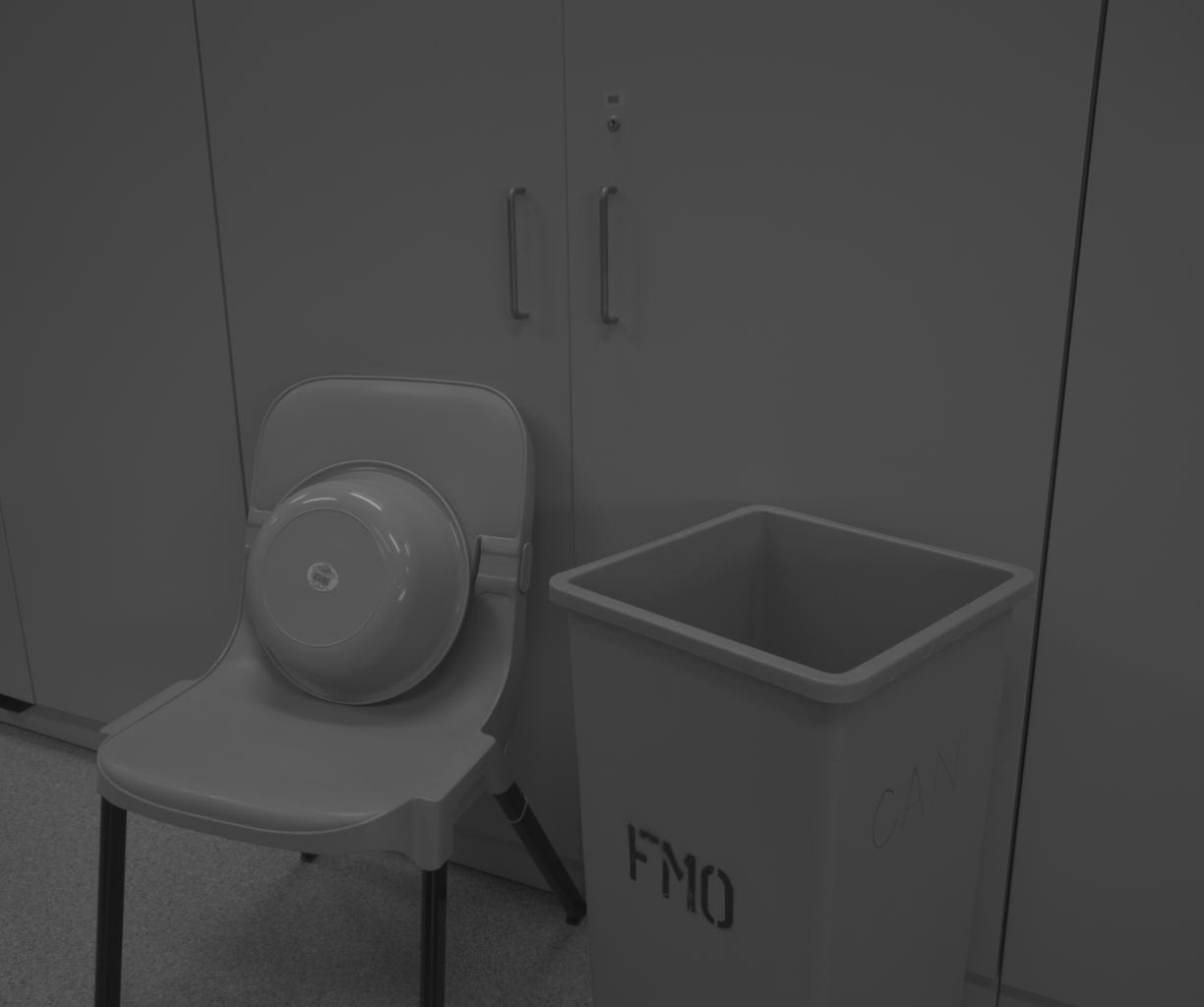}&
\includegraphics[width=0.159\linewidth,height=0.131\linewidth]{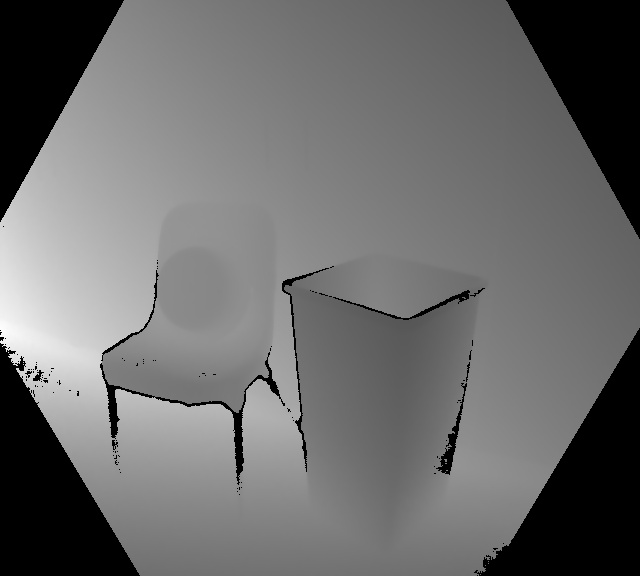}&

\includegraphics[width=0.159\linewidth,height=0.131\linewidth]{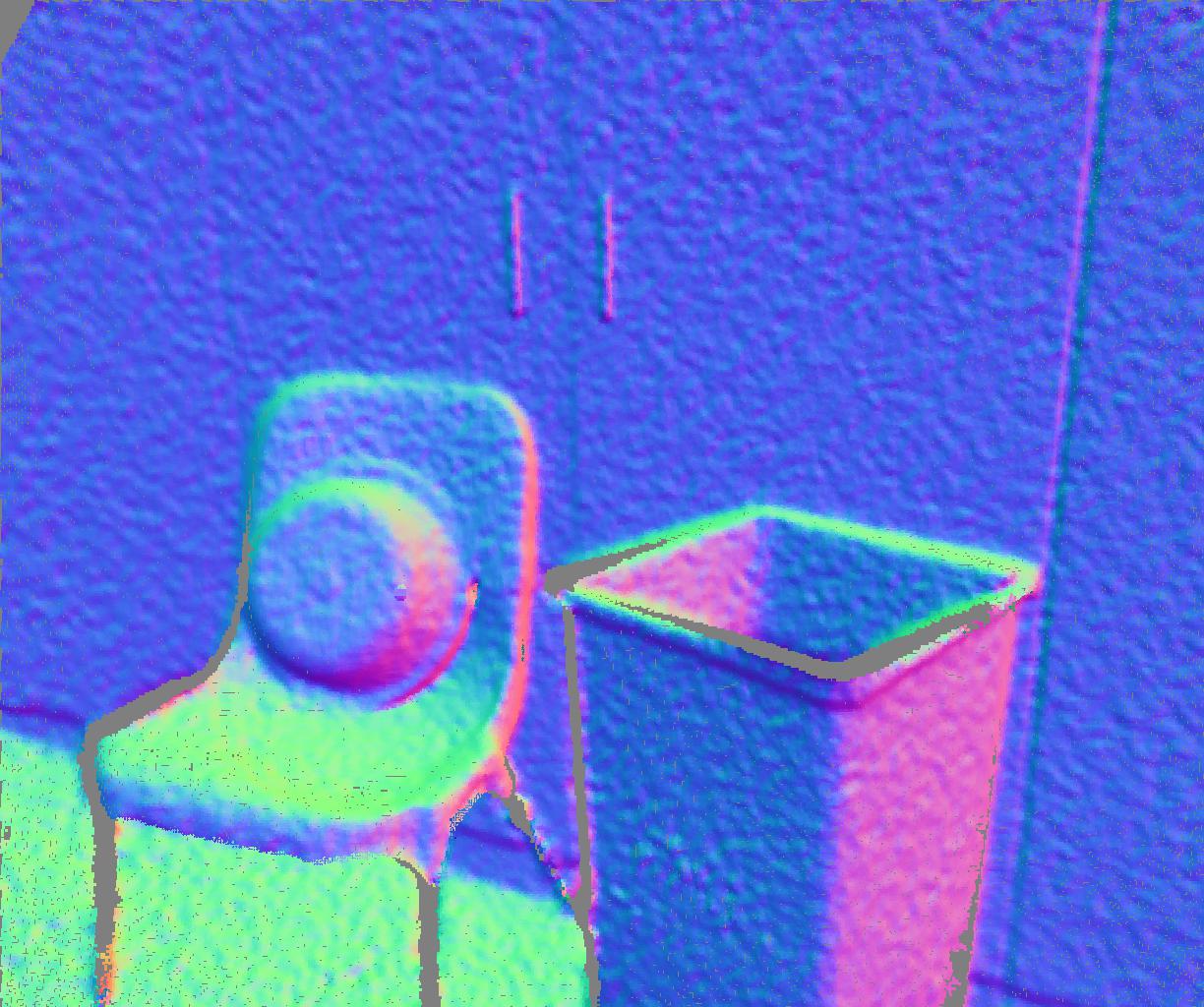} &
\includegraphics[width=0.159\linewidth,height=0.131\linewidth]{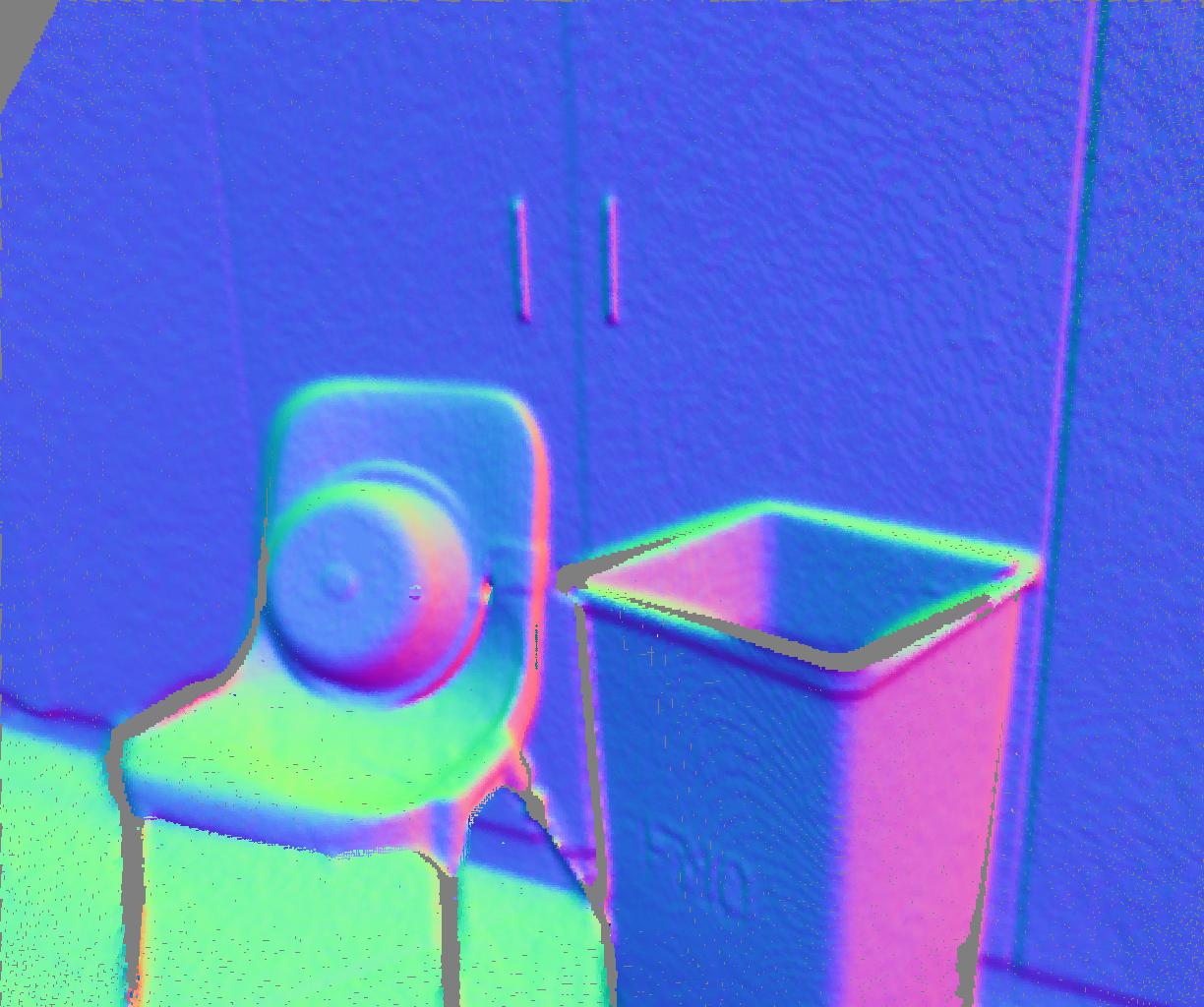} &
\includegraphics[width=0.159\linewidth,height=0.131\linewidth]{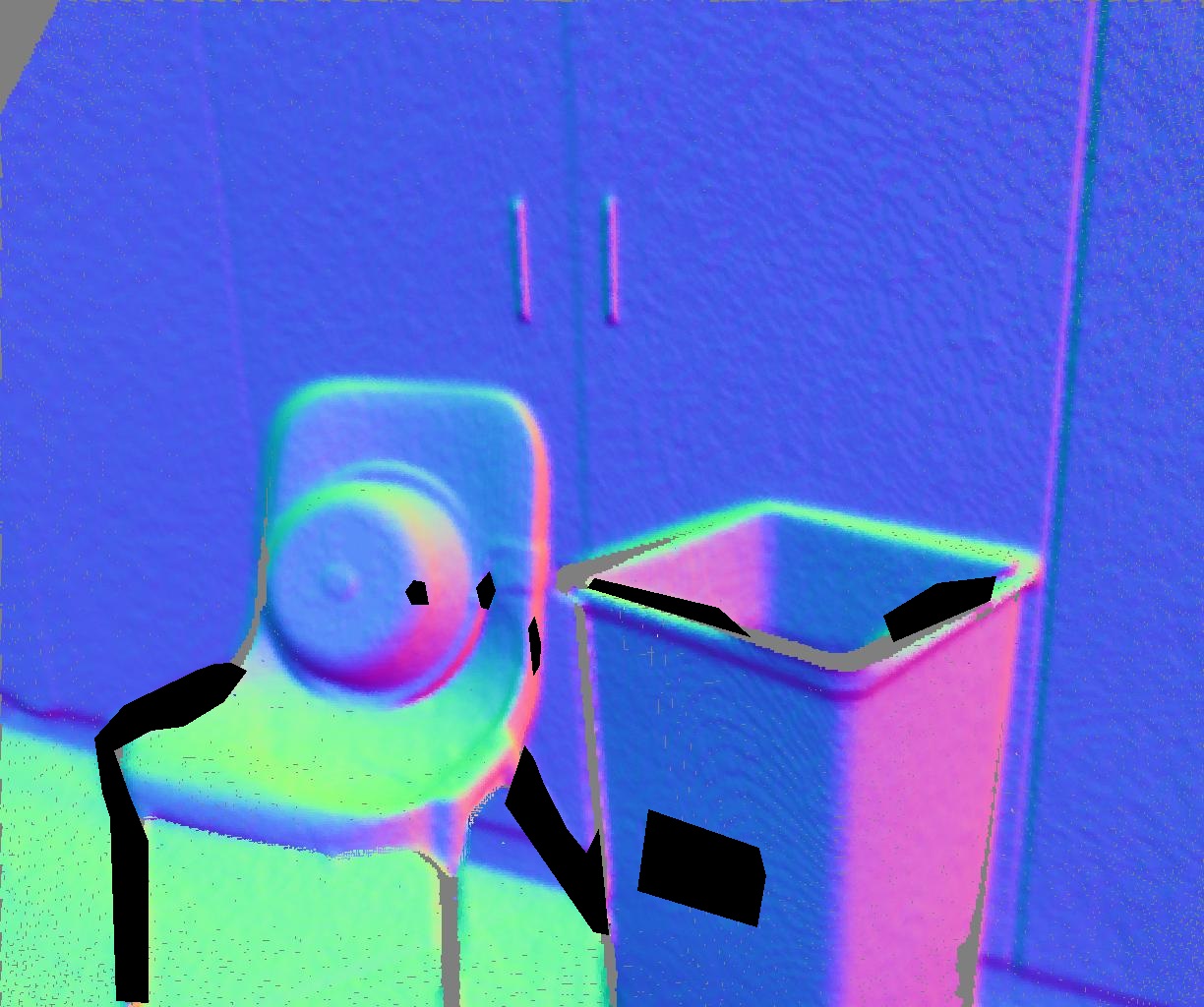}\\

\small{(a) Capture setup} & \small{(b) Polarization image} &  \small{(c) Depth map} & \small{(d) Coarse normal}&  \small{(e) Denoised normal} & \small{(f) Final normal} \\

\end{tabular}
\caption{\textbf{Our capture setup (a)} fixes a polarization camera (red arrow) and a depth sensor (black arrow) with a custom mount. The polarization camera captures polarization images (b), and the depth sensor collects scene-level depth (c). We use PCA-based normal estimation to obtain normal maps from the depth data, as shown in (d) and (e). (d) uses a single depth map, and (e) uses a median-denoised depth map. Post-processing the denoised normal map in (e) yields the final normal map shown in (f), there we exclude normals in areas where the depth sensor returns the inaccurate or sparse depth (e.g., dark region). }
\label{fig:dataset}
\end{figure*}

\section{SPW Dataset}
\label{sec:dataset}



We construct the SPW (Shape from Polarization in the Wild) dataset, the first real-world dataset that contains scene-level ground-truth surface normals for polarization images in the wild. Table~\ref{table:Dataset_cmp} provides a comparison between SPW and prior SfP datasets. The only existing real-world SfP dataset is DeepSfP~\cite{ba2020deep}; however, it only provides ground-truth normals on masked objects. Kondo et al.~\cite{kondo2020accurate} built a big SfP dataset, but it is synthetic and not publicly available. 

Our dataset consists of 522 sets of images from 110 different scenes containing diverse object materials and lighting conditions, and each scene includes 3-7 sets of images with different depths and viewing directions. A polarization image and the corresponding normal map are provided in each set. 
In addition to these image sets with ground truth normals, we also capture a separate set of images in outdoor scenes. Since most depth sensors cannot easily acquire dense depth for faraway scene content, these images are used for perceptual evaluation only, to assess generalization beyond the depth range of the training data.

\begin{table}[t]
\small
\centering
\renewcommand{\arraystretch}{1.2}
\begin{tabular}{@{}l@{\hspace{2mm}}c@{\hspace{2mm}}c@{\hspace{2mm}}c@{\hspace{2mm}}c@{\hspace{2mm}}c@{}}
\toprule[1pt]

Dataset &  Level  &  Collection & Size &Resolution & Public\\ 
    \midrule 
DeepSfP~\cite{ba2020deep}& Object& Real-world& 263& \small{1224$\times$1024} & \checkmark  \\ 
Kondo ~\cite{kondo2020accurate}&  Scene &  Synthetic &44305 &\small{256$\times$192} & $\times$  \\ 
Ours&  Scene &  Real-world &522 &\small{1224$\times$1024} & \checkmark   \\ 

\bottomrule[1pt]
\end{tabular}
\caption{\textbf{Comparison among different datasets.} DeepSfP~\cite{ba2020deep} is real-world but focuses on object-level, Konda et al.~\cite{kondo2020accurate} has a big dataset size, but it is synthetic and not publicly available. Ours is the first real-world scene-level SfP dataset.}
\label{table:Dataset_cmp}
\end{table}





Fig.~\ref{fig:dataset} shows our data preparation pipeline. The pipeline can be divided into the following four parts.

\textbf{a) Devices.} Since there is no existing polarization-depth camera, we need to choose a depth sensor to capture dense scene-level depth. We notice that most LiDAR cannot produce dense point clouds efficiently, and the 360-degree rotating device used in~\cite{ba2020deep} can only reconstruct small objects. Hence, we use a ToF sensor (Azure Kinect) to capture scene-level depth, and this depth sensor's resolution is $640\times 576$. For polarization, we use a PHX050S-P polarization camera that can capture four polarization images with polarizer angles of $0^{\circ}, 45^{\circ}, 90^{\circ}, 135^{\circ}$ in a single shot, and the resolution of this polarization camera is $1224\times 1024$. We fix the two sensors with a custom mount to make sure the camera pose between the depth sensor and polarization camera is the same in each capture, as shown in Fig.~\ref{fig:dataset}(a).


\textbf{b) Depth-polarization alignment.} We obtain the intrinsic and the initial extrinsic parameters between the polarization and depth sensor from stereo calibration. We then use coordinate descent to improve depth-polarization alignment further. Specifically, we optimize the extrinsic parameters with fixed intrinsic parameters to minimize the projection error between the reprojected RGB image and the polarization image. The optimized extrinsic parameters are used to produce polarization-aligned depth.


\textbf{c) Depth denoising.} Since surface normals computed from a single polarization-aligned depth map are noisy, we capture 50 depth images with a stationary setup and compute the median at each pixel to reduce noise in the depth map. Finally, we generate a point cloud from the denoised depth map and calculate the surface normals from the aligned point cloud using Principal Component Analysis (PCA) from the Open3D library~\cite{Zhou2018open3d}. As shown in Fig.~\ref{fig:dataset}(d,e), denoising the depth map yields much cleaner normals. 



\textbf{d) Post-processing.}
Even though we get high-quality normals by the above steps, we further improve the quality by excluding normals in areas where the depth sensor returns inaccurate values, such as dark and occluded regions. We also exclude normals on thin structures where the depth sensor only captures very sparse point clouds, such as chair legs or wires. The final normals are shown in Fig.~\ref{fig:dataset}(f).



\begin{figure*}[t]
\centering
\begin{tabular}{@{}c@{}}

\includegraphics[width=1.0\linewidth]{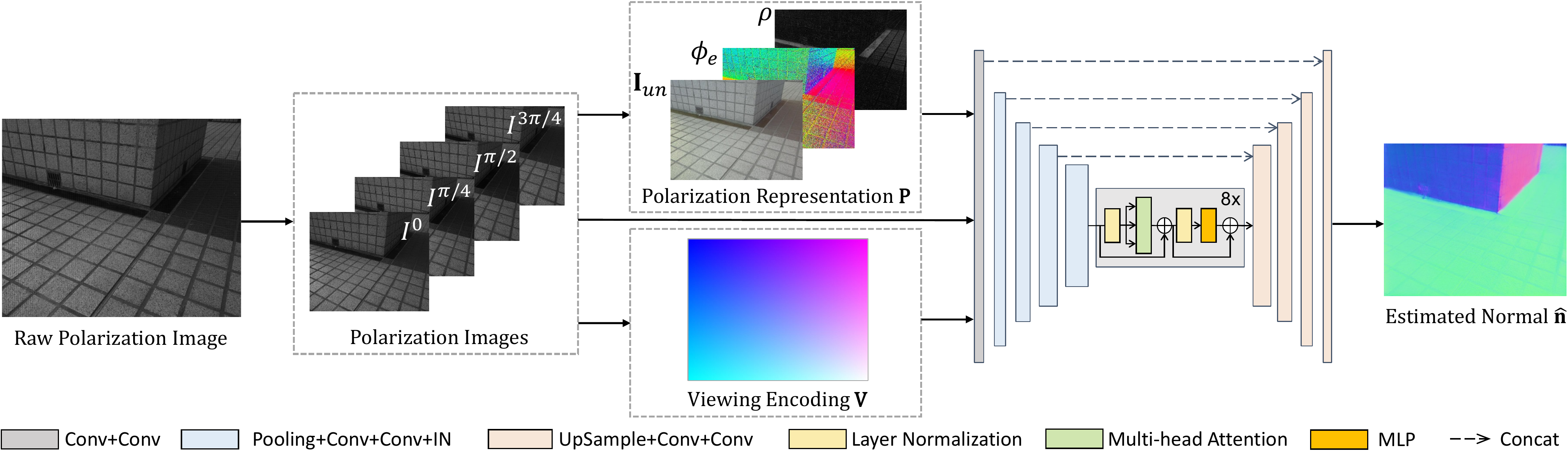}
\end{tabular}
\caption{\textbf{An overview of our approach.}
The input to our network includes three parts:
(1) Polarization images $\mathbf{I}^0, \mathbf{I}^{{\pi}/{4}}, \mathbf{I}^{{\pi}/{2}}, \mathbf{I}^{{3\pi}/{4}}$. (2) The polarization representation $\mathbf{I}_{un}$, $\phi_{e}$, and $\rho$ computed from polarization images. (3) The viewing encoding $\mathbf{V}$ is vital to handle the perspective projection for scene-level SfP. The concatenated inputs are fed into the neural network to output an estimated normal $\mathbf{\hat n}$.  }
\label{fig:framework}
\end{figure*}

\section{Method}

\label{sec:background}

In this paper, we consider linear polarization. A polarization camera can measure the intensity of light $\mathbf{I}^{\phi_{pol}}$ passing a polarizer~\cite{collett2005field,zhu2019depth}, which is determined by the polarizer angle $\phi_{pol}$ and the polarization of the light:
\begin{align}
\label{eq:polar}
    \mathbf{I}^{\phi_{pol}} =  \mathbf{I}_{un}\{1 + \rho{\rm cos}(2\phi-2\phi_{pol})\},  
\end{align}
where $\phi$ is the angle of polarization (AoP), $\rho$ is the degree of polarization (DoP), $\mathbf{I}_{un}$ is the unpolarized intensity of light. $\phi$, $\rho$, and $\mathbf{I}_{un}$ can be computed from images with  different polarizer angles by~\cite{LI2014Stokes,lei2020polarized}.



 \textbf{Degree of polarization (DoP)} $\rho$ contains cues for the viewing angle $\theta_\mathbf{v}$ between surface normal $\mathbf n$ and viewing direction $\mathbf v$. Specifically, the DoP $\rho$ is decided by the viewing angle $\theta_\mathbf{v}$, the refractive index $\eta$ of the object and reflection type $r$ (specular or diffuse reflection). More details are provided in the supplement.
 


 \textbf{Angle of polarization (AoP) $\phi$} is the projection of the polarization direction $\mathbf{d}$ on the image plane. In terms of physical properties, $\mathbf{d}$ is always parallel or perpendicular to the incidence plane, which is defined by surface normal $\mathbf{n}$ and viewing direction $\mathbf{v}$. There are two ambiguities for the polarization angle: $\pi$-ambiguity and diffuse/specular-ambiguity. The $\pi$-ambiguity is because $\phi$ is from 0 to $\pi$ and there is no difference between $\phi$ and $\phi + \pi$ (Eq.~\ref{eq:polar}). The reflection type causes the diffuse/specular-ambiguity: the polarization direction is parallel or perpendicular to the incidence plane respectively for diffuse/specular dominant reflection.

\subsection{Overview}
Fig.~\ref{fig:framework} provides an overview of our approach. The raw polarization image $\mathbf{I} \in \mathbb{R}^{H\times W \times 4}$ consists of four polarization images $\mathbf{I}^{\phi_{pol}}\in \mathbb{R}^{H\times W \times 1}$ under four polarizer angles $\phi_{pol} \in \{0, \pi/4, \pi/2, 3\pi/4 \}$. We firstly compute a polarization representation $\mathbf{P}$ for normal estimation (Sec.~\ref{sec:polarprior}). Then, to handle the perspective projection for scene-level SfP, we provide the viewing encoding $\mathbf{V}$ as an extra input (Sec.~\ref{sec:viewing_encoding}). At last, we predict the normal $\mathbf{\hat n}$ from all the provided information with our designed architecture (Sec.~\ref{sec:architecture}). 

\subsection{Polarization representation}
\label{sec:polarprior}
Having a proper polarization representation $\mathbf{P}$ as the input to a neural network can effectively improve the performance of SfP~\cite{ba2020deep,kondo2020accurate}. Kondo et al.~\cite{kondo2020accurate} directly compute AoP $\phi$, DoP $\rho$ and $I_{un}$ as the polarization representation. DeepSfP~\cite{ba2020deep} calculates possible SfP solutions under the assumption of orthographic projection:
\begin{align}
    \label{eq:normal_2angle}
    \mathbf{n} = (\rm{sin}\theta \rm{cos}\alpha,\rm{sin}\theta \rm{sin}\alpha , \rm{cos}\theta)^\intercal,
\end{align}
where $\theta$ and $\alpha$ are the zenith angle and azimuth angle computed from DoP $\rho$ and AoP $\phi$, respectively. While it is effective, computing their polarization representation is quite time-consuming, as reported in their paper~\cite{ba2020deep}.

We propose a new polarization representation $\mathbf{P}$ that is efficient and more effective compared with existing polarization representations~\cite{ba2020deep,kondo2020accurate}, as shown in our experiments (Table~\ref{table:AblationPolar}). $\mathbf{P} \in \mathbb{R}^{H\times W \times 4}$ consists of $\mathbf{I}_{un}, \phi_e, \rho$, where $\phi_e$ is the encoded AoP: 
\begin{align}
    \phi_e = (cos 2\phi, sin 2\phi).
\end{align}
The encoded AoP $\phi_e$ is designed to address a weakness of raw AoP $\phi$. For example, given two polarization angles $0^\circ$ and $179^\circ$, the distance between them should be $1^\circ$ in physics for polarization. However, in the calculated $\phi$ space, the difference is $179^\circ$. Encoding helps solve the weakness of raw AoP representation since there is no difference between $\phi$ and $\phi + \pi$ in the encoding space. 

We input the DoP $\rho$ as cues for solving specular/diffuse ambiguity since the DoP $\rho$ is usually large when specular reflection dominates. This strategy improves the performance but does not fully resolve the specular/diffuse ambiguity. 

\paragraph{Importance of polarization.} 
Polarization contains useful cues about physical 3D information of objects based on real-world reflection. Thus, utilizing polarization can improve the fidelity of estimated normals, especially for areas with rare or wrong semantic information. Fig.~\ref{fig:percep_physics} shows an example about the advantage of polarization: given an image printed on a flat sheet of paper, the RGB-based baselines are distracted by the content of the image and fail to predict correct normals for the physical content of the scene (i.e., the flat paper). Polarization provides an alternative modality that can convey the true shape of objects in the scene. Hence, the polarization can give a robust cue to distinguish that the wall (paper) is exactly flat.

\begin{figure}[t]
\centering
\begin{tabular}{@{}c@{\hspace{1mm}}c@{\hspace{1mm}}c@{\hspace{1mm}}c@{}}
\includegraphics[width=0.245\linewidth]{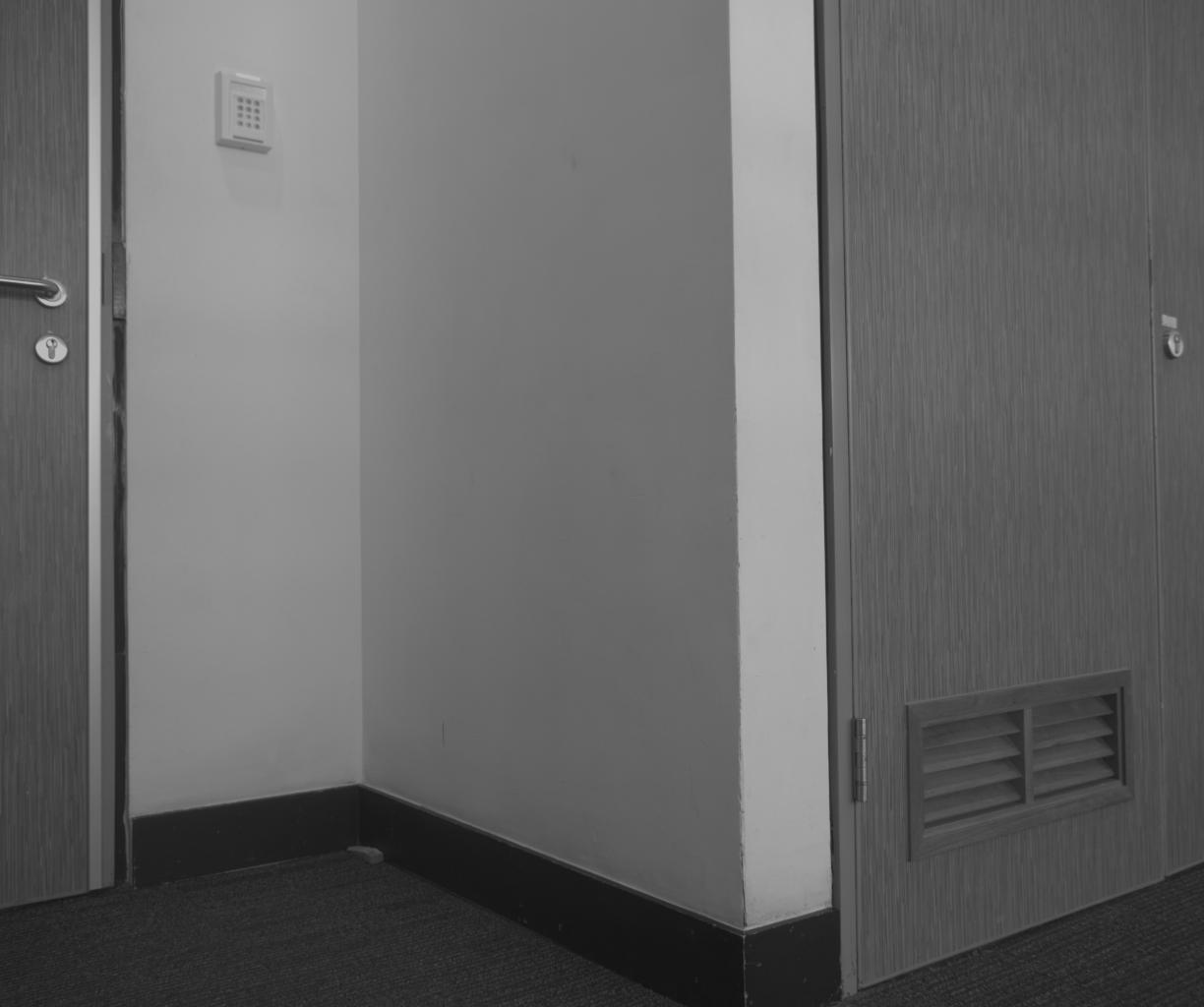}&
\includegraphics[width=0.245\linewidth]{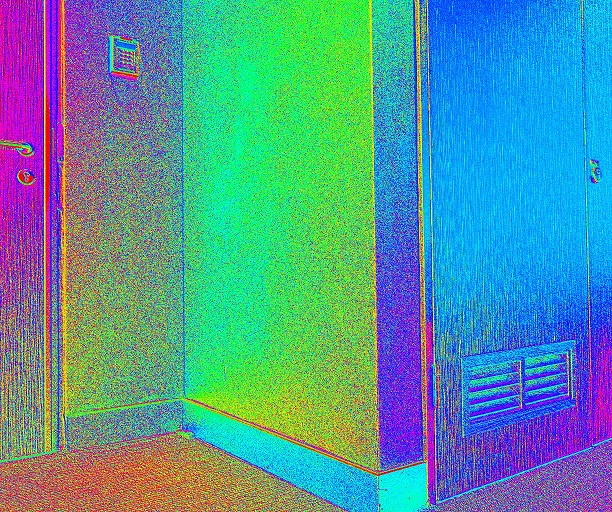}&
\includegraphics[width=0.245\linewidth]{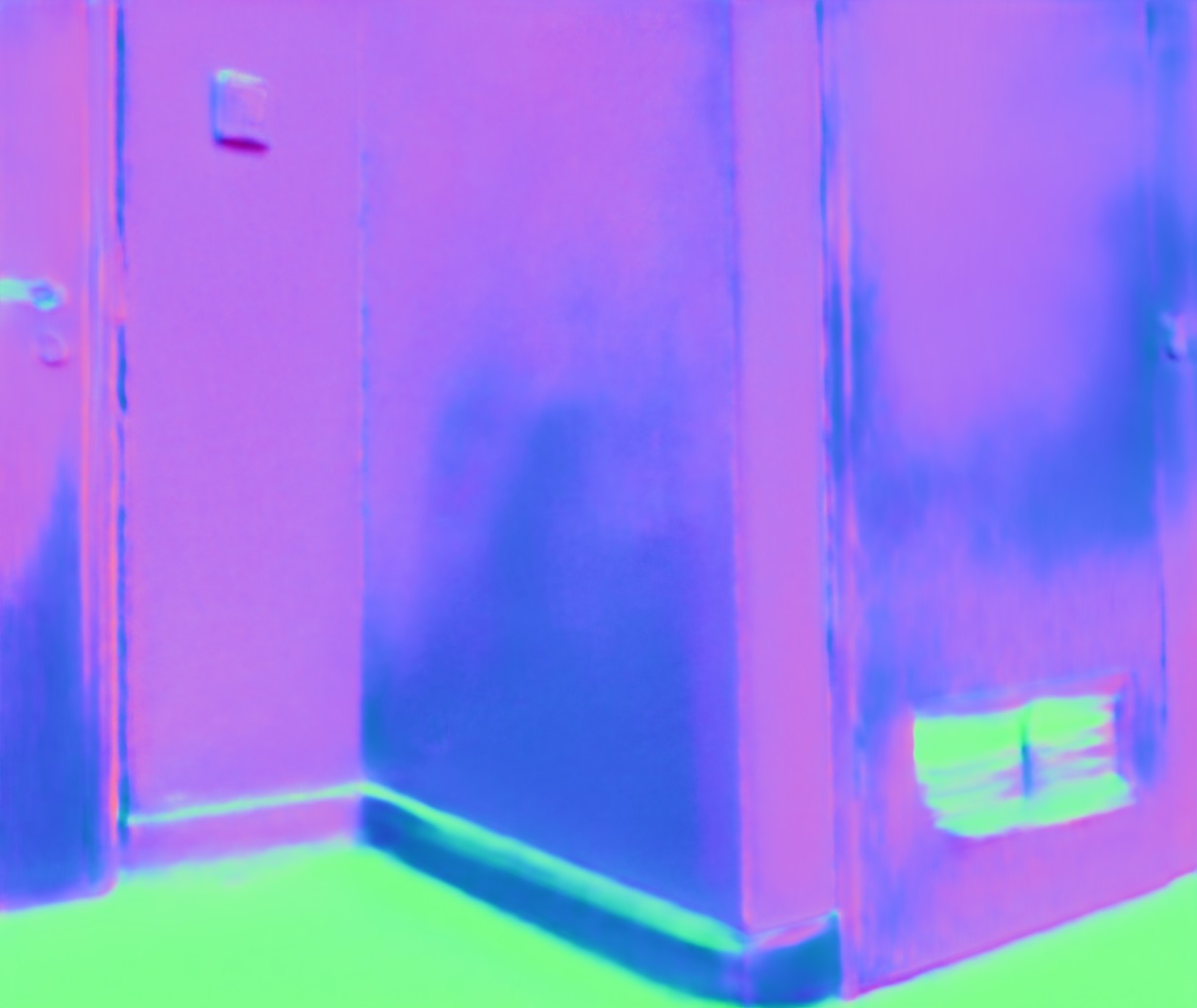}&
\includegraphics[width=0.245\linewidth]{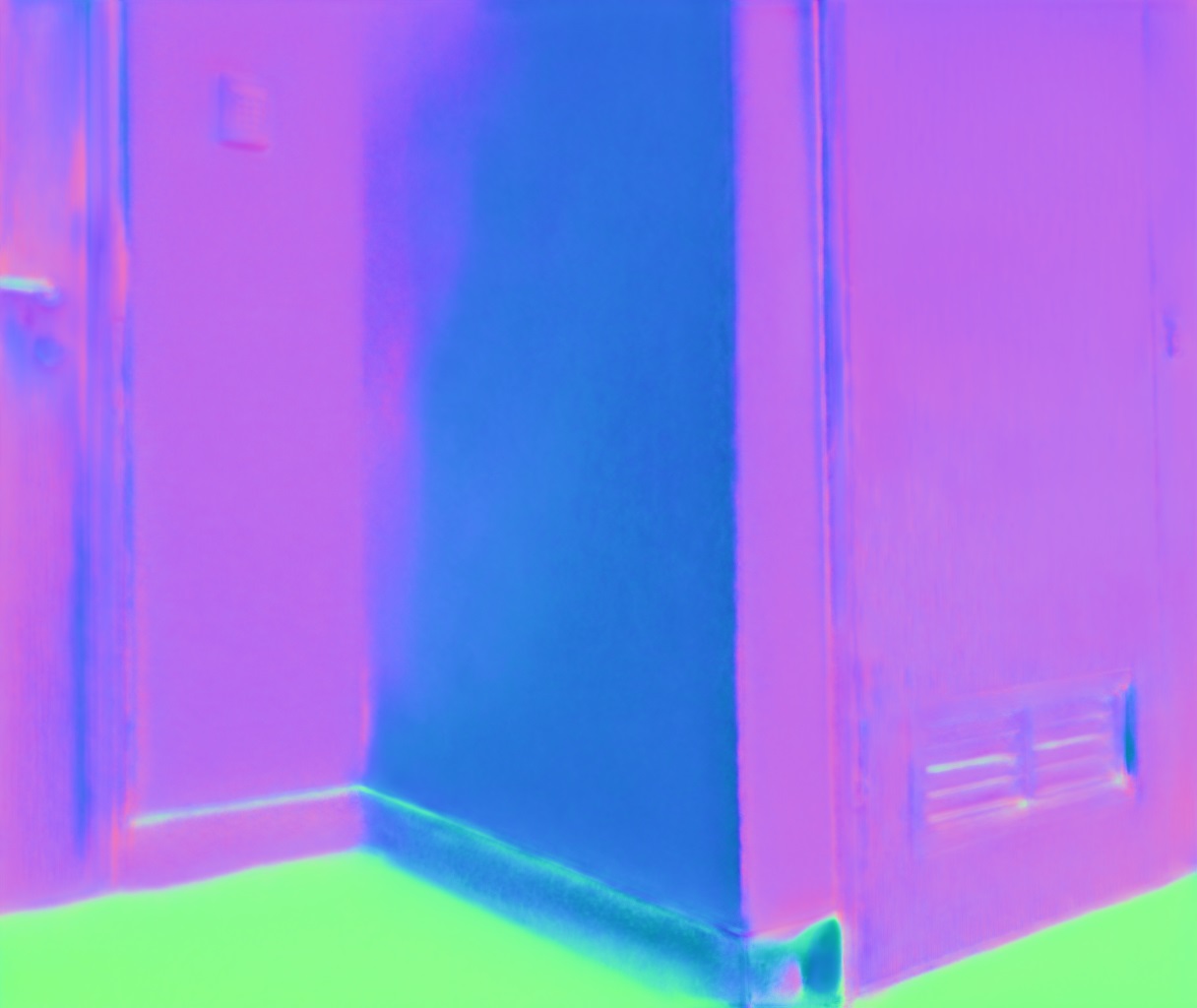}\\

\small{Input} $\mathbf{I}_{un}$ & \small{Input $\phi$}  & \small{Without $\mathbf{V}$} & \small{With $\mathbf{V}$}\\

\end{tabular}
\caption{\textbf{An analysis of our proposed viewing encoding $\mathbf{V}$.} The polarization representation is affected by spatial-varying viewing directions in scene-level SfP. Enforcing the viewing encoding $\mathbf V$ can effectively calibrate the impact of viewing direction on the polarization representation.}
\label{fig:Abl-view}
\end{figure}

\begin{table*}[t]
\small
\centering
\renewcommand{\arraystretch}{1.2}
\begin{tabular}{l@{\hspace{5mm}}c@{\hspace{5mm}}c@{\hspace{3mm}}c@{\hspace{3mm}}c@{\hspace{5mm}}c@{\hspace{3mm}}c@{\hspace{3mm}}c}
\toprule[1pt]
\small{Method}&\small{Task} & \multicolumn{3}{c}{Angular Error $\downarrow$ } & \multicolumn{3}{c}{Accuracy $\uparrow$}  \\ 
 & & \small{Mean} & \small{Median}  & \small{RMSE}& \small{$11.25^{\circ}$} & \small{$22.5^{\circ}$} & \small{$30.0^{\circ}$} \\
 
\midrule[0.6pt]

 \small{Miyazaki et al.~\cite{miyazaki2003polarization}}  &\small{Physics-based SfP} & 
  55.34 & 55.19 & 60.35 & 2.6 & 10.4 & 18.8\\ 
  
 \small{Mahmoud et al.~\cite{mahmoud2012direct}} &\small{Physics-based SfP} & 
 52.14  & 51.93  & 56.97  & 2.7  & 11.6  & 21.0\\

 \small{Smith et al.~\cite{smith2019height}}  &\small{Physics-based SfP} & 
 50.42  & 47.17 & 55.53  & 11.0 & 24.7  & 33.2 \\ 

 \small{DeepSfP$^\dag$ ~\cite{ba2020deep}}  &\small{Learning-based SfP} & 
28.43 & 24.90 & 33.17 & 18.8 & 48.3 & 62.3  \\ 
\small{Kondo et al.$^\dag$~\cite{kondo2020accurate}}  &\small{Learning-based SfP} & 
28.59 & 25.41 &33.54 & 17.5 & 47.1 & 62.6  \\

 \small{Ours}       &\small{Learning-based SfP}  &         
 \textbf{17.86} & \textbf{14.20} & \textbf{22.72} & \textbf{44.6} & \textbf{76.3} & \textbf{85.2} \\ 
\bottomrule[1pt]
\end{tabular}

\caption{\textbf{Quantitative evaluation on the SPW dataset.} Our approach outperforms all baselines by a large margin on all evaluation metrics. $\dag$: our implementation. }
\label{table:baseline_comparison}
\end{table*}

\begin{table}
\small
\centering
\renewcommand{\arraystretch}{1.2}
\begin{tabular}{l@{\hspace{3mm}}c}
\toprule[1pt]
Method & Mean Angular Error$\downarrow$ \\
\midrule
 {Miyazaki et al.~\cite{miyazaki2003polarization}}& {43.94}   \\ 
 {Mahmoud et al.~\cite{mahmoud2012direct}}& {51.79}  \\ 
 {Smith et al.~\cite{smith2019height}}& 45.39 \\ 
 {DeepSfP~\cite{ba2020deep}}& {18.52}   \\ 
 {Ours}& \textbf{{14.68}}     \\ 
\bottomrule[1pt]
\end{tabular}
\caption{\textbf{Quantitative evaluation on the DeepSfP dataset~\cite{ba2020deep}.} Our approach obtains the best score. The results of other baselines are collected from the official results in DeepSfP~\cite{ba2020deep}.}
\label{table:cmp_deepsfp}
\end{table}

\subsection{Viewing encoding}
\label{sec:viewing_encoding}
We introduce the viewing encoding $\mathbf V$ to account for non-orthographic projection in scene-level SfP, which contains the viewing direction cues for every pixel of the polarization representation. Previous object-level SfP approaches~\cite{atkinson2007shape_twoviews,ba2020deep} assume viewing directions are $(0,0,1)^\intercal$ for all pixels (i.e., orthographic projection) since an object is always put at the image center. However, the viewing direction is spatially varying in scene-level SfP, and the polarization representation is heavily influenced by the viewing direction. As shown in Fig.~\ref{fig:Abl-view}, for pixels with the same material and surface normal, their polarization representations are quite different. Also, since the CNN is translation-invariant, it is hard for CNN to know the viewing direction without explicitly providing it to the CNN. We thus propose to input the viewing encoding to the CNN.

A direct representation of viewing encoding is the viewing direction, which is computed from the intrinsic parameters of the polarization camera. If the intrinsic parameters of the camera are not available, we can also use the 2D coordinate $(u,v)$ of each pixel and normalize it to $[-1,1]$ as input~\cite{Liu2018an}. Fig.~\ref{fig:Abl-view} presents an example for the effectiveness of our viewing encoding. 





  \textbf{Discussion.} Note that our viewing encoding is different from the positional encoding used in NeRF~\cite{mildenhall2020nerf} or transformers~\cite{vaswani2017attention}. Our design is inspired by the fact that per-pixel viewing directions influence polarization. Besides, viewing encoding yields better performance than the positional encoding in our experiments.

\subsection{Network architecture and training}
\label{sec:architecture}
To handle the ambiguities that exist in polarization cues, we introduce multi-head self-attention~\cite{vaswani2017attention} to SfP for utilizing the global context information. As shown in Fig.~\ref{fig:framework}, the self-attention block is added in the bottleneck of an Encoder-Decoder architecture~\cite{Ronneberger2015Unet}. Different from similar architectures that combine CNN encoder with transformer~\cite{chen2021transunet,yang2021transformers}, we remove the linear projection layer since the CNN encoder already extracts the embeddings. Besides, similar to position embedding of transformer~\cite{chen2021transunet,yang2021transformers}, our viewing encoding can provide the position information to self-attention, and we thus remove the position embedding. Finally, we add instance normalization to the encoder since we notice it helps convergence.

We adopt a cosine similarity loss~\cite{ba2020deep} for training. We implement our model in PyTorch. The model is trained for 1000 epochs with batch size 16 on four Nvidia Tesla V100 GPUs, each with 16 GB memory. We use the Adam optimizer~\cite{Kingma2015Adam} with initial learning rate 1e-4 and we adopt a cosine decay scheduler for the learning rate. The learning rate is linearly scaled with the batch size. We crop images to 512×512 patches in each iteration for data augmentation.

\section{Experiments}

\subsection{Experimental setup}
\noindent \textbf{Evaluation metrics.} Following previous surface normal estimation works~\cite{DBLP:pixelnet,Wang2015deep3d}, we adopt six widely used metrics. The first three are \textit{Mean}, \textit{Median}, and \textit{RMSE} (lower is better $\downarrow$), which are the mean, median, and RMSE of angular errors. The last three are {$\mathit{11.25^{\circ}}$}, {$\mathit{22.5^{\circ}}$}, and {$\mathit{30.0^{\circ}}$} (higher is better $\uparrow$), and each shows the percentage of pixels within a specific angular error.

\noindent \textbf{Datasets.} We use two datasets in the experiments. 
\begin{itemize}
    \setlength{\itemsep}{0pt}
    \setlength{\parsep}{0pt}
    \setlength{\parskip}{0pt}
    \item \textbf{DeepSfP~\cite{ba2020deep}}. DeepSfP is the only publicly available SfP dataset that contains real-world ground-truth surface normals. There is only one object in each image, but the surface normal is high-quality. We use the train/test split provided in the original paper~\cite{ba2020deep}.
    \item\textbf{SPW}. We use our SPW dataset, presented in Sec.~\ref{sec:dataset}. We use 403 and 119 images for training and evaluation, respectively. Train and test sets do not include the images from the
same scene to avoid overfitting. We also use the far-field data for perceptual evaluation.
randomly divided based on scenes (instead of images). 
\end{itemize}

\begin{figure*}
\centering
\begin{tabular}{@{}c@{\hspace{1mm}}c@{\hspace{1mm}}c@{\hspace{1mm}}c@{\hspace{1mm}}c@{\hspace{1mm}}c@{\hspace{1mm}}c@{\hspace{1mm}}c@{}}

\includegraphics[width=0.120\linewidth]{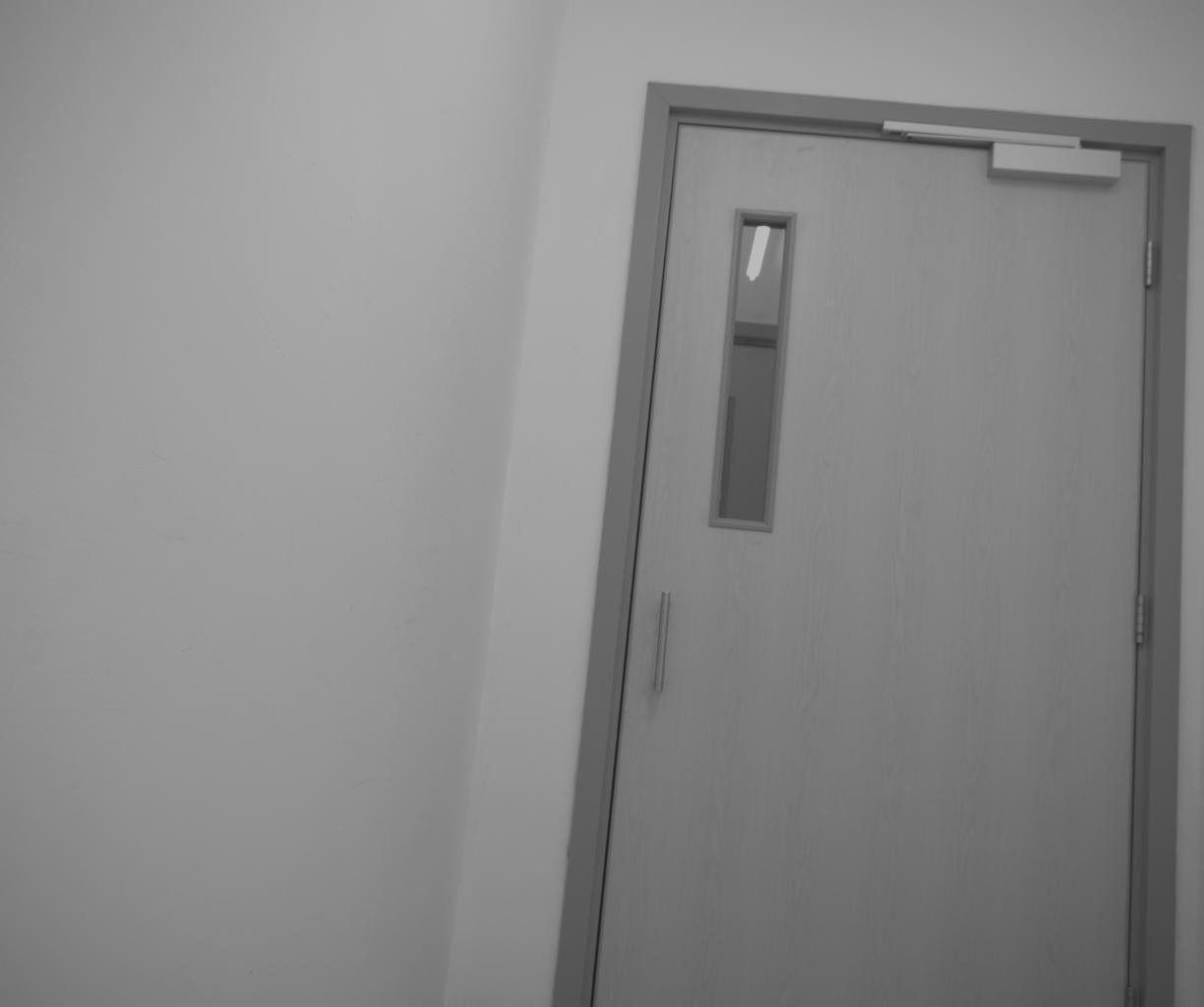}&
\includegraphics[width=0.120\linewidth]{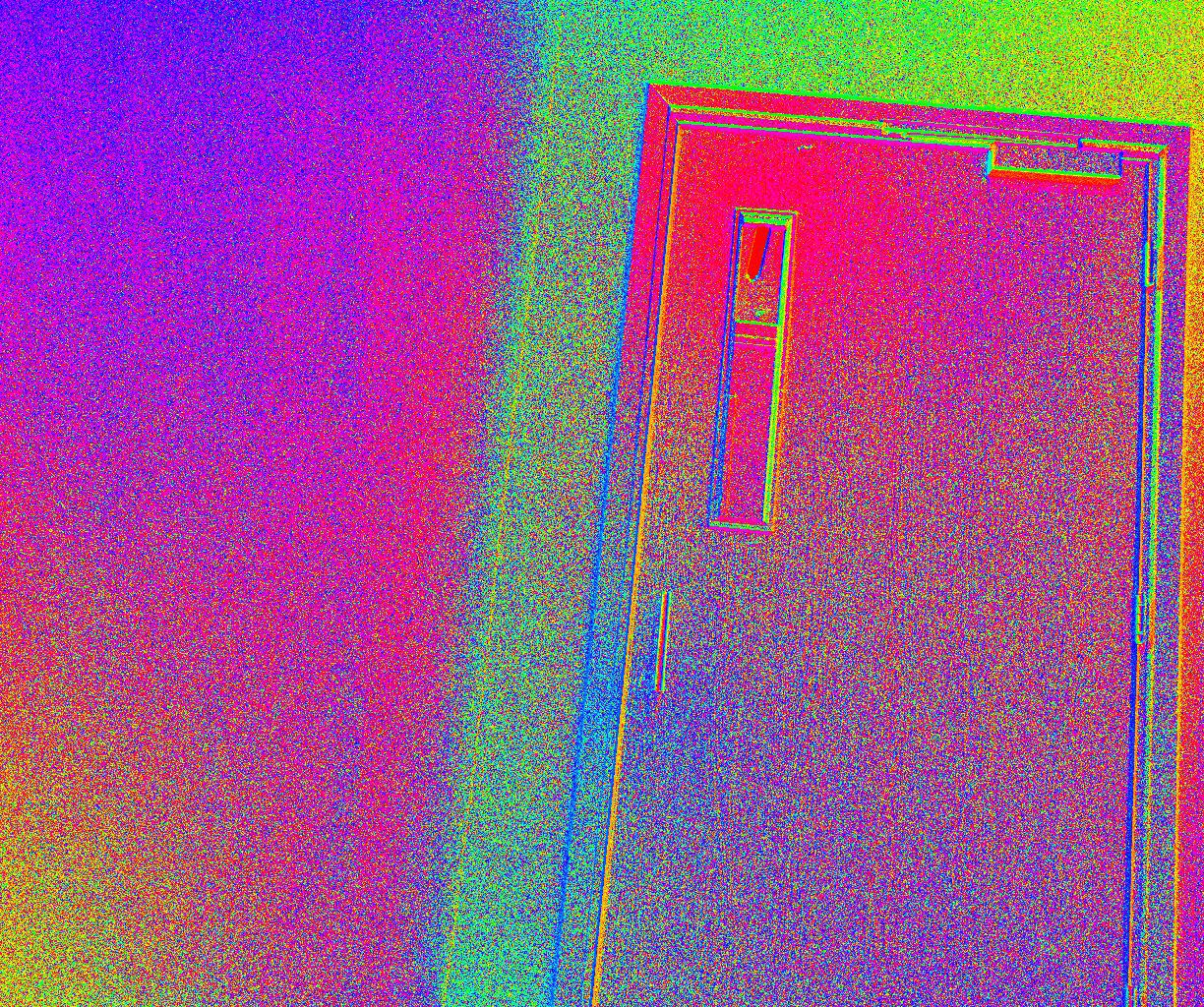}&
\includegraphics[width=0.120\linewidth]{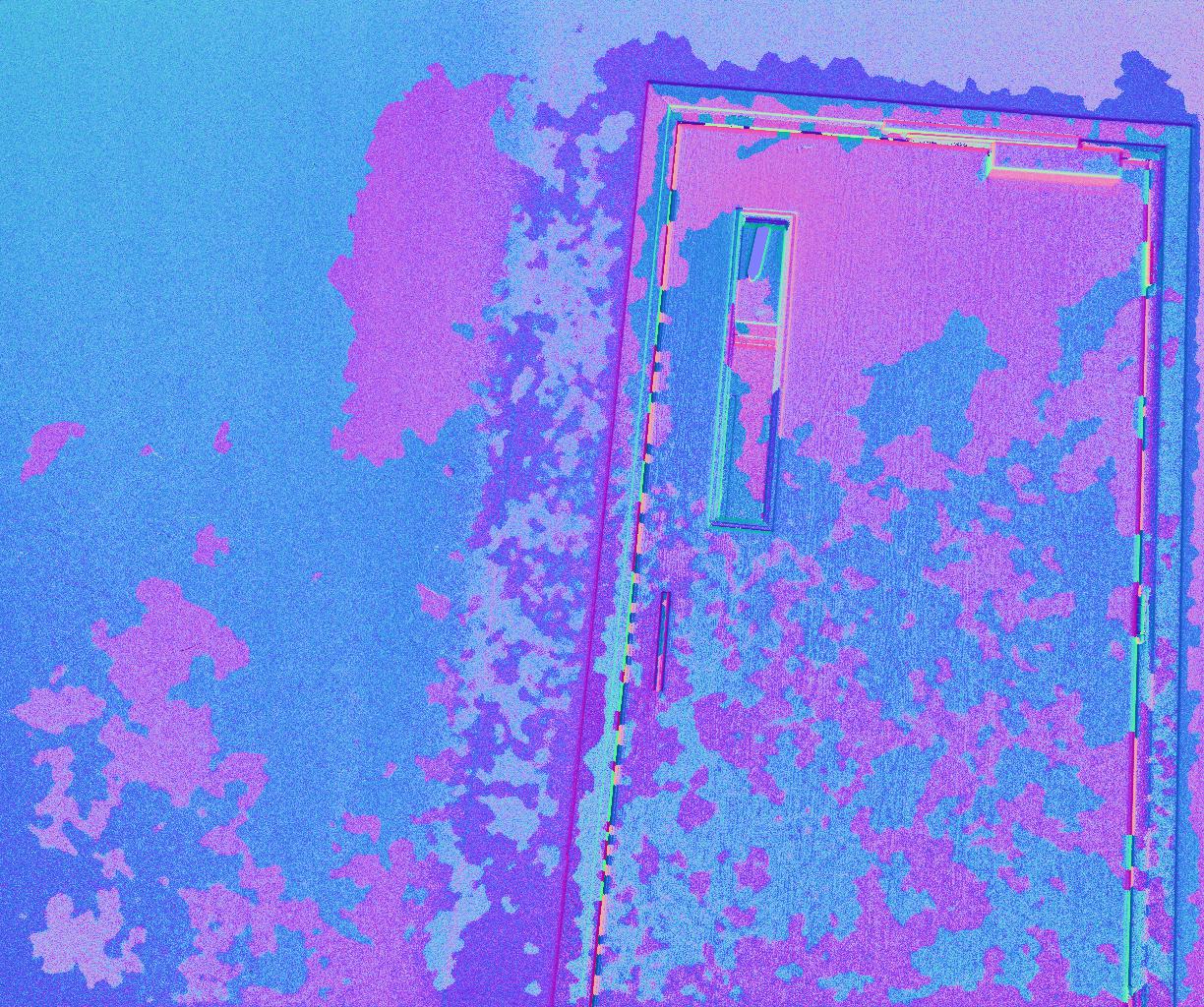}&
\includegraphics[width=0.120\linewidth]{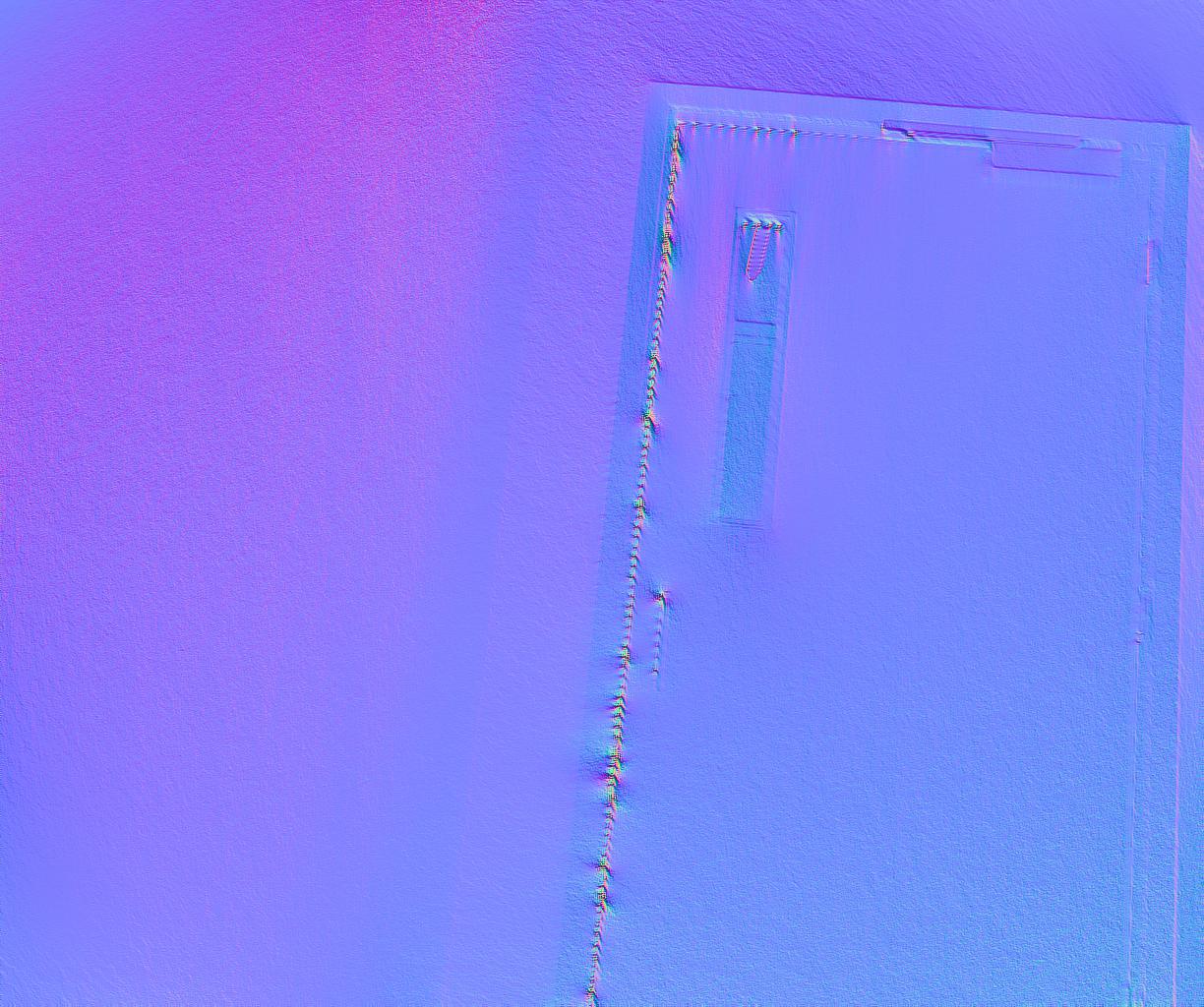}&
\includegraphics[width=0.120\linewidth]{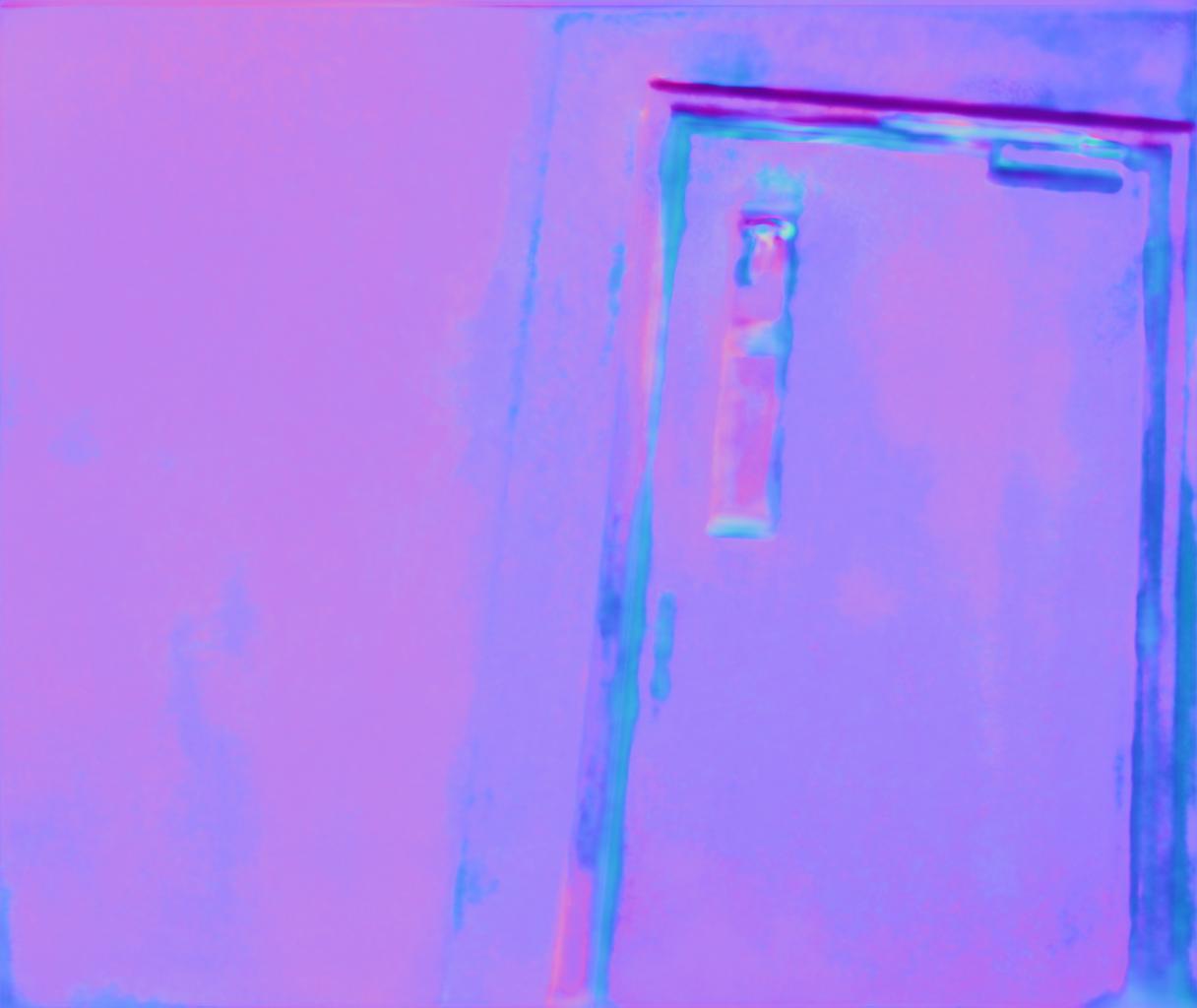}&
\includegraphics[width=0.120\linewidth]{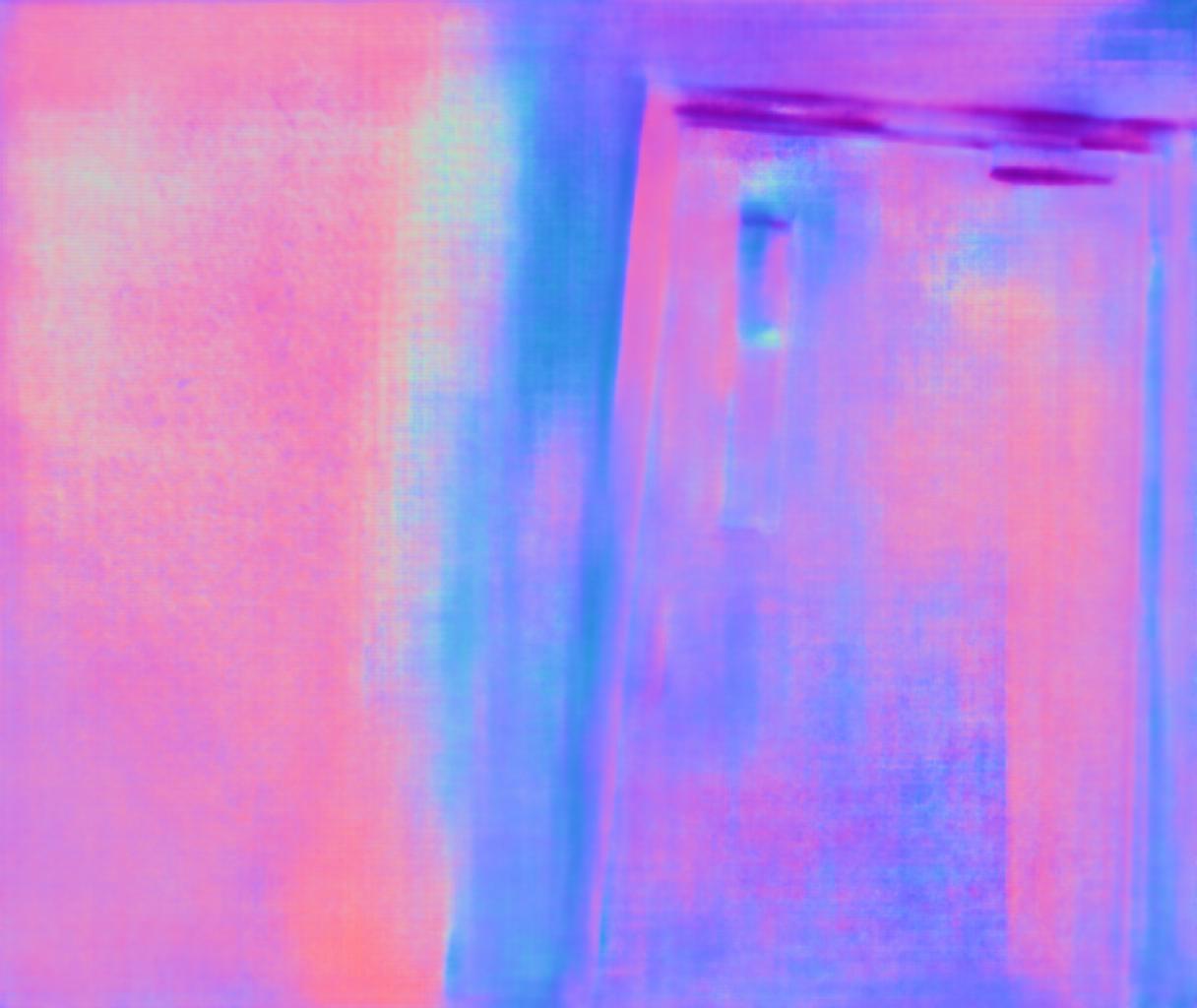}&
\includegraphics[width=0.120\linewidth]{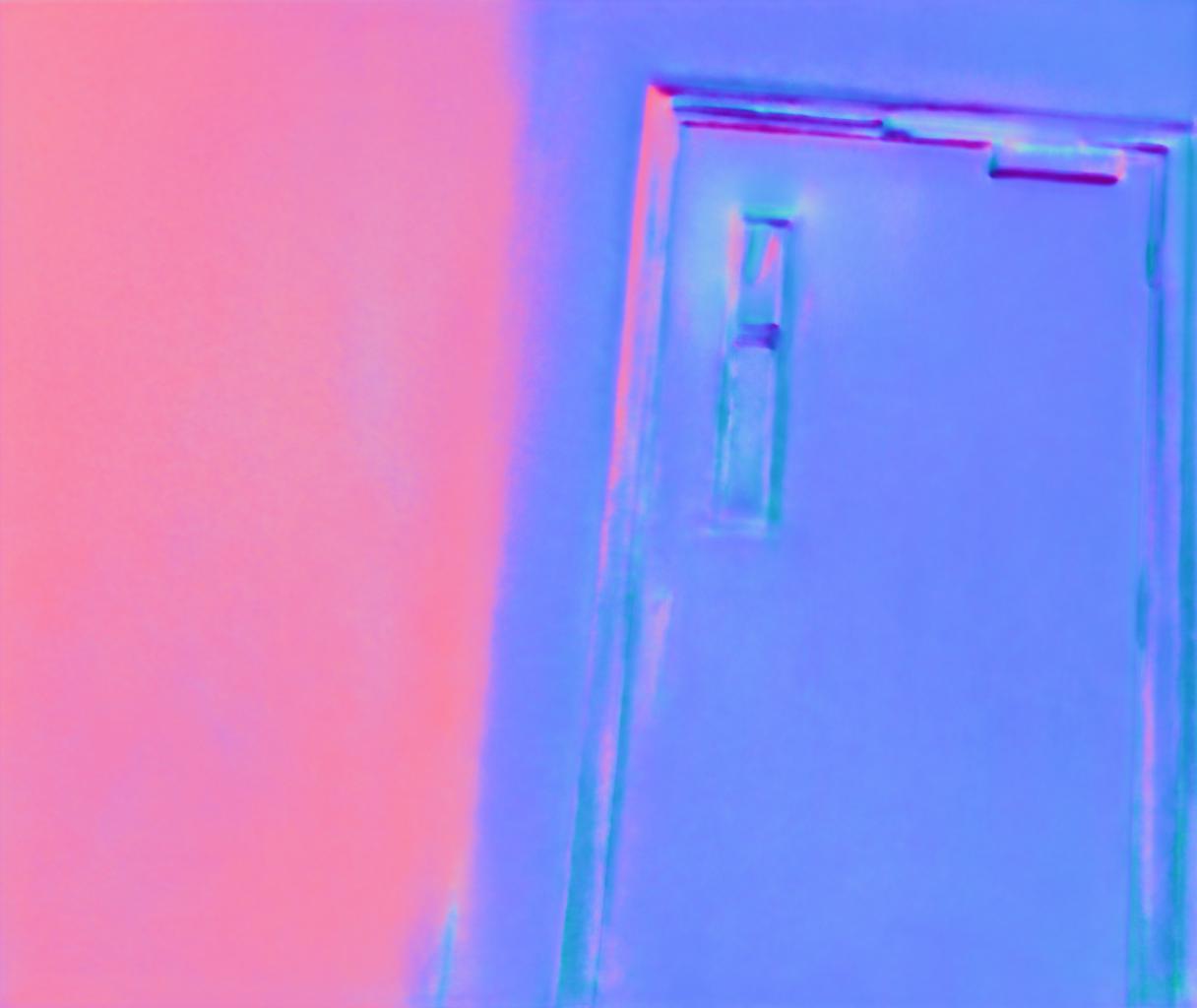}&
\includegraphics[width=0.120\linewidth]{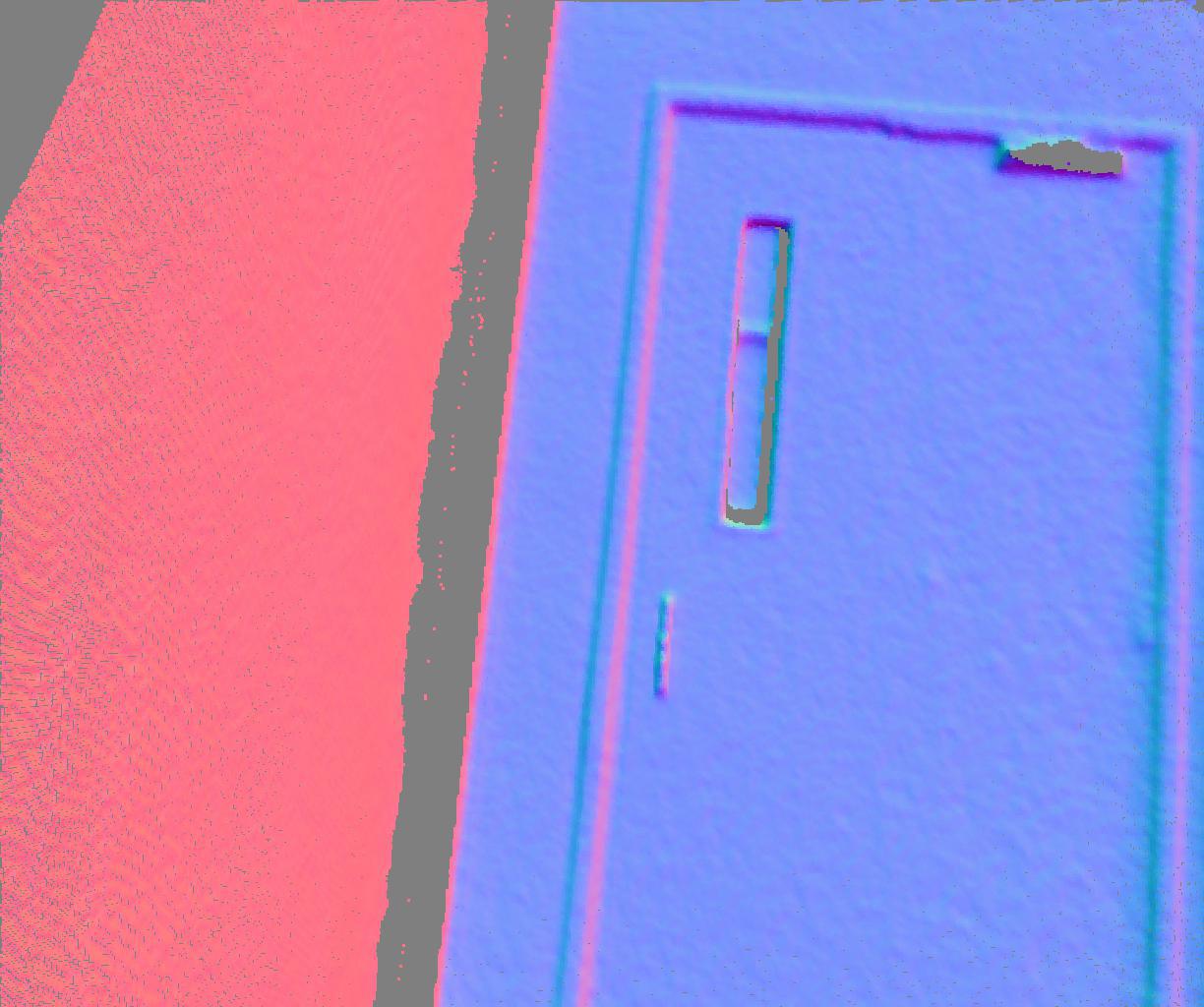}
\\
\includegraphics[width=0.120\linewidth]{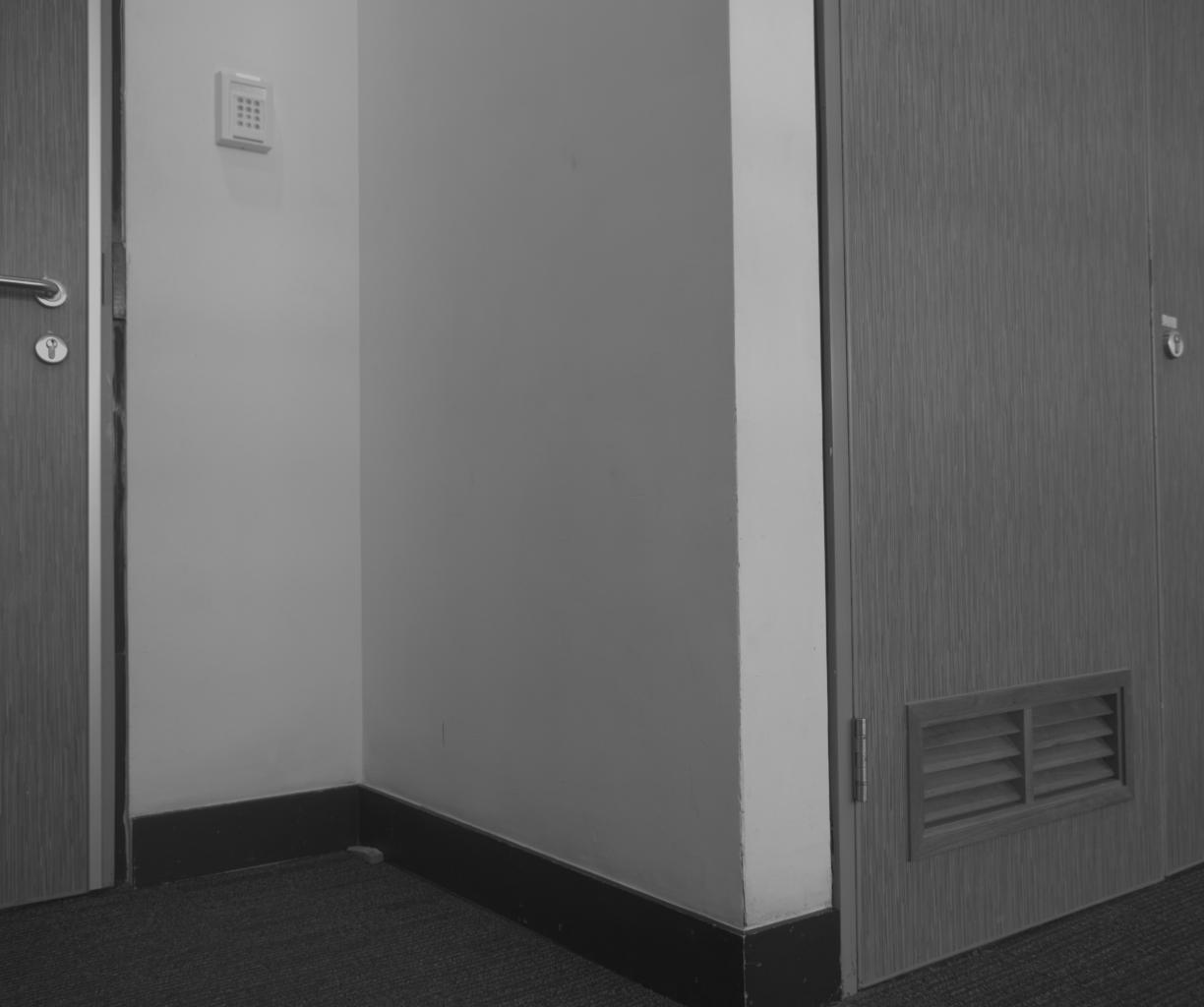}&
\includegraphics[width=0.120\linewidth]{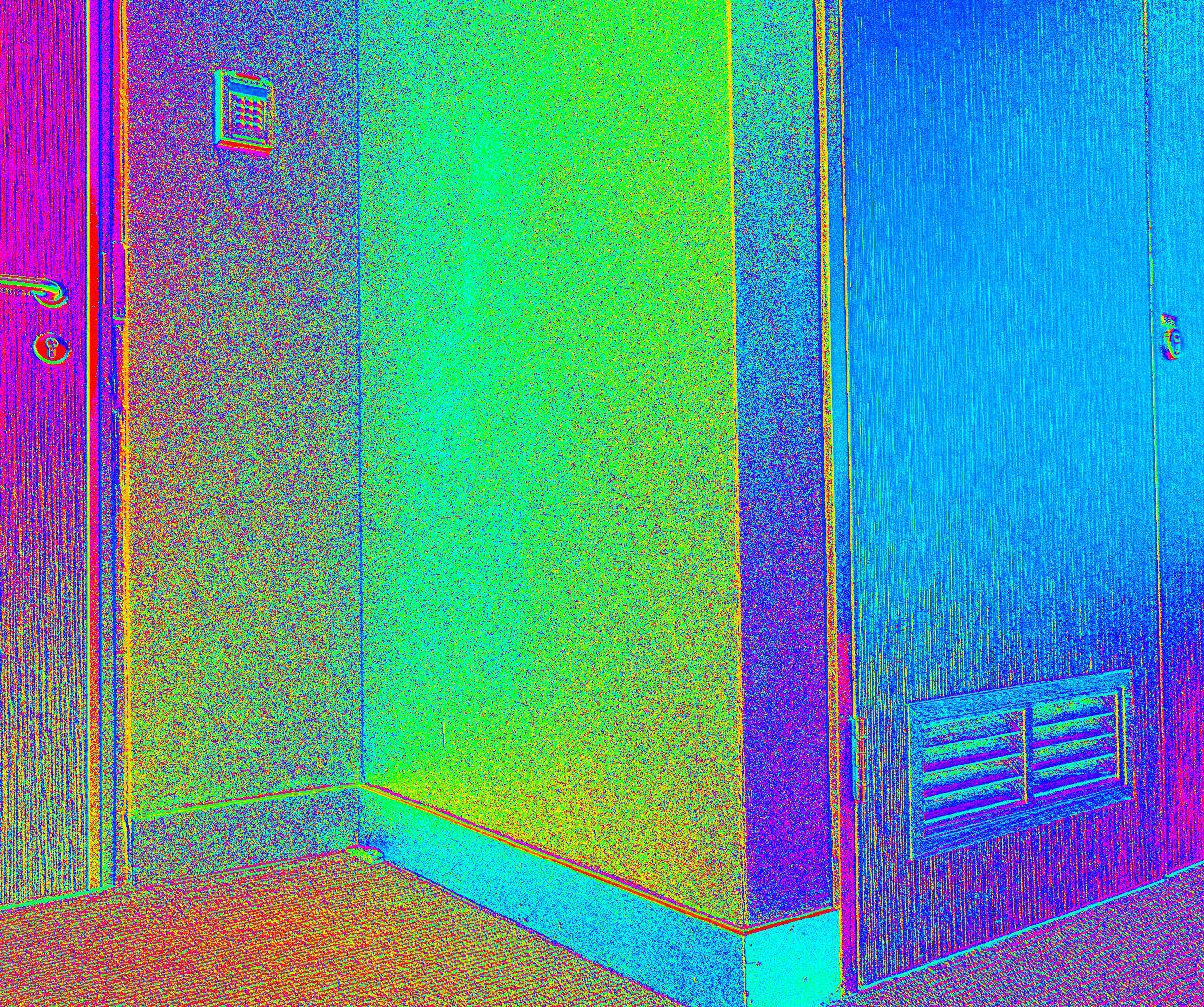}&
\includegraphics[width=0.120\linewidth]{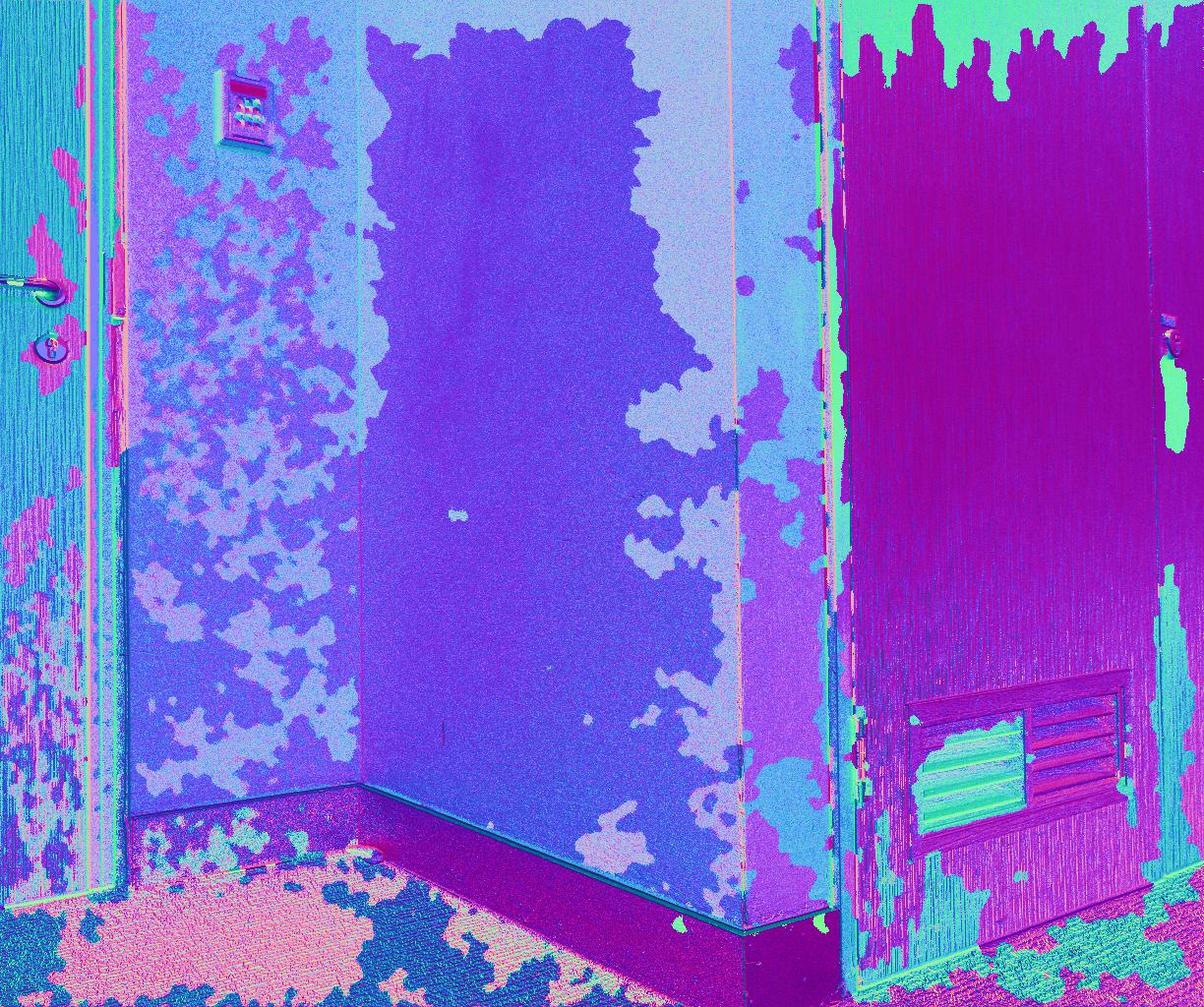}&
\includegraphics[width=0.120\linewidth]{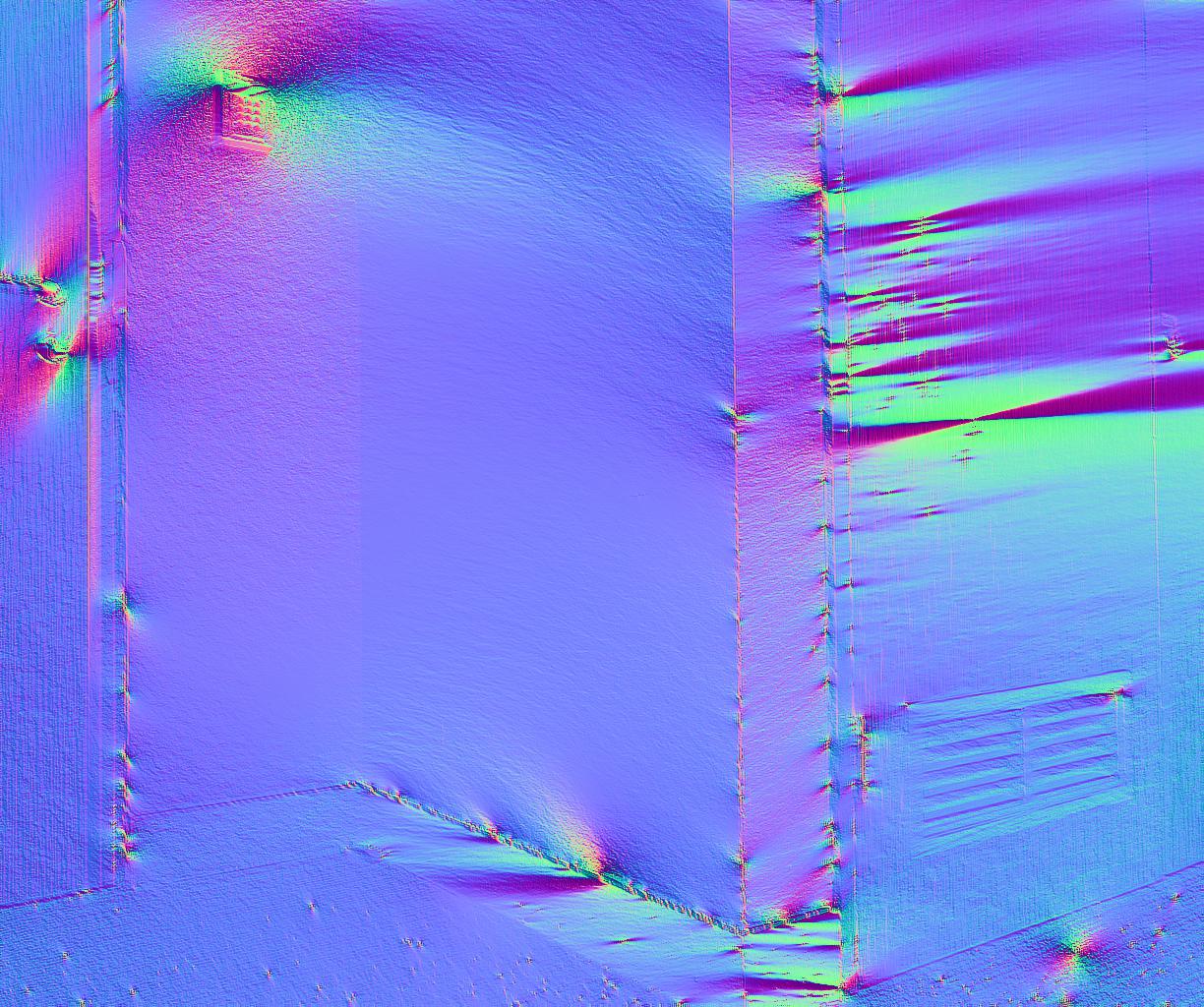}&
\includegraphics[width=0.120\linewidth]{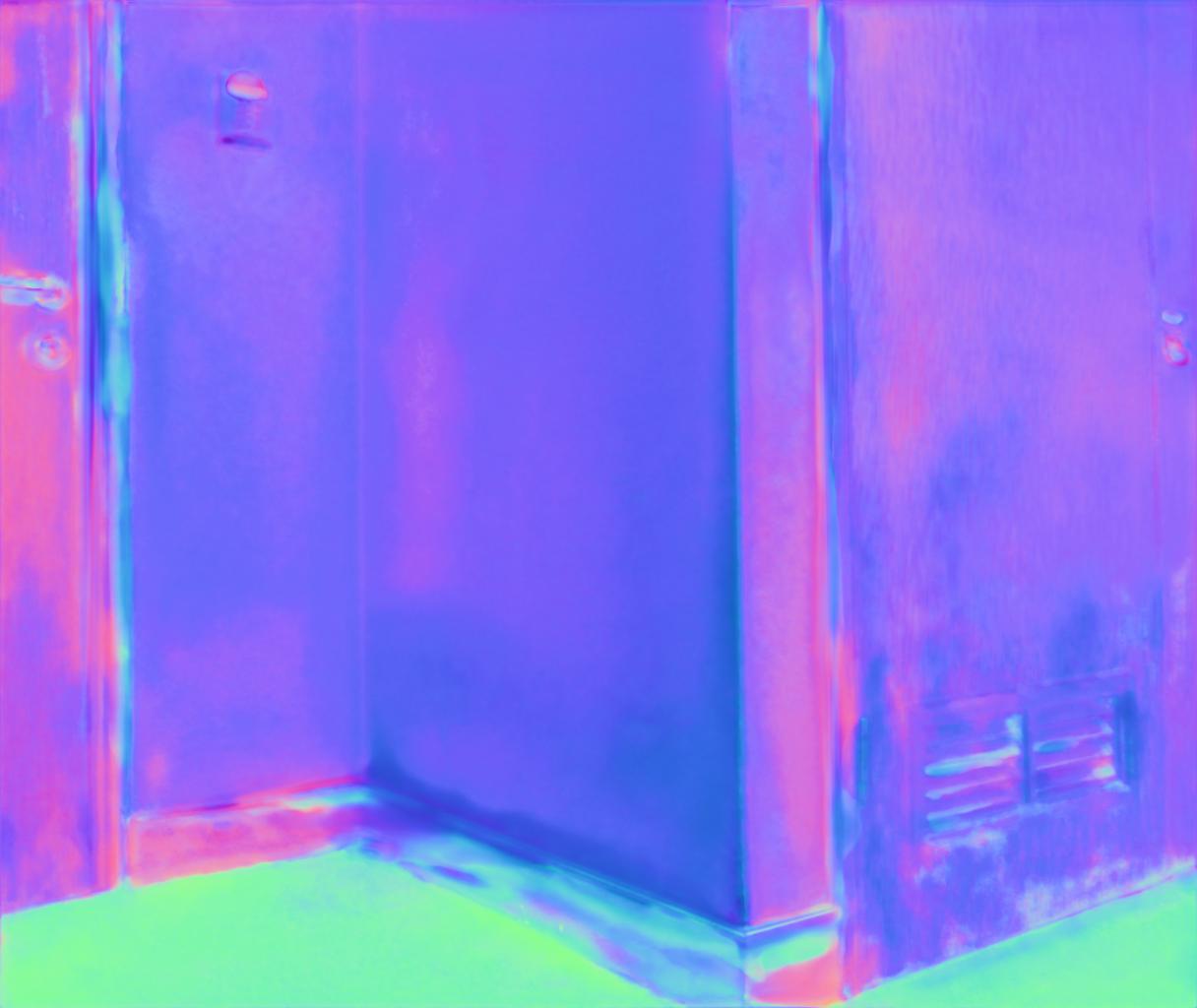}&
\includegraphics[width=0.120\linewidth]{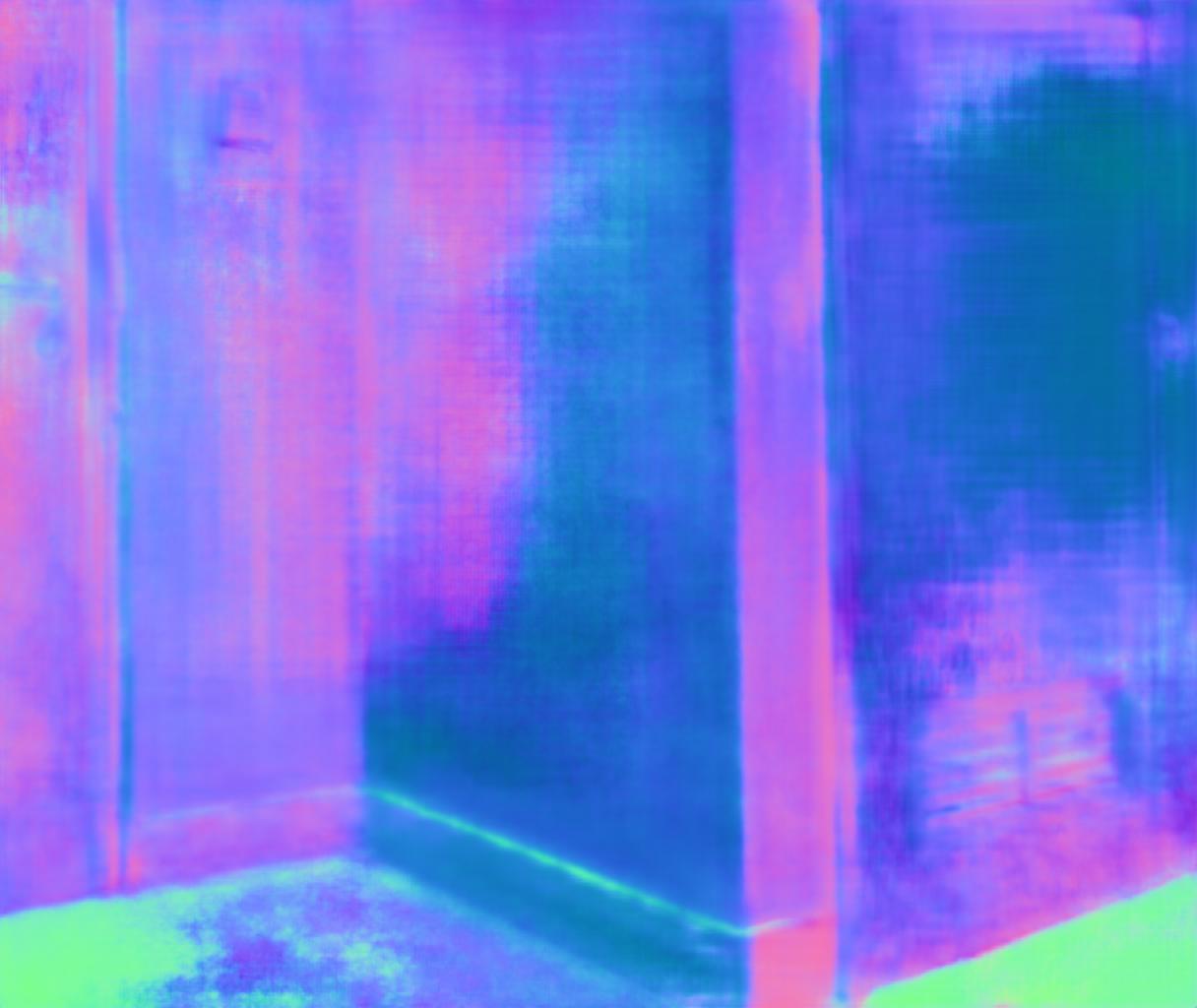}&
\includegraphics[width=0.120\linewidth]{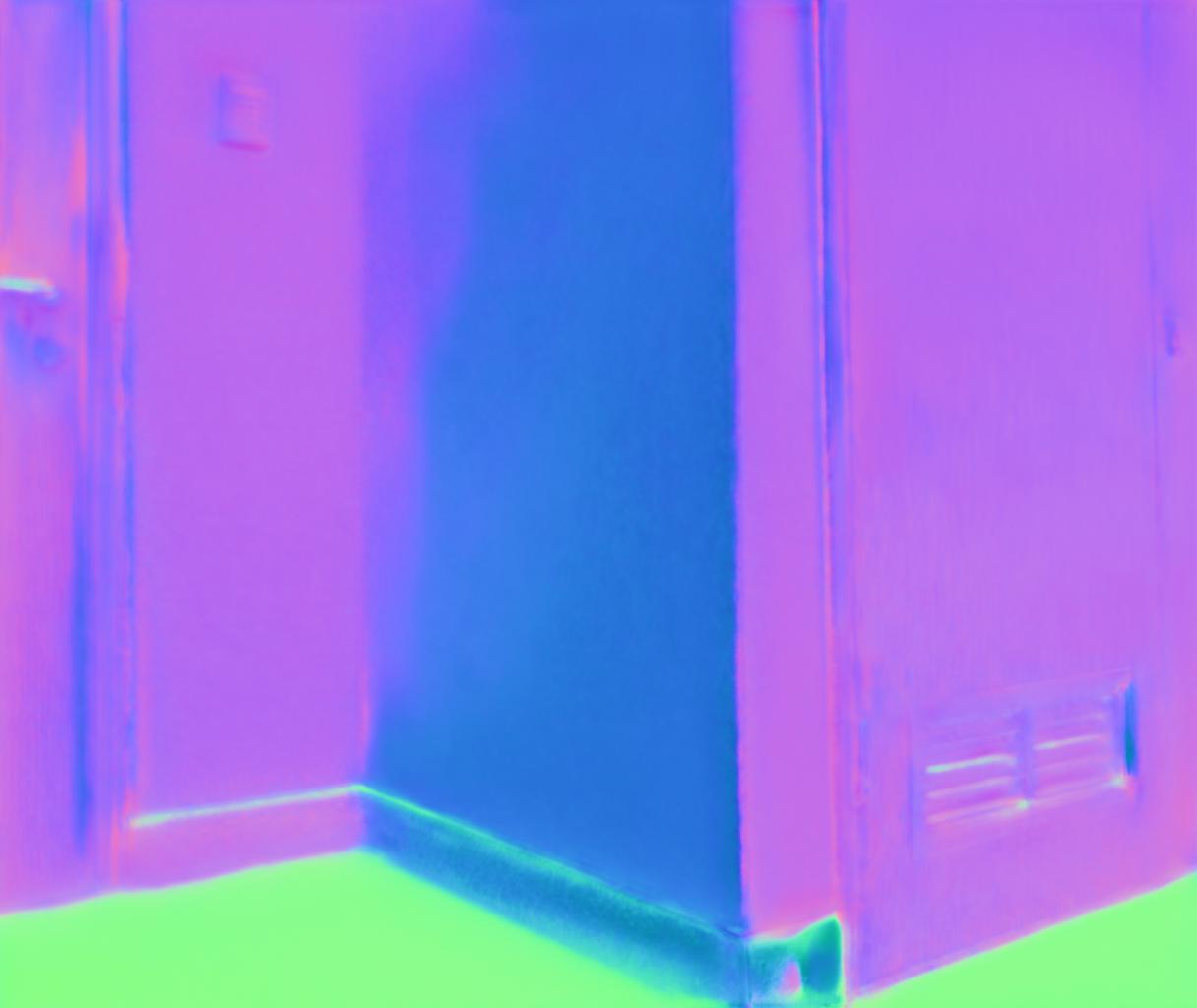}&
\includegraphics[width=0.120\linewidth]{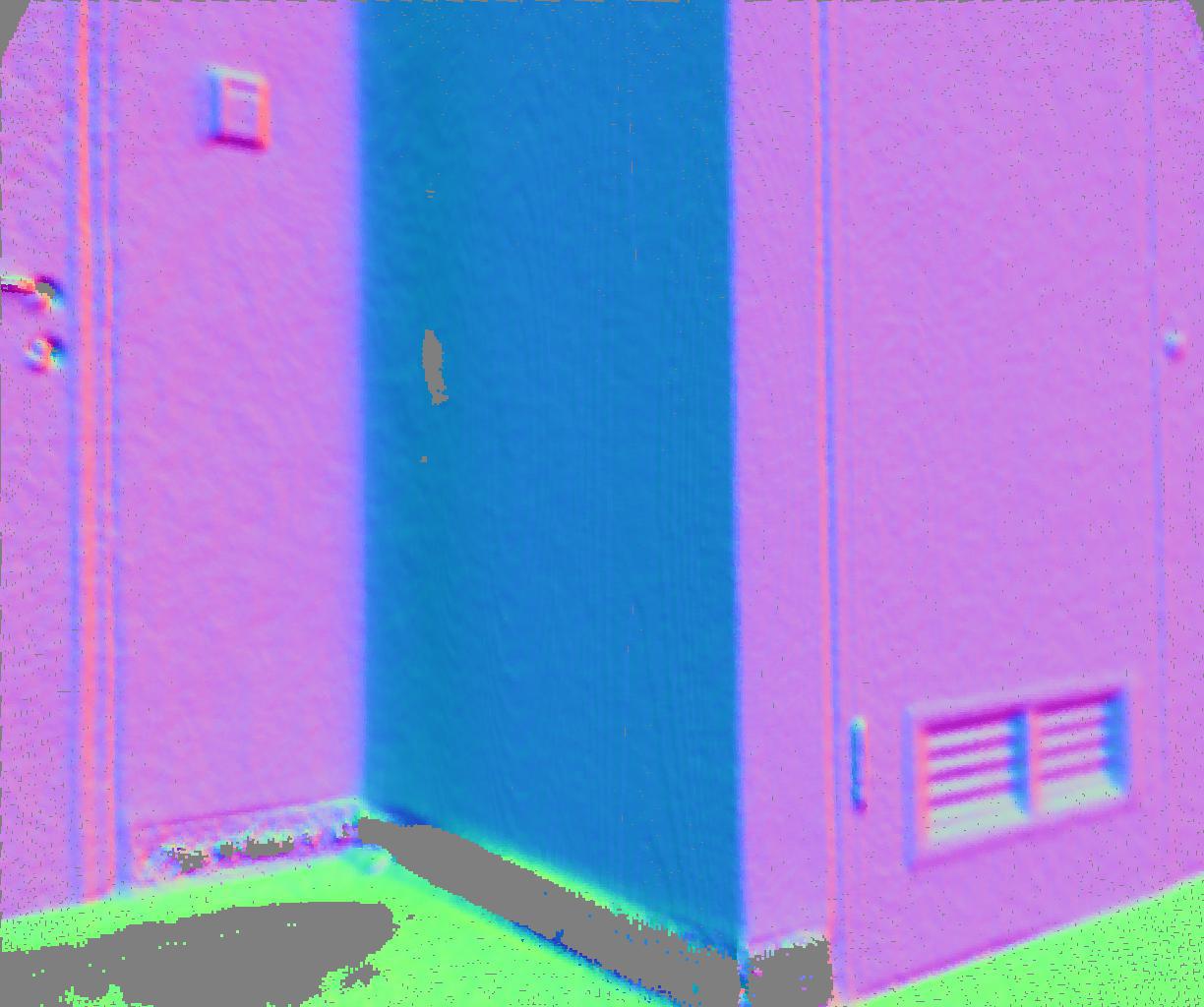}
\\

\small{Input $\mathbf{I}_{un}$}  & \small{Input $\phi$}  & \small{Miyazaki~\cite{miyazaki2003polarization}} & \small{Smith et al.~\cite{smith2019height}}  & \small{DeepSfP~\cite{ba2020deep}}& \small{Kondo et al.~\cite{kondo2020accurate}} &\small{Ours} &Ground truth \\
\end{tabular}

\caption{Qualitative comparison between our approach and other shape from polarization methods baselines~\cite{miyazaki2003polarization,smith2019height,ba2020deep,kondo2020accurate} on the SPW dataset~\cite{ba2020deep}. }
\label{fig:percep_cmp_sfp}
\end{figure*}

\subsection{Comparison to SfP baselines}

Our approach is compared with three physics-based SfP methods (Miyazaki et al.~\cite{miyazaki2003polarization}, Mahmoud et al.~\cite{mahmoud2012direct}, and Smith et al.~\cite{smith2019height}) and two learning-based SfP methods (DeepSfP~\cite{ba2020deep} and Kondo et al.~\cite{kondo2020accurate}). The source code and results of DeepSfP~\cite{ba2020deep} and Konda et al.~\cite{kondo2020accurate} are not available. We reimplement these two approaches and retrain their models on our SPW dataset.

Table~\ref{table:baseline_comparison} presents the quantitative results of all the methods on our SPW dataset. Our approach outperforms all baselines by a large margin on all metrics. 

Fig.~\ref{fig:percep_cmp_sfp} provides a qualitative comparison on images from the SPW dataset. Our estimated surface normal maps are more accurate. Besides, 
Our approach can produce high-quality normals while other methods do not.

Table~\ref{table:cmp_deepsfp} presents the quantitative results on the public DeepSfP dataset~\cite{ba2020deep}, in which we also achieve the best performance. Our approach reduces the mean angular error by $20\%$ compared to the second-best result reported by DeepSfP~\cite{ba2020deep}.

For physics-based SfP~\cite{mahmoud2012direct,miyazaki2003polarization,smith2019height}, since the assumptions of these methods do not hold in the wild, the quantitative accuracy of these approaches is low on the SPW dataset. They cannot obtain satisfying performance on the DeepSfP dataset, either. For example, Mahmoud et al.~\cite{mahmoud2012direct} assume a distant light source, which is not common in the real world (e.g., multiple light sources can exist in a room). As for the learning-based SfP methods~\cite{ba2020deep,kondo2020accurate}, our approach is the best-performing one and we analyze the designs (i.e., polarization representation, viewing encoding and architectures) that contribute to our model in Sec.~\ref{sec:analysis}.

\subsection{Generalization to outdoor scenes}
\begin{figure}
\centering
\begin{tabular}{@{}c@{\hspace{1mm}}c@{\hspace{1mm}}cc@{}}

\includegraphics[width=0.33\linewidth]{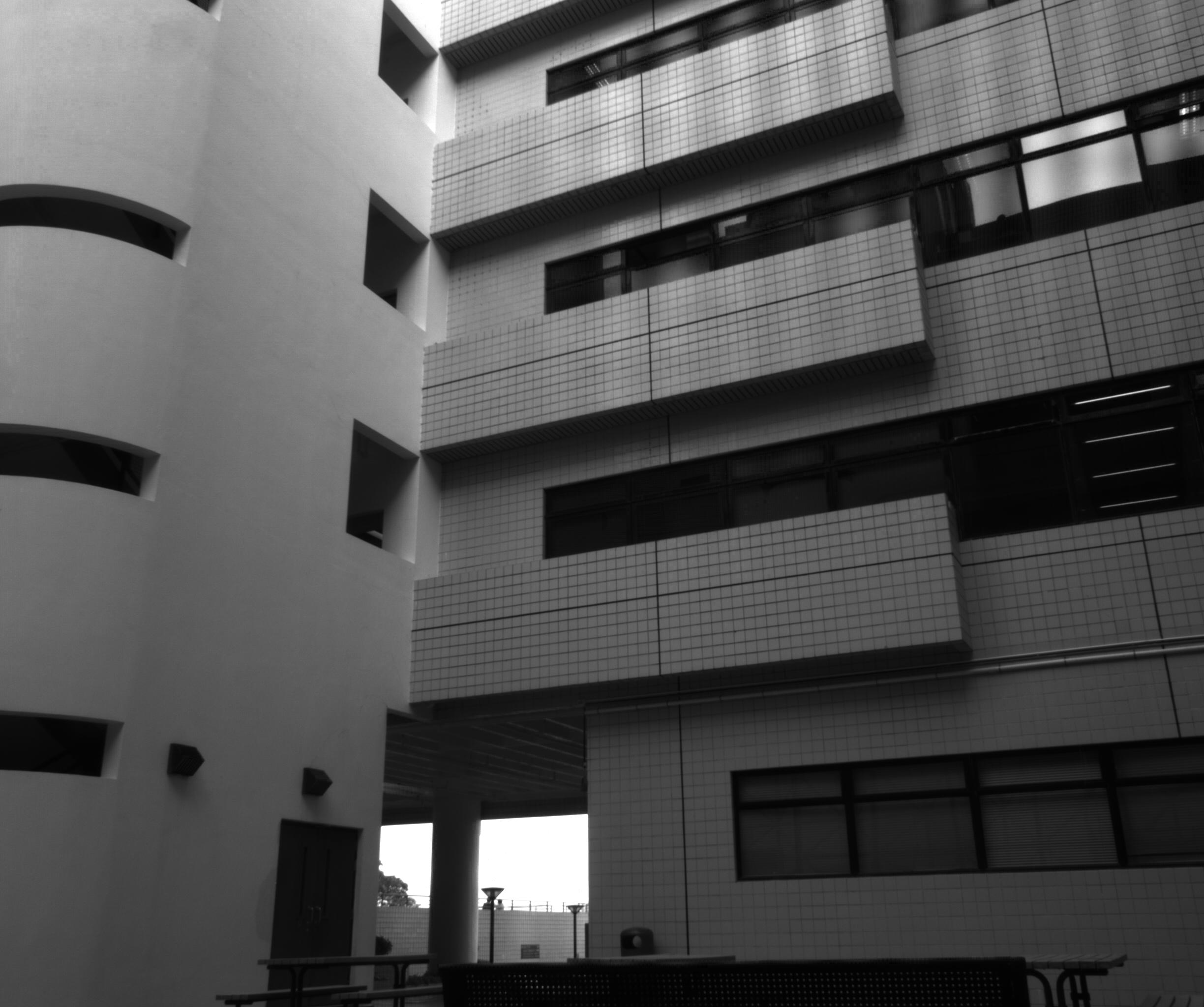}&
\includegraphics[width=0.33\linewidth]{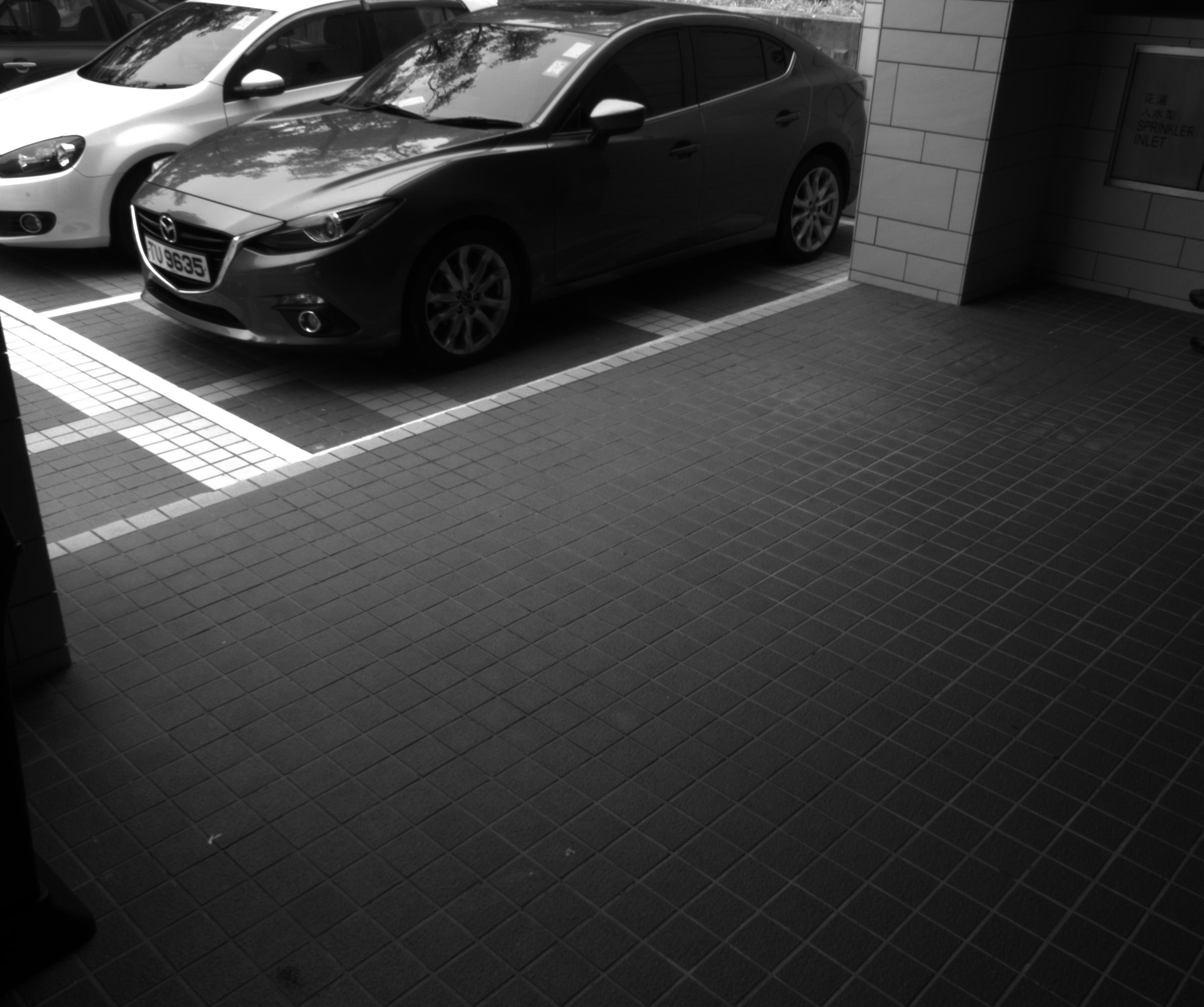}&
\includegraphics[width=0.33\linewidth]{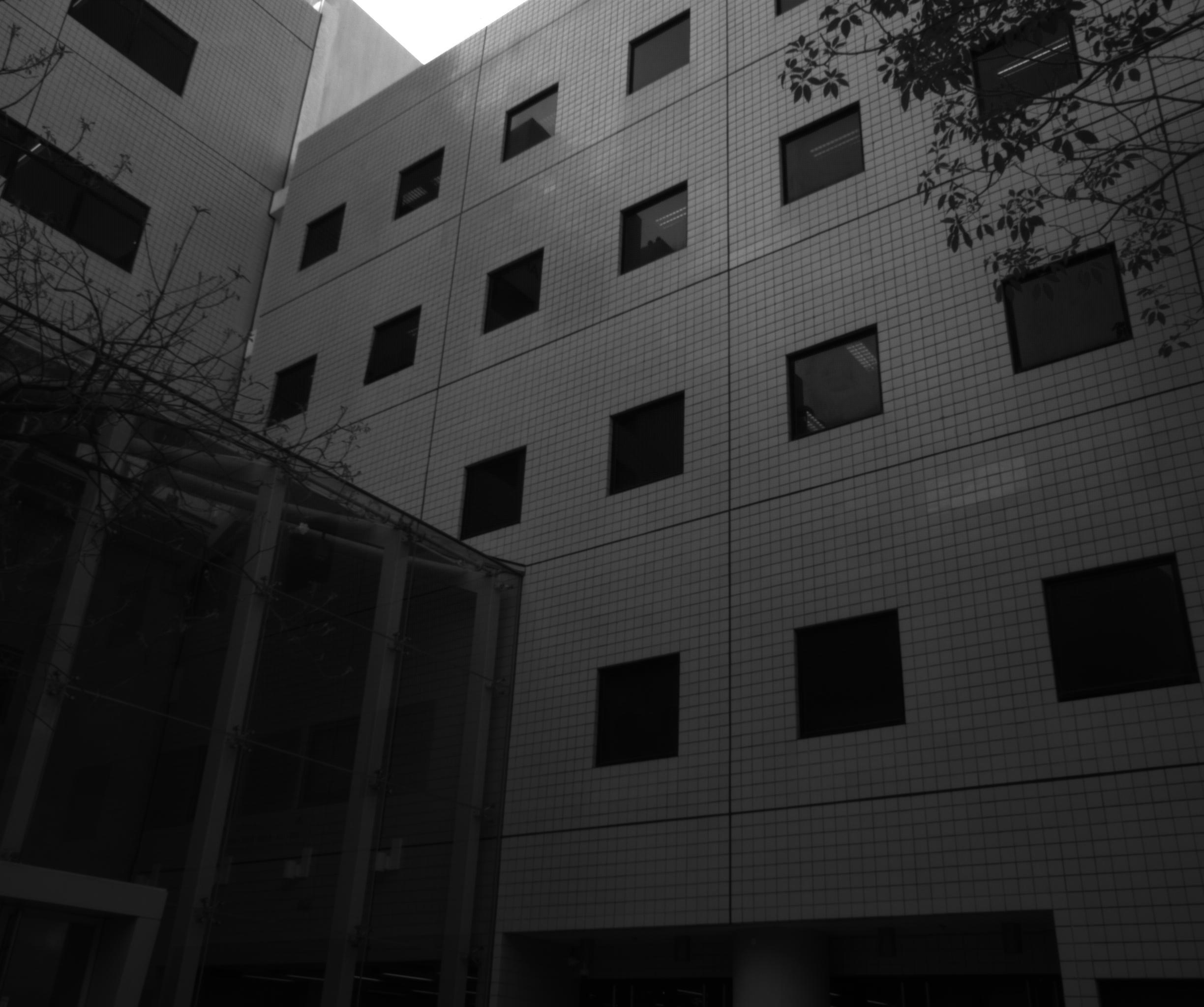}\\

\includegraphics[width=0.33\linewidth]{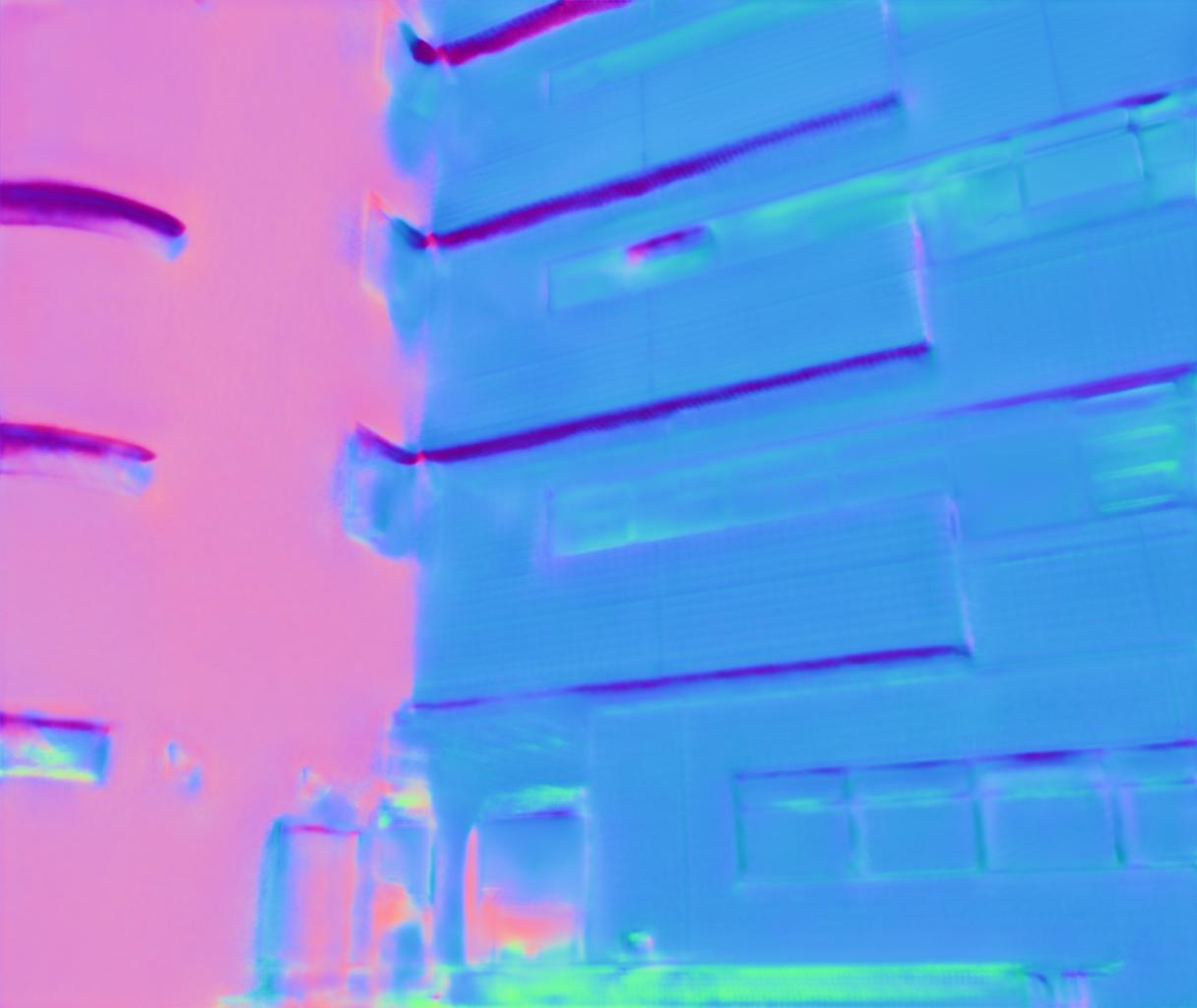}&
\includegraphics[width=0.33\linewidth]{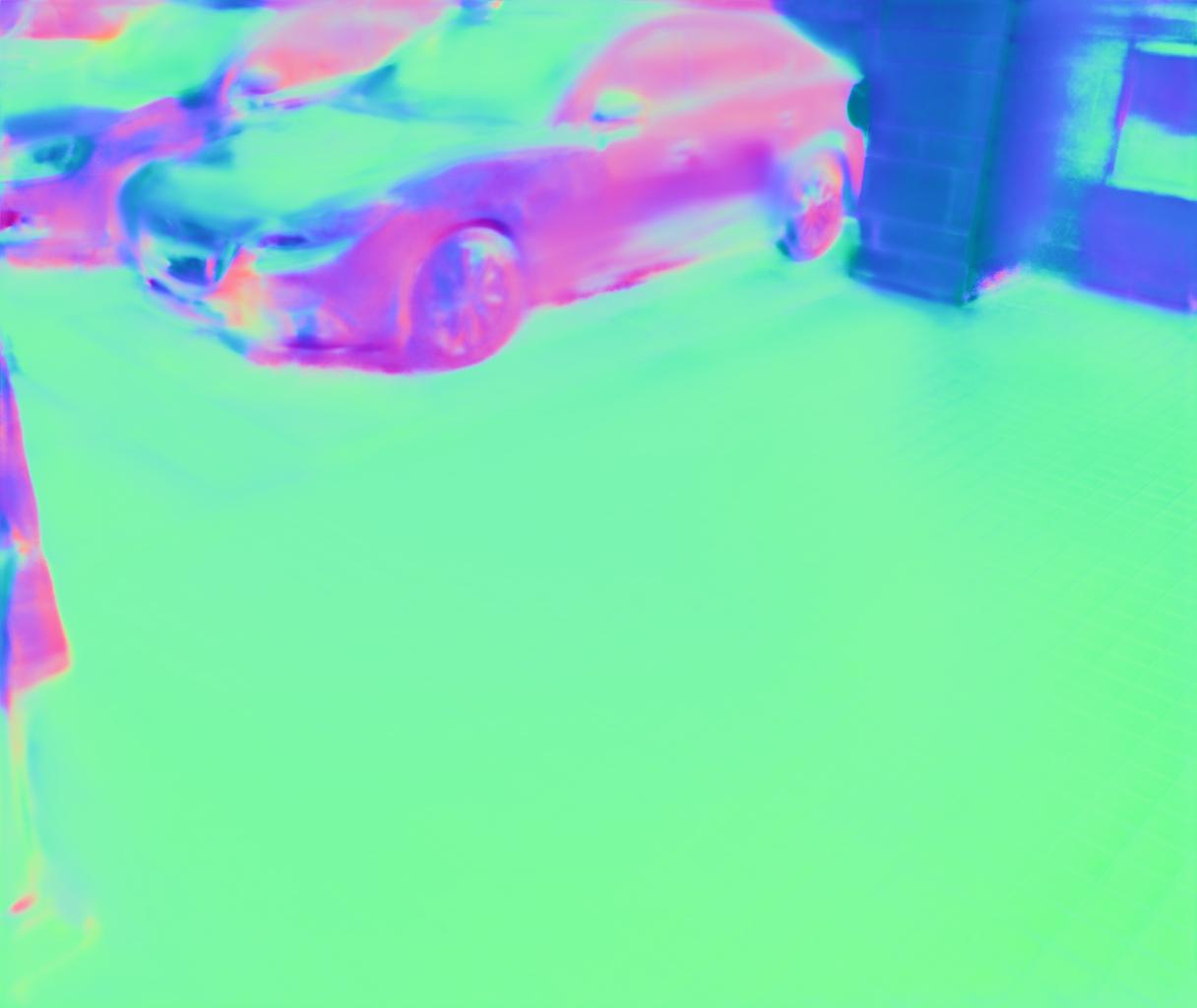}&
\includegraphics[width=0.33\linewidth]{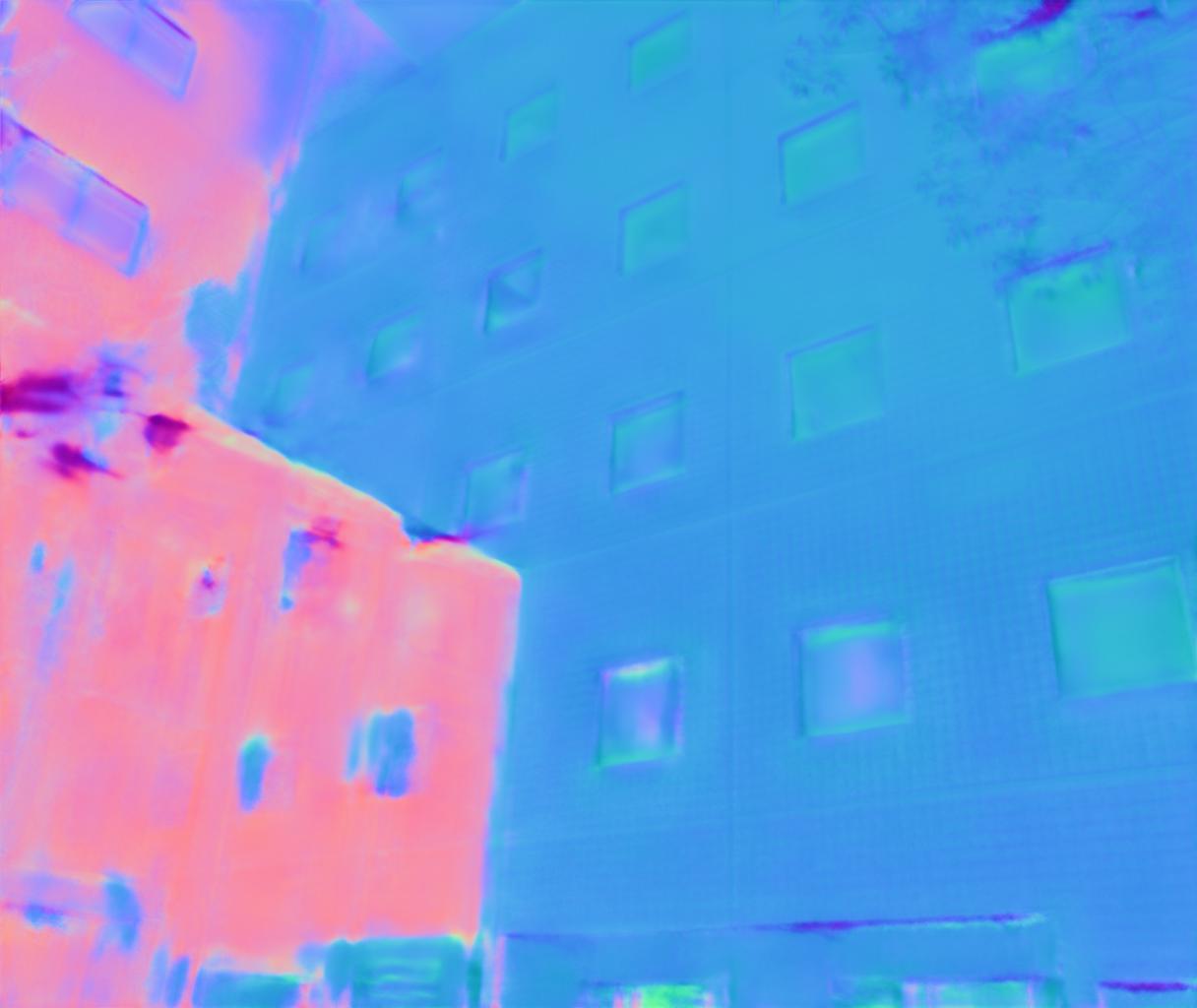}\\

\end{tabular}

\caption{\textbf{Our results on outdoor scenes.} Although our model is trained on near-field content, it appears to successfully generalize to large-scale outdoor scenes. }
\label{fig:distant_scene}
\end{figure}

Although our model is trained on near-field depth estimated by a Kinect camera, it can generalize to outdoor scenes with distances far beyond the Kinect depth range.
This is illustrated qualitatively in Fig.~\ref{fig:distant_scene}. Quantitative results are not provided due to the lack of ground-truth normals in this regime.
This generalization is possible because the relationship between polarized light and surface normals is not affected by distance.
Thus our model that learns to estimate normals from near-field polarization data can generalize to outdoor scenes. %


\subsection{Controlled experiments}
\label{sec:analysis}

\subsubsection{Polarization representation}

\begin{table}
\small
\centering
\renewcommand{\arraystretch}{1.2}
\begin{tabular}{l@{\hspace{5mm}}c@{\hspace{3mm}}c@{\hspace{5mm}}c}
\toprule[1pt]
Polarization & \multicolumn{2}{@{\hspace{-1.5mm}}c}{Mean Angular Error$\downarrow$} & Time(s)\\
  representation $\mathbf{P}$ & SPW   & DeepSfP~\cite{ba2020deep}   \\ 
\midrule
 {Without polarization}& {27.52}  & {19.14} & 0.000 \\ 
 
 \midrule
 {Raw polarization}& {21.77}   & {14.89}   & 0.000 \\ 
 {$\mathbf{P}$ from Kondo et al. ~\cite{kondo2020accurate}}& {18.26}    & {15.44}   & 0.203 \\ 
 {$\mathbf{P}$ from DeepSfP ~\cite{ba2020deep}}& 18.05   & {14.82}  & 1.514 \\ 
 {$\mathbf{P}$} from our approach & \textbf{{17.86}}  & \textbf{14.68} & 0.281 \\ 
\bottomrule[1pt]
\end{tabular}
\caption{\textbf{Controlled experiments for polarization representations on SPW and DeepSfP~\cite{ba2020deep} datasets.} Please check Sec.~\ref{sec:polarprior} for the details of various representations. We test the preprocessing time of a raw image with resolution $1024\times1224$ using a single thread on Intel Xeon Gold CPU with 2.30GHz frequency.}
\label{table:AblationPolar}
\end{table}


The experiments are conducted on both DeepSfP dataset~\cite{ba2020deep} and SPW dataset. We remove the polarization information or replace our proposed polarization representation with other representations as input to our model. Table~\ref{table:AblationPolar} provides the quantitative results of various polarization representations. Utilizing our polarization representation reduces the mean angular error by $9^{~\circ}$. Besides, we obtain the lowest MAE on both datasets and the running time of our representation is much shorter than DeepSfP~\cite{ba2020deep}.

\subsubsection{Viewing encoding} 

\begin{figure}
\centering
\begin{tabular}{@{}c@{}}
\includegraphics[width=0.85\linewidth]{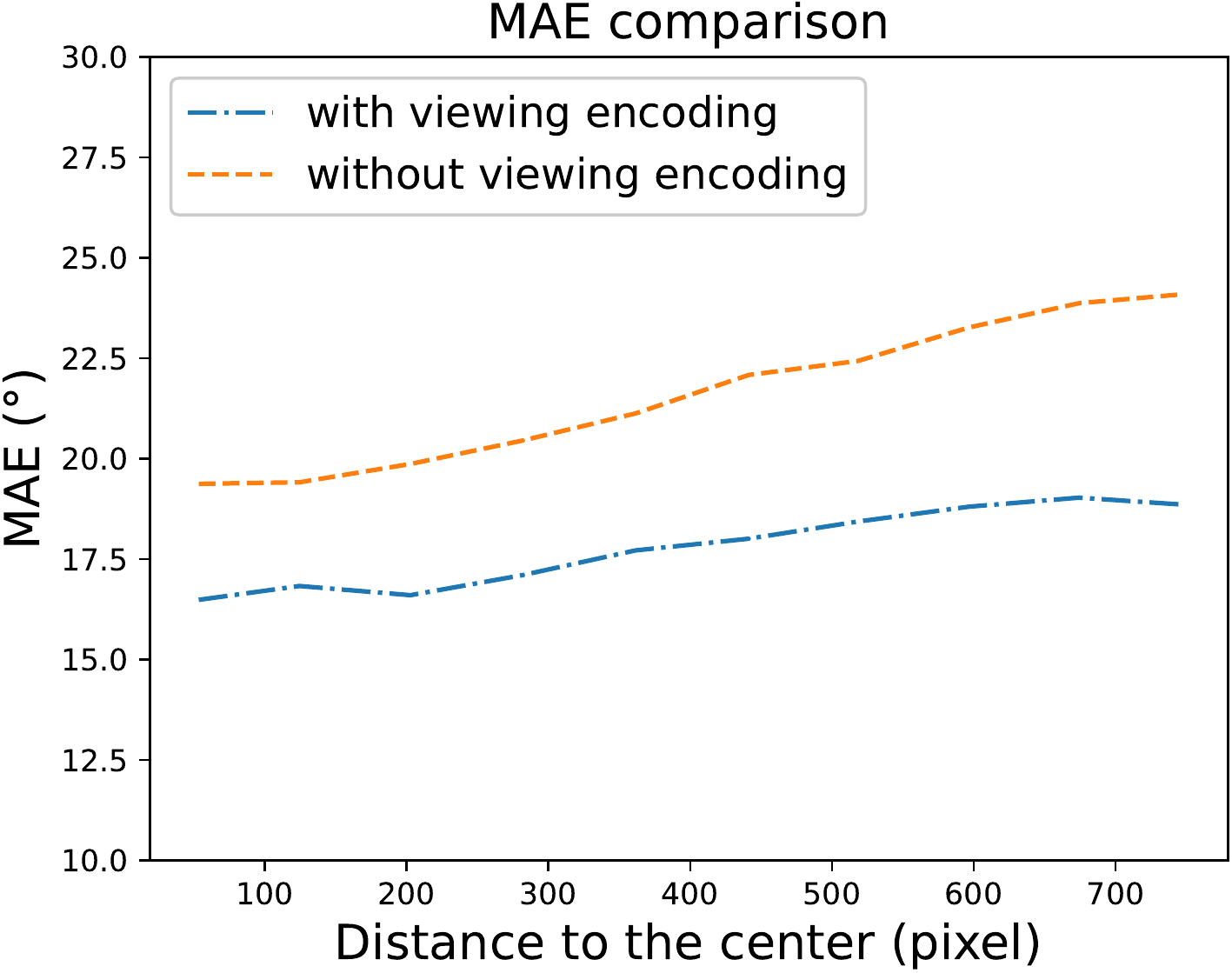}
\end{tabular}
\caption{\textbf{An analysis of the viewing encoding.} We calculate the mean angular error (MAE) for each pixel on the test set. We notice that the improvement brought by viewing encoding increases with the distance to the image center. We believe this is because the impact of non-orthographic projection is more severe in the corners of images. }
\label{fig:pe_cmp}
\end{figure}

Since the DeepSfP dataset is not designed for scene-level SfP, we only conduct the controlled experiments on the SPW dataset. For each model, we remove the viewing encoding from the input or use different types of viewing encoding. In Table~\ref{table:AblationVE}, using viewing encoding improves our model in all the metrics effectively by a large margin. The model that uses raw viewing directions achieves the best performance. The model that uses positional encoding of NeRF~\cite{mildenhall2020nerf} also improves the performance but is not as good as ours. When the images in a dataset are captured with the same intrinsic parameters, using normalized coordinates as viewing encoding also obtains satisfying performance. We further analyze the impact of viewing encoding in Fig.~\ref{fig:Abl-view} and Fig.~\ref{fig:pe_cmp}. 





\begin{table}
\small
\centering
\renewcommand{\arraystretch}{1.2}
\begin{tabular}{@{}l@{\hspace{3mm}}c@{\hspace{1mm}}c@{\hspace{1mm}}c@{\hspace{3mm}}c@{\hspace{1mm}}c@{\hspace{1mm}}c@{}}
\toprule[1pt]
\small{Viewing encoding} & \multicolumn{3}{c}{Angular Error $\downarrow$ } & \multicolumn{3}{c}{Accuracy $\uparrow$}  \\ 
  &  \small{Mean} & \small{Median} & \small{RMSE}& \small{$11.25^{\circ}$}  & \small{$22.5^{\circ}$} & \small{$30.0^{\circ}$} \\ 
\midrule

\small{Ours without $\mathbf{V}$}  & 22.12 & 18.00 & 27.03 & 32.2 & 66.9 & 77.8 \\
 Ours with $\mathbf{V}_p$ & 20.31 & 16.02 & 25.68 & 40.4& 71.0 & 80.5 \\
 Ours with $\mathbf{V}_c$ & 18.44 & 14.62 & 23.46 & 43.7 & {76.1} & 84.8 \\
 Ours with $\mathbf{V}$ &  \textbf{17.86} & \textbf{14.20} & \textbf{22.72} & \textbf{44.6} & \textbf{76.3} & \textbf{85.2} \\ 


\bottomrule[1pt]
\end{tabular}

\caption{\textbf{Controlled experiments for the viewing encoding on the SPW dataset. }Previous learning-based SfP methods~\cite{ba2020deep,kondo2020accurate} do not input any viewing encoding. In addition to our viewing encoding $\mathbf{V}$, we also try to use the positional encoding of NeRF~\cite{mildenhall2020nerf} $\mathbf{V}_p$ and normalized coordinates $\mathbf{V}_c$ as the viewing encoding.}
\label{table:AblationVE}
\end{table}


\subsubsection{Network architectures}

We study different network architectures in this section. In addition to networks of previous SfP methods (DeepSfP~\cite{ba2020deep} and Kondo et al.~\cite{kondo2020accurate}), we also compare with two RGB-based normal estimation methods: TransDepth~\cite{fukao2021polarimetric} and DORN~\cite{fu2018deep}. For all the experiments, we provide the same polarization representation and viewing encoding to these compared network architectures. Table~\ref{table:AblationNet} presents the quantitative results of different architectures. Our architecture obtains the best performance. Besides, removing self-attention or replacing instance normalization with batch normalization leads to performance degradation.

\begin{table}
\small
\centering
\renewcommand{\arraystretch}{1.2}
\begin{tabular}{@{}l@{\hspace{3mm}}c@{\hspace{1mm}}c@{\hspace{1mm}}c@{\hspace{3mm}}c@{\hspace{1mm}}c@{\hspace{1mm}}c@{}}
\toprule[1pt]
\small{Network} & \multicolumn{3}{c}{Angular Error $\downarrow$ } & \multicolumn{3}{c}{Accuracy $\uparrow$}  \\ 
 &  \small{Mean} & \small{Median} & \small{RMSE}& \small{$11.25^{\circ}$}  & \small{$22.5^{\circ}$} & \small{$30.0^{\circ}$} \\ 

\midrule

 {Kondo et al.$^\dag$~\cite{kondo2020accurate}}&26.43 & 22.69 & 31.80 & 23.8 & 54.1 & 67.6  \\ 
 {DeepSfP$^\dag$~\cite{ba2020deep}}& 24.97 & 20.83 & 30.13 & 25.6 & 58.4  &70.9  \\ 
  {U-Net~\cite{Ronneberger2015Unet}}& 26.35 & 22.45& 31.97 & 25.4 & 54.5 & 67.6 \\ 
    {DORN~\cite{fu2018deep}}& 20.16 & 15.60 & 25.47 & 39.8 & 71.3 & 81.1 \\ 
{TransDepth~\cite{yang2021transformers}}& 22.05 & 17.46 & 27.77 & 33.0 & 66.6 & 77.9  \\ 
  \hline
  {Ours without IN}& 20.74 &16.63 & 25.98 & 38.5 & 69.3 & 79.0 \\ 
  {Ours without SA}&   21.08 & 16.54 & 26.62 &36.1 & 68.5 & 79.3 \\
 Ours &  \textbf{17.86} & \textbf{14.20} & \textbf{22.72} & \textbf{44.6} & \textbf{76.3} & \textbf{85.2} \\ 


\bottomrule[1pt]
\end{tabular}

\caption{\textbf{Controlled experiments for network architectures on the SPW dataset.} 
We retrain Kondo et al.~\cite{kondo2020accurate}, DeepSfP~\cite{ba2020deep} and other networks with the same representation as ours (e.g. viewing encoding and our novel polarization representation) for fair comparison. 
SA: self-attention. IN: instance normalization. $\dag$: our implementation.}
\label{table:AblationNet}
\end{table}



\section{Conclusion}

We present the first approach dedicated to scene-level surface normal estimation from a single polarization image in the wild. The accuracy of our model is demonstrated on SPW, the first scene-level dataset for real-world SfP. By introducing the viewing encoding, a self-attention module and a novel polarization representation to SfP, our model substantially outperforms prior work on both SPW and the object-level DeepSfP dataset. In addition, our model can generalize from near-field scenes (used during training) to far-field outdoor scenes. This is possible because the polarization sensor is based on passive sensing, so our trained model is expected to generalize to distant scenes. 
We hope our work including the proposed SPW dataset and our technical designs can contribute to high-quality normal estimation, especially shape from polarization.

\paragraph{Limitations.} One of the limitations of our work is the lack of quantitative evaluation in outdoor scenes. Note that the quantitative experiments in outdoor scenes will require long-range high-resolution depth and normal estimation with high-end depth sensors.
{\small
\bibliographystyle{ieee_fullname}
\bibliography{egbib}
}

\end{document}


\title{Shape from Polarization for Complex Scenes in the Wild\\
\textit{Supplementary Material}}

\author{First Author\\
Institution1\\
Institution1 address\\
{\tt\small firstauthor@i1.org}
\and
Second Author\\
Institution2\\
First line of institution2 address\\
{\tt\small secondauthor@i2.org}
}
\maketitle


\section*{Summary of the Supplementary Material}
\noindent This supplementary document is organized as follows: 
\begin{itemize}
  \item Section~\ref{sec:Additional} provides addition results.

 \item Section~\ref{sec:details} presents the details of our implementation.

 \item Section~\ref{sec:polarization} introduces the background of shape from polarization.

\end{itemize}

\section{Additional Results}
\label{sec:Additional}

\subsection{Comparison to non-polarization baselines}
We choose the latest RGB-based normal estimation method~\cite{yang2021transformers} for comparison. According to the evaluation results of Yang et al.~\cite{yang2021transformers}, their model obtains the best score on the NYU dataset when models under the same training setting. We retrain their model on the SPW dataset without using the polarization information. 

Table \ref{table:NonPolarization} shows the quantitative results. Our results are significantly better than the results of Yang et al.~\cite{yang2021transformers}. Note that the results are for reference as the experimental setting is unfair: we use the extra polarization information; they use the pretrained weights on ImageNet~\cite{deng2009imagenet}.
\begin{table}[h]
\small
\centering
\renewcommand{\arraystretch}{1.2}
\begin{tabular}{@{}l@{\hspace{3mm}}l@{\hspace{1mm}}c@{\hspace{1mm}}c@{\hspace{3mm}}c@{\hspace{1mm}}c@{\hspace{1mm}}c@{\hspace{1mm}}c@{}}
\toprule[1pt]
\small{Method} & \multicolumn{3}{c}{Angular Error $\downarrow$ } & \multicolumn{3}{c}{Accuracy $\uparrow$}  \\ 
 & \small{Mean} & \small{Median}  & \small{RMSE}& \small{$11.25^{\circ}$} & \small{$22.5^{\circ}$} & \small{$30.0^{\circ}$} \\
 
\midrule[0.6pt]

  


\small{TransDepth$^\dag$} & 
25.96 & 21.71 & 31.77 & 26.9 & 56.7 & 68.3  \\

 \small{Ours}   &         
 \textbf{17.86} & \textbf{14.20} & \textbf{22.72} & \textbf{44.6} & \textbf{76.3} & \textbf{85.2} \\ 
\bottomrule[1pt]
\end{tabular}

\caption{\textbf{Quantitative evaluation on the SPW dataset.} Our approach outperforms rgb-based method TransDepth~\cite{yang2021transformers} by a large margin on all evaluation metrics. $\dag$: we retrain the model on the unpolarized intensity images in SPW dataset. }
\label{table:NonPolarization}
\end{table}

\subsection{Ablation experiments}
We report the quantitative results with various number of self-attention blocks in Table~\ref{table:AttentionBlockNumber}. On the SPW dataset, using 8 blocks obtains almost the best performance, while the gain from even more blocks is marginal. An interesting phenomenon is that using only 1 block can also improve the performance substantially.

\subsection{Visualization on Deepsfp Dataset}
\begin{figure}
\centering
\begin{tabular}{@{}c@{\hspace{1mm}}c@{\hspace{1mm}}c@{\hspace{1mm}}c@{}}

\includegraphics[width=0.32\linewidth]{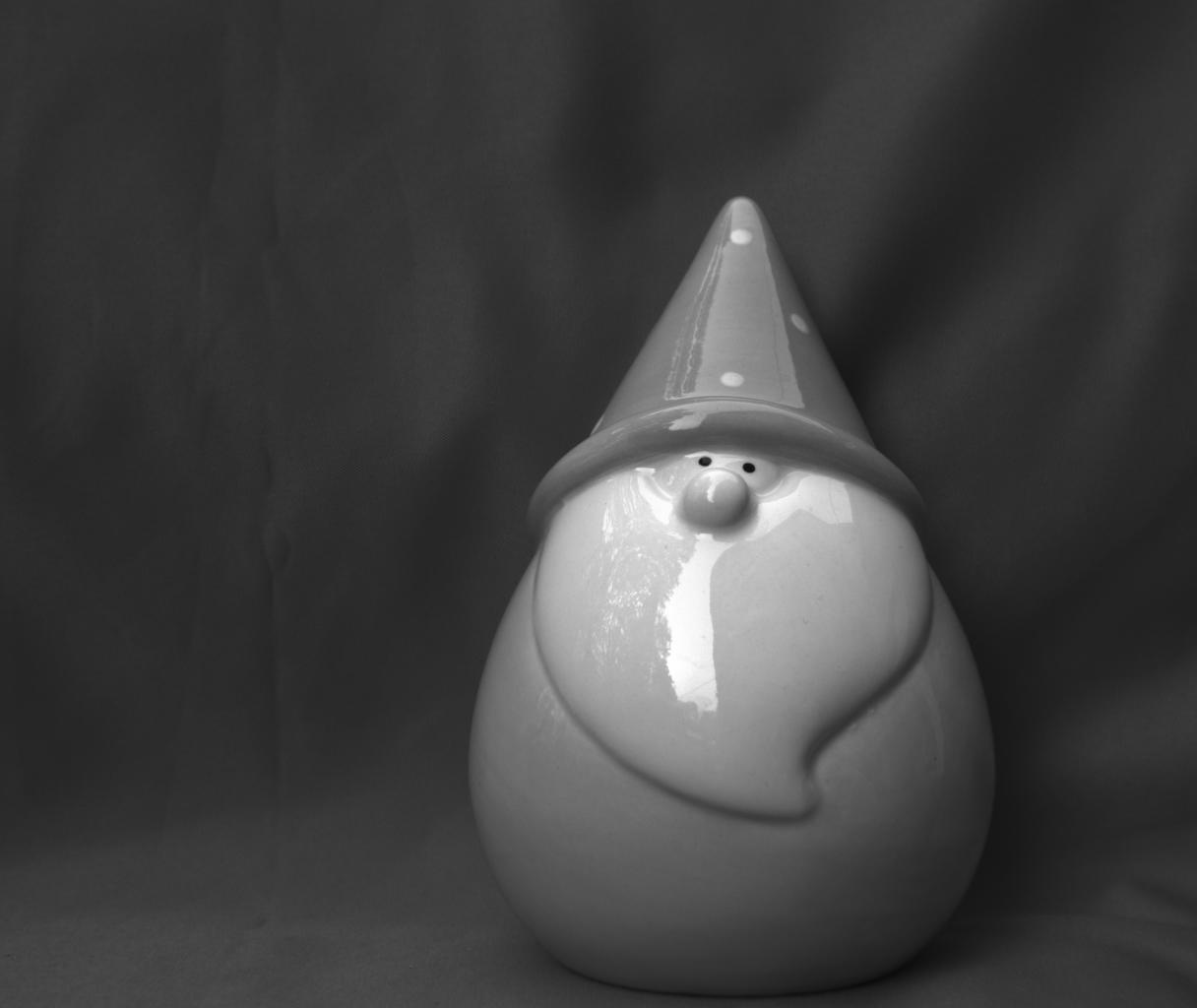}&
\includegraphics[width=0.32\linewidth]{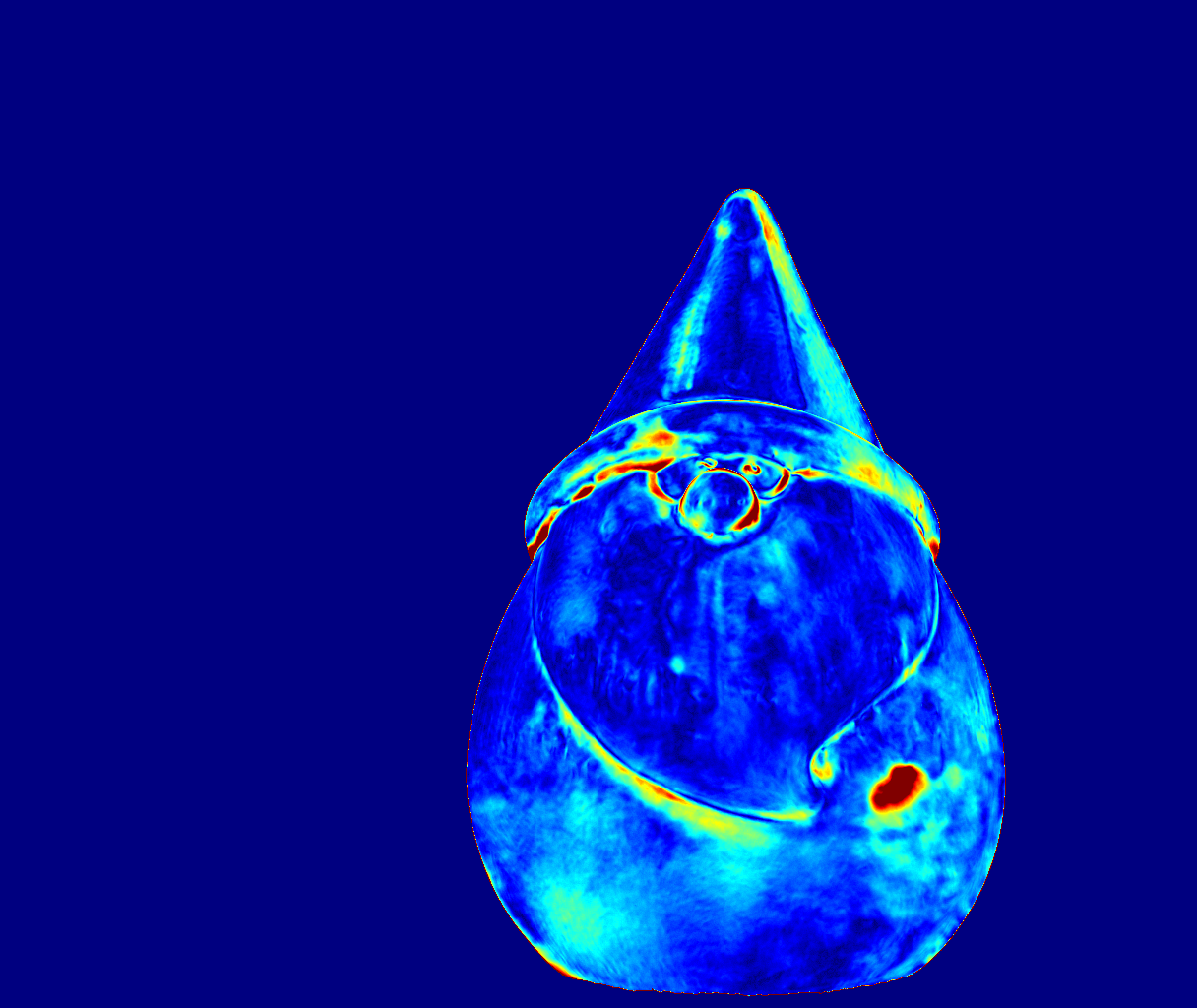}&
\includegraphics[width=0.32\linewidth]{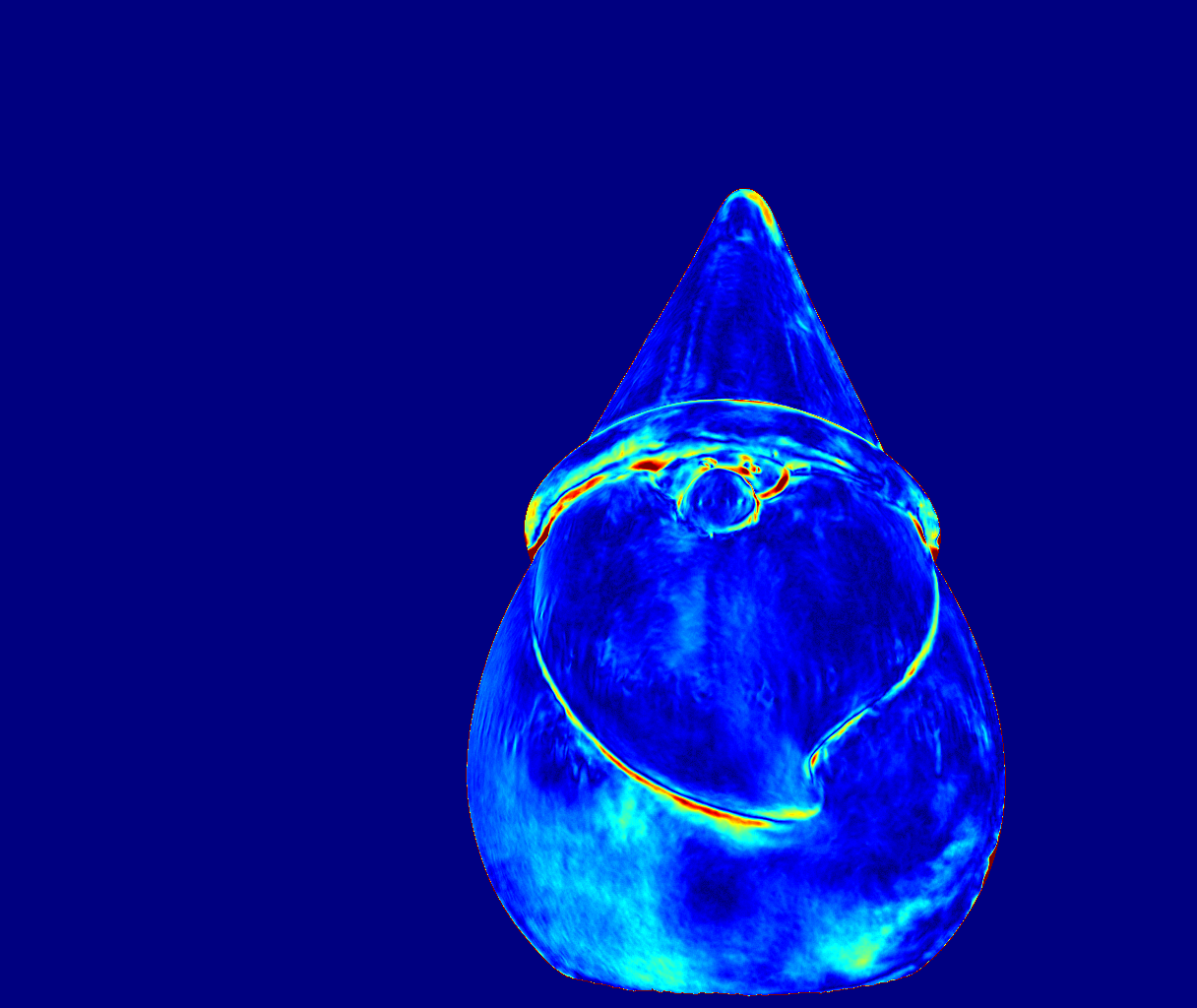}\\

\includegraphics[width=0.32\linewidth]{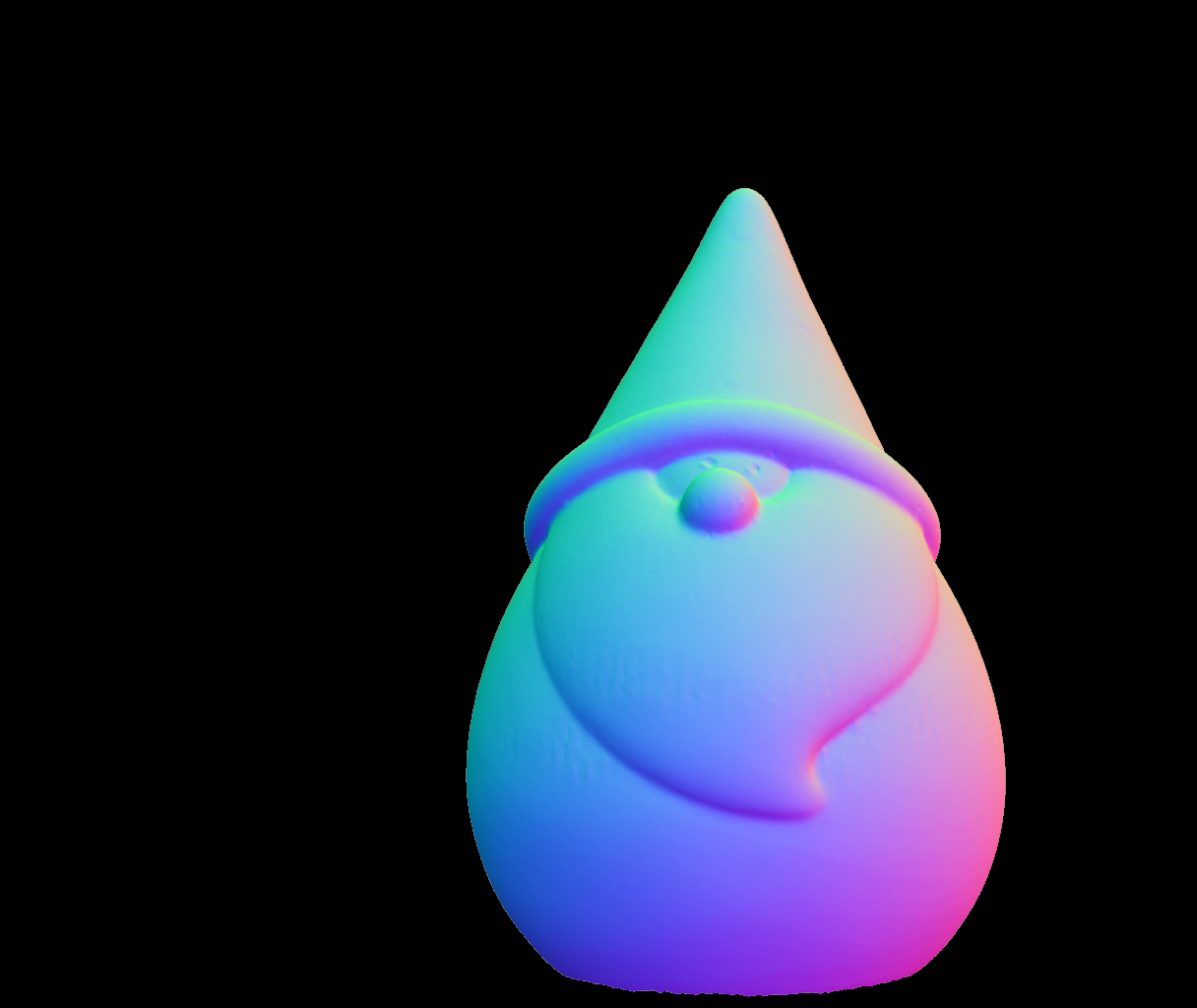}&
\includegraphics[width=0.32\linewidth]{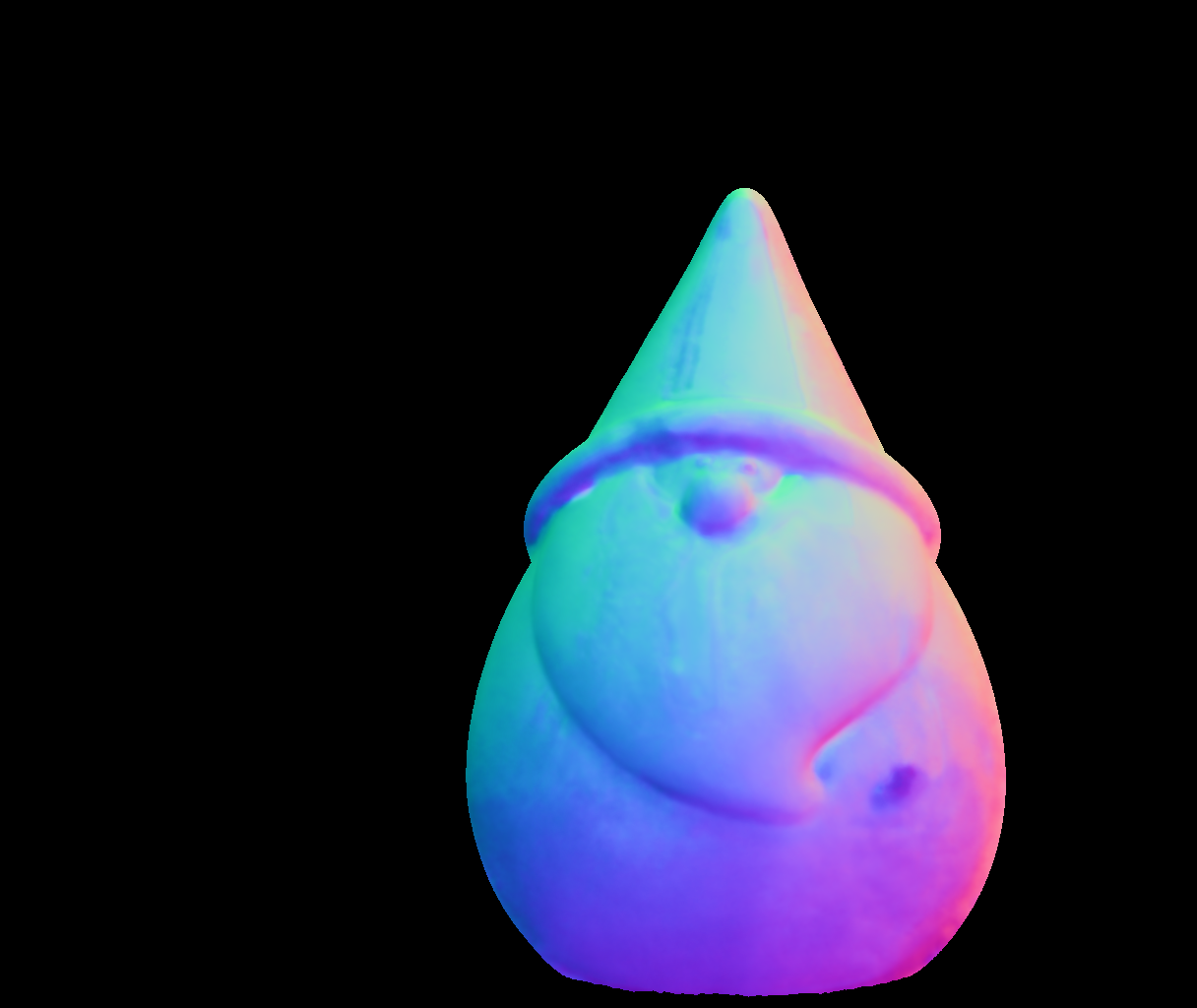}&
\includegraphics[width=0.32\linewidth]{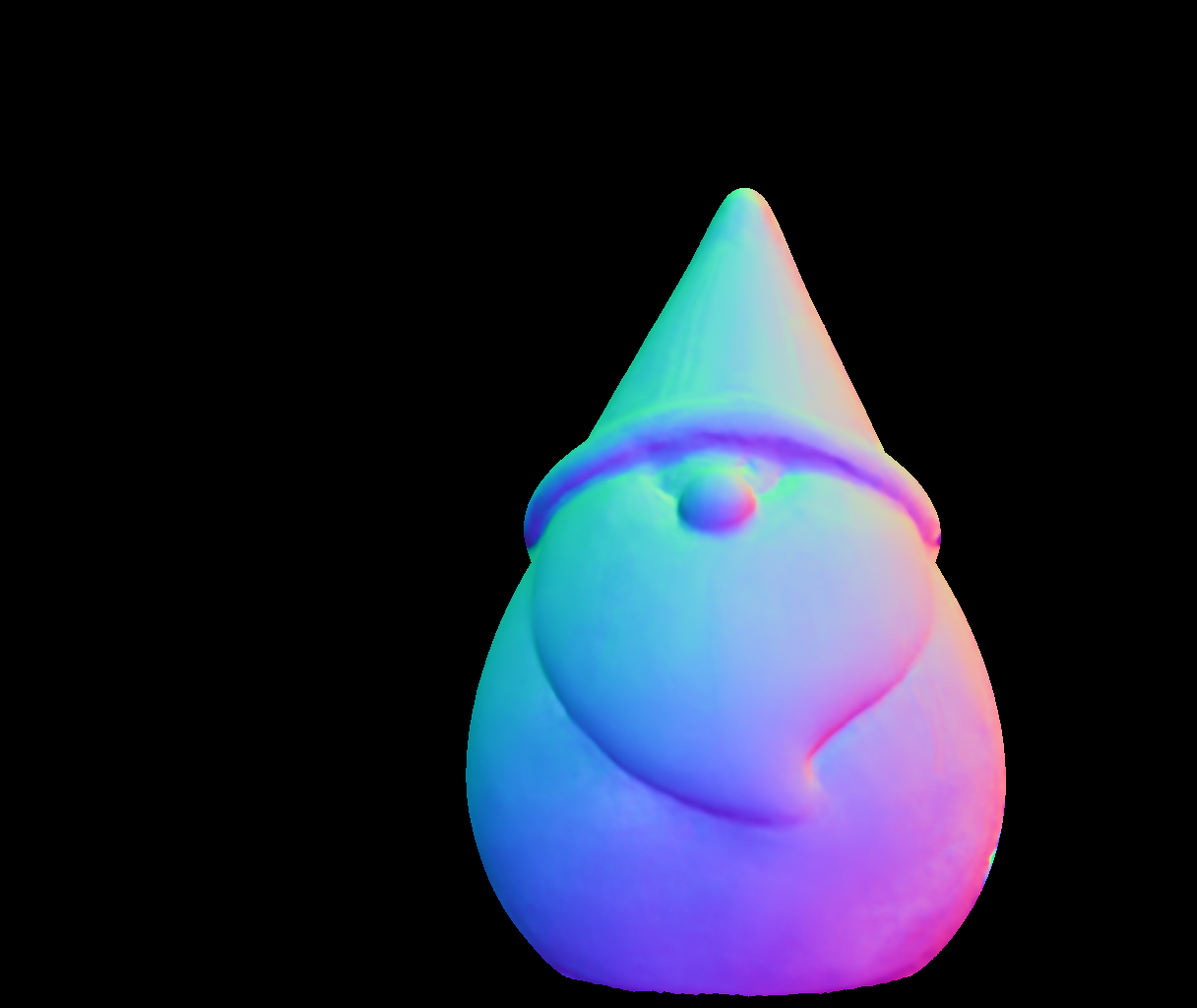}\\

\small{Input $\mathbf{I}_{un}$ and GT}  & DeepSfP\cite{ba2020deep}& Ours
\end{tabular}

\caption{Visual comparison results of estimated normals. We show error maps of DeepSfP\cite{ba2020deep} and ours.}
\label{fig:vis_deepsfp}
\end{figure}

In Fig.~\ref{fig:vis_deepsfp}, we present more perceptual results of our method and DeepSfP\cite{ba2020deep} baseline on the DeepSfP\cite{ba2020deep} dataset. Our result is more accurate than the baseline since our hybrid architecture and polarization representation can handle the diffuse/specular-ambiguity better.

\begin{table}
\small
\centering
\begin{tabular}{@{}c@{\hspace{3mm}}c@{\hspace{1mm}}c@{\hspace{1mm}}c@{\hspace{3mm}}c@{\hspace{1mm}}c@{\hspace{1mm}}c@{}}
\toprule[1pt]
\small{Blocks} & \multicolumn{3}{c}{Angular Error $\downarrow$ } & \multicolumn{3}{c}{Accuracy $\uparrow$}  \\
 &  \small{Mean} & \small{Median} & \small{RMSE}& \small{$11.25^{\circ}$}  & \small{$22.5^{\circ}$} & \small{$30.0^{\circ}$} \\ 
 \midrule
{0}&   21.08 & 16.54 & 26.62 &36.1 & 68.5 & 79.3 \\
{1}& 19.54 & 15.44 & 24.71 & 40.6 & 72.6 & 82.1  \\ 
{2}& 19.21 & 15.35 & 24.17 & 41.4 & 73.6 & 82.7  \\
{4}&18.27 & 14.32 & 23.20 & 44.7 & 74.8 & 84.3  \\
{8}& 17.76 & 13.92 & 22.80 & {45.8} & \textbf{77.5} & \textbf{85.7} \\ 
{12}& \textbf{17.67} & \textbf{13.60} & 22.75 & \textbf{46.4} & 77.2 & 85.2  \\
\bottomrule[1pt]
\end{tabular}
\caption{\textbf{Ablation experiments for the number of self-attention blocks on the SPW dataset.} We choose 8 blocks in our model according to the quantitative results.}
\label{table:AttentionBlockNumber}
\end{table}

\section{Implementation Details}
\label{sec:details}
\subsection{Network architecture}
Table \ref{table:Architectures} shows details of our architecture specification. ``3×3, 64," denotes a 2d convolution operation of kernel size 3, output channel 64. ``BN, ReLU'' denotes batch normalization~\cite{ioffe2015batchnorm} and ReLU activation~\cite{nair2010rectified}. ``IN" denotes instance normalization~\cite{ulyanov2016instancenorm}, while ``LN" denotes layer normalization~\cite{ba2016layernorm}. In multi-head self-attention blocks, ``dim 512 (head 8) MHA" indicates a 8-heads attention block each with head dimention 64. ``2048-d MLP" denotes a MLP with a hidden layer of 2048 dimensions. 

\begin{table*}[t]
\small
\centering
\renewcommand{\arraystretch}{1.2}
\begin{tabular}{c|c|c}
\hline
stage & building block & output size\\
\hline

\hline
$\begin{array}{l}
    \text{input convolution} \\
\end{array}$ & {$\left[\begin{array}{l}
3 \times 3, 64 \\
\text{ BN, ReLU}\end{array}\right] \times 2$ }  & ${H} \times {W} \times 64 $ \\ 






\hline
$\begin{array}{l}
    \text{downsampling convolution1} \\
\end{array}$ & {$\left[\begin{array}{l}
3 \times 3, 128 \\
\text{ IN, ReLU}\end{array}\right] \times 2$ }  & $\frac{H}{2} \times \frac{W}{2} \times 128$  \\ 

\hline
$\begin{array}{l}
    \text{downsampling convolution2} \\
\end{array}$ & {$\left[\begin{array}{l}
3 \times 3, 256 \\
\text{ IN, ReLU}\end{array}\right] \times 2$ }  & $\frac{H}{4} \times \frac{W}{4} \times 256$  \\ 

\hline
$\begin{array}{l}
    \text{downsampling convolution3} \\
\end{array}$ & {$\left[\begin{array}{l}
3 \times 3, 512 \\
\text{ IN, ReLU}\end{array}\right] \times 2$ }  & $\frac{H}{8} \times \frac{W}{8} \times 512$  \\ 

\hline
$\begin{array}{l}
    \text{downsampling convolution4} \\
\end{array}$ & {$\left[\begin{array}{l}
3 \times 3, 512 \\
\text{ IN, ReLU}\end{array}\right] \times 2$ }  & $\frac{H}{16} \times \frac{W}{16} \times 512$  \\

\hline
\small{multi-head attention} & {$\left[\begin{array}{l} \text{LN}, \text{dim 512 (head 8) MHA} \\
\text{LN, 2048-d MLP}\end{array}\right] \times 8$}   & $ \frac{H}{16} \times \frac{W}{16} \times 512$ \\
\hline
$\begin{array}{l}
    \text{skip-connection and upsampling convolution1} \\
\end{array}$ & {$\left[\begin{array}{l}
3 \times 3, 512 \\ 
\text{ BN, ReLU} \\
3 \times 3, 256 \\
\text{ BN, ReLU}\end{array}\right] \times 1 $}  & $\frac{H}{8} \times \frac{W}{8} \times 256$  \\ 

\hline
$\begin{array}{l}
    \text{skip-connection and upsampling convolution2} \\
\end{array}$ & {$\left[\begin{array}{l}
3 \times 3, 256 \\ 
\text{ BN, ReLU} \\
3 \times 3, 128 \\
\text{ BN, ReLU}\end{array}\right] \times 1 $}  & $\frac{H}{4} \times \frac{W}{4} \times 128$  \\ 

\hline
$\begin{array}{l}
    \text{skip-connection and upsampling convolution3} \\
\end{array}$ & {$\left[\begin{array}{l}
3 \times 3, 128 \\ 
\text{ BN, ReLU} \\
3 \times 3, 64 \\
\text{ BN, ReLU}\end{array}\right] \times 1 $}  & $\frac{H}{2} \times \frac{W}{2} \times 64$  \\ 

\hline
$\begin{array}{l}
    \text{skip-connection and upsampling convolution4} \\
\end{array}$ & {$\left[\begin{array}{l}
3 \times 3, 64 \\ 
\text{ BN, ReLU}\end{array}\right] \times 2 $}  & ${H} \times {W} \times 64$  \\





\hline
\small{output convolution} & $1 \times1, 3$ & $H \times W \times 3$  \\ 
\hline


\hline
\end{tabular}

\caption{\textbf{Architectures of our hybrid model.} Building blocks are shown in brackets, with the numbers of blocks stacked. Downsampling is performed at the beginning of downsampling convolution layer using max pooling of stride 2. $2\times$ bilinear upsampling and skip-connection with the encoder features are conducted at the beginning of upsampling convolution layer.}
\label{table:Architectures}
\end{table*}

\section{Shape from Polarization}
\label{sec:polarization}
In this section, we provide a detailed introduction to polarization. Table~\ref{table:Symbol} shows all the used symbols and notations.

\begin{table}[t]
\small
\centering
\renewcommand{\arraystretch}{1.2}
\begin{tabular}{l@{\hspace{2mm}}l@{\hspace{2mm}}}
\toprule[1pt]
 Symbol &  Description  \\ 
\midrule
$\mathbf{n}$ & Surface normal\\ 
$\mathbf{P_i}$ & Incidence plane. \\
$\mathbf{n_i}$ & Normal of the incidence plane\\ 
$\mathbf{n_c}$ & Normal of the camera plane\\ 

$\mathbf{v}$ & Viewing direction \\ 
${\alpha}$ & Azimuth angle \\ 
${\theta}$ & Zenith angle \\ 
${\theta_\mathbf{v}}$ & Viewing angle, the angle between $\mathbf{n}$ and $\mathbf{v}$ \\ 
${\eta}$ & Refractive index \\

\hline
$\rho$ & Degree of polarization \\ 
$\phi$ & Angle of polarization \\ 
$I_{un}$ & Unpolarized intensity \\ 
$I$ & Intensity of incident light \\ 
$\mathbf{d}$ & 3D polarization direction \\
$\phi_{pol}$ & Polarizer angle \\ 
$\mathbf{\Phi}$ & The vector representation of $\phi$ in the camera plane. \\

\bottomrule[1pt]
\end{tabular}
\vspace{1mm}
\caption{Symbols and notations used in the paper.}
\label{table:Symbol}
\end{table}

\subsection{Polarization measurement}


Given polarization images $I^{0}, I^{\pi/2}, I^{\pi/4}, I^{3\pi/4}$ obtained by different polarizer angles, the polarization information can be obtained through the following equation:
\begin{align}
\label{eq:pol}
&I = (I^{0} + I^{\pi/2} + I^{\pi/4} + I^{3\pi/4})/2, \\
\label{eq:pol2}
&\textbf{$\rho$} = \frac{\sqrt{((I^{0} - I^{\pi/2})^{2} + (I^{\pi/4} - I^{3\pi/4})^{2})}}{I}, \\
\label{eq:pol3}
&\textbf{$\phi$} = \frac{1}{2}\arctan\frac{I^{\pi/4}-I^{3\pi/4}}{I^{0}-I^{\pi/2}}.
\end{align}

\subsection{Preliminary}
\begin{figure}[t]
\centering
\begin{tabular}{@{}c@{}}
\includegraphics[width=1.0\linewidth]{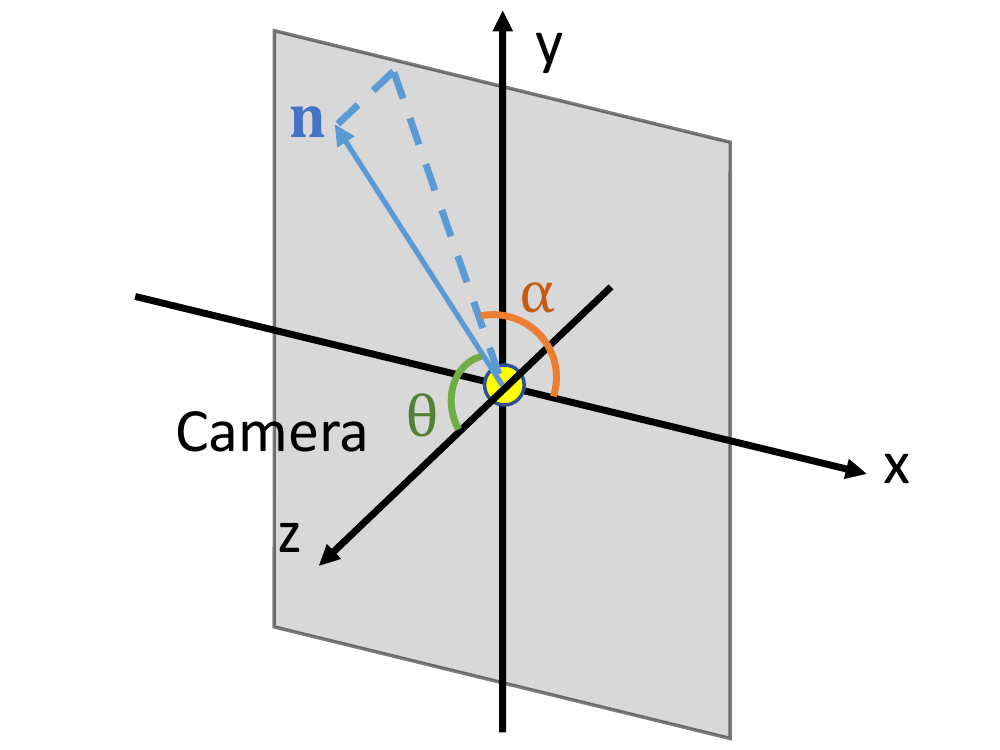}
\end{tabular}
\caption{Our coordinate system.}
\label{fig:framework}
\end{figure}

\textbf{Coordinate system.} We represent the surface normal and viewing direction in a global coordinate system, as shown in Fig.~\ref{fig:framework}. The x-axis is rightward. The y-axis is upward. The z-axis is pointing out of paper. The original point of this coordinate system coincides with the camera's Principle Point. The camera plane is perpendicular to the z-axis.

\textbf{Normal representation.} Surface normal can be represented by two angles $\theta$ and $\alpha$:
\begin{align}
    \label{eq:normal_2angle}
    \mathbf{n} = [\rm{sin}\theta \rm{cos}\alpha,\rm{sin}\theta \rm{sin}\alpha , \rm{cos}\theta]^\intercal,
\end{align}
where $\mathbf{n}$ is the surface normal, $\theta$ is the zenith angle, and $\alpha$ is the azimuth angle, as shown in Fig.~\ref{fig:framework}.

\subsection{Polarization under orthographic projection}
\label{subsec:orthographic}
More details about the relationship between the surface normal and polarization information are presented.

\subsubsection{Zenith angle}
\label{subsec:ortho_zenith}
The viewing angle $\theta_\mathbf{v}$ is the angle between viewing direction and surface normal. Under orthographic projection, the zenith angle $\theta$ equals to the viewing angle $\theta_\mathbf{v}$ according to Equation~\ref{eq:normal_2angle}:
\begin{align}
    {\rm cos}  \theta_\mathbf{v} &= \mathbf{n}  \cdot \mathbf{v} = {n}_x  {v}_x + {n}_y  {v}_y + {n}_z  {v}_z,  \label{eq:dop1} \\
    &=  {\rm cos}  \theta,  \ \ \rm{if}  \ \ \mathbf{v} = [0, 0, 1]  
\end{align}


The viewing angle $\theta_\mathbf{v}$ influences the degree of polarization $\rho$ directly. Specifically, given the refractive index $\eta$ of the object, the degree of polarization $\rho$ is decided by the viewing angle $\theta_\mathbf{v}$ with a function $\rho = g(\theta_\mathbf{v};\eta)$. The function $g$ is decided by many factors, such as the reflection type. For example, for specular reflection, we have: 
\begin{align}
    \rho= 
    \frac{2{\rm sin}^{2}\theta_\mathbf{v} {\rm cos}\theta_\mathbf{v} \sqrt{\eta^2 - {\rm sin}^{2}\theta_\mathbf{v}}}
    {\eta^2 - {\rm sin}^2\theta_\mathbf{v} - \eta^2 {\rm sin}^2 \theta_\mathbf{v} + 2{\rm sin}^4\theta_\mathbf{v}}. \label{eq:dop_ortho}
\end{align}
For diffuse reflection, we have:
\begin{align}
    \label{eq:rho_d}
    &\rho = 
    &\frac{(\eta - \frac{1}{\eta})^2{\rm sin}^2\theta_\mathbf{v}}
    {2+2\eta^2 - (\eta+\frac{1}{\eta})^2{\rm sin}^2\theta_\mathbf{v} + 4{\rm cos}\theta_\mathbf{v} \sqrt{\eta^2 - {\rm sin}^2\theta_\mathbf{v}}}.
\end{align}

\begin{figure*}[t]
\centering
\begin{tabular}{@{}c@{\hspace{4mm}}c@{}}
\includegraphics[width=0.4\linewidth]{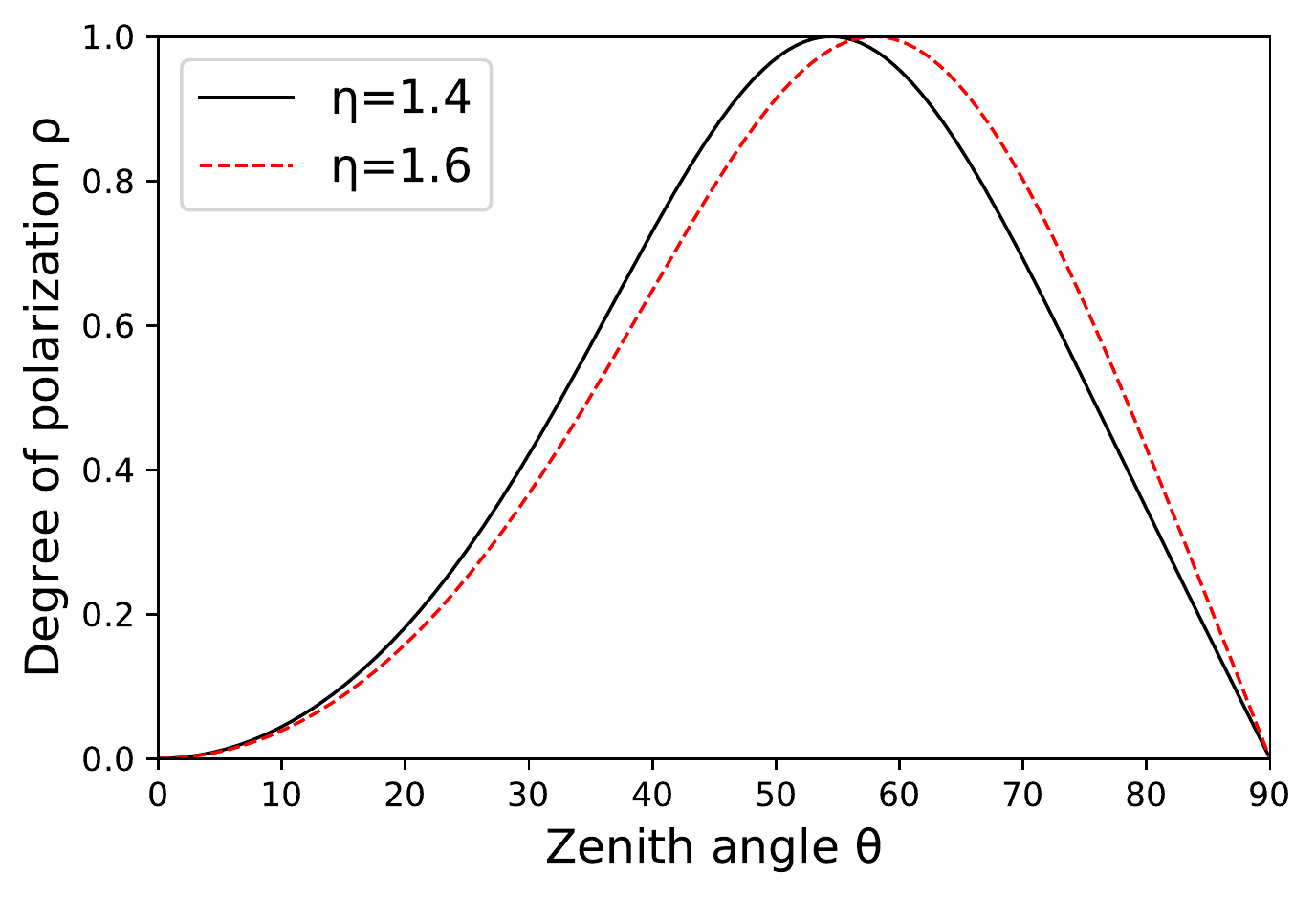}&\includegraphics[width=0.4\linewidth]{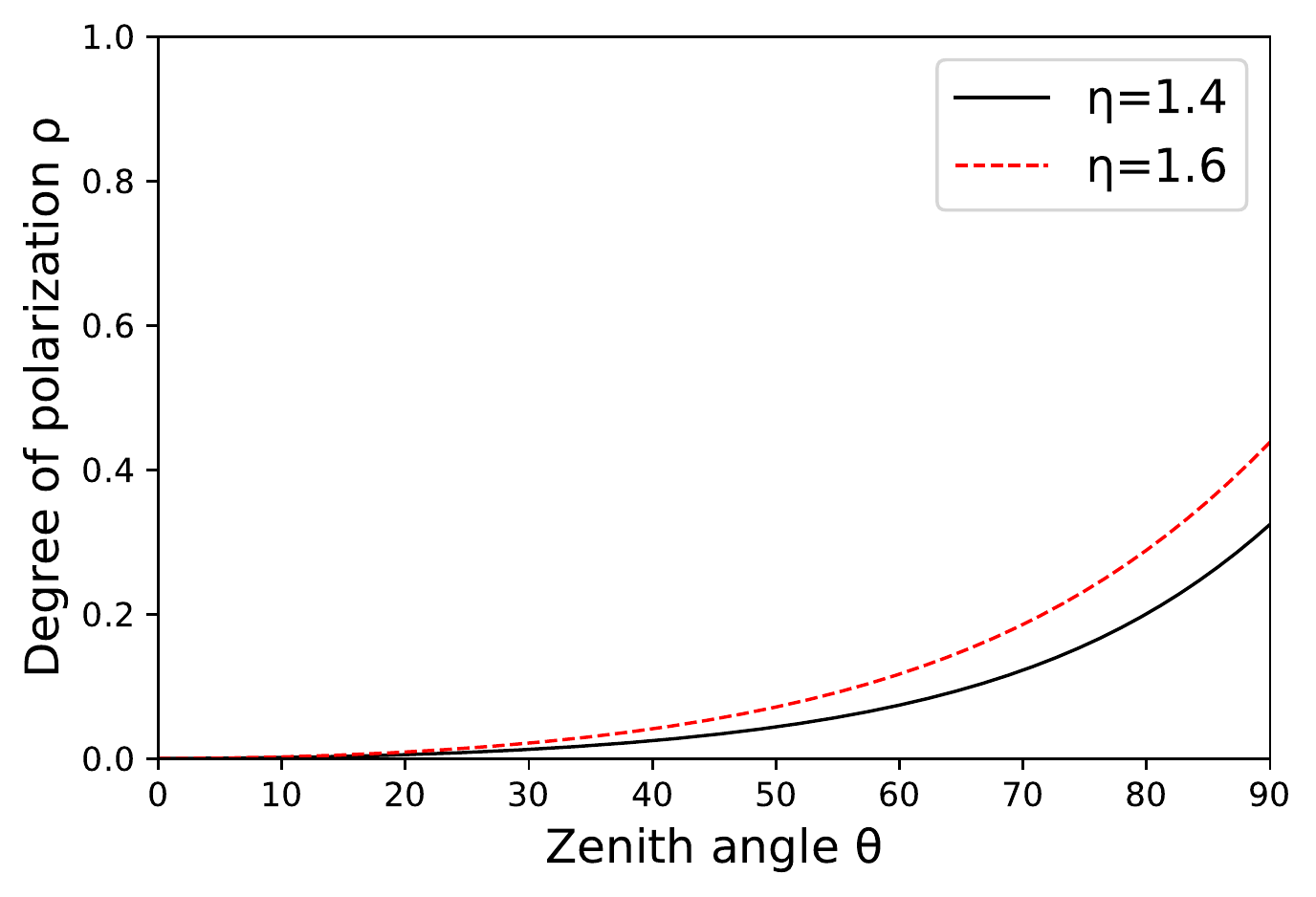}\\
(a) Specular reflection & (b) Diffuse reflection
\end{tabular}
\caption{Degree of polarization changes differently for (a) specular and (b) diffuse reflection. $\eta$: refractive index.}
\label{fig:rho_and_theta}
\end{figure*}

Equation~\ref{eq:dop_ortho} can be inverted to obtain an estimation of viewing angle from the degree of polarization $\rho$:
\begin{align}
  &  \cos{\theta} = \nonumber \cos{\theta_\mathbf{v}} = \nonumber \\ 
    &\resizebox{.45 \textwidth}{!}{
    \sqrt{\frac{\eta^{4} (1-\rho^{2})+2 \eta^{2}\left(2 \rho^{2}+\rho-1\right)+\rho^{2}+2 \rho-4 \eta^{3} \rho \sqrt{1-\rho^{2}}+1}{(\rho+1)^{2}\left(\eta^{4}+1\right)+2 \eta^{2}\left(3 \rho^{2}+2 \rho-1\right)}}.} \label{eq:zenith}
\end{align}
As shown in Fig.~\ref{fig:rho_and_theta}, we can estimate the zenith angle $\theta$ given the degree of polarization $\rho$ under a specific refractive index $\eta$ and the reflection type.

\subsubsection{Azimuth angle}
\label{subsec:ortho_azimuth}

The azimuth angle $\alpha$ is closely related to the polarization angle $\phi$. Specifically, there are four possible solutions for $\alpha$ based on the measured $\phi$ under the \textit{orthographic assumption} (i.e., $\mathbf{v} = [0, 0, 1] ^\intercal$):
\begin{equation}
\label{eq:aop}
    \alpha \in \{ \phi,  \phi + \pi, \phi + \pi / 2, \phi - \pi/2 \}, \ \ 0 \leq \alpha < 2\pi. 
\end{equation}
There are two ambiguities in the solution: $\pi$-ambiguity and $\pi/2$-ambiguity. The $\pi$-ambiguity is because $\phi$ is from 0 to $\pi$ and there is no difference between $\phi$ and $\phi + \pi$. The $\pi/2$-ambiguity is decided by the reflection type. If diffuse reflection dominates, $\alpha$ equals to $\phi$ or $\phi + \pi$; if specular reflection dominates, there is a $\pi/2$ shift compared with $\phi$.

\subsubsection{Solutions for surface normal}
As analyzed in Section~\ref{subsec:ortho_zenith} and Section~\ref{subsec:ortho_azimuth}, zenith angle $\theta$ and azimuth angle $\alpha$ can be estimated through the degree of polarization $\rho$ and angle of polarization $\phi$ respectively. At last, we can obtain possible solutions using Equation~\ref{eq:normal_2angle} directly.

\subsection{Polarization under perspective projection}


    
    
    
    

Most equations in Section~\ref{subsec:orthographic} do not hold under perspective projection since we cannot assume $\mathbf{v}=[0,0,1]^\intercal$ for all pixels. However, we can still derive other equations from utilizing the relationship between polarization and surface normal $\mathbf{n}$.

\subsubsection{Degree of polarization}
Zhu et al.~\cite{zhu2019depth} extend the linear formulation of Smith et al.~\cite{smith2019height} to the perspective case for the zenith angle. Note the degree of polarization $\rho$ is decided by the viewing angle $\theta_\mathbf{v}$ and refractive index $\eta$. Since $\theta_\mathbf{v}$ is angle between surface normal $\mathbf n$ and viewing direction $\mathbf v$:
\begin{align}
    {\rm cos}  \theta_\mathbf{v} &= \mathbf{n}  \cdot \mathbf{v} = {n}_x  {v}_x + {n}_y  {v}_y + {n}_z  {v}_z,  \label{eq:dop1} \\
    {\rm cos}\theta_\mathbf{v} &= \mathit{{v}_x}\rm{sin}\theta \rm{cos}\alpha   + \mathit{{v}_y}\rm{sin}\theta \rm{sin}\alpha + \mathit{{v}_z}\rm{cos}\theta. \label{eq:dop2}
\end{align}
Obviously, we do not have $\theta=\theta_{\mathbf{v}}$ under perspective projection ($\mathbf{v}$ is not $[0,0,1]^\intercal$) for all pixels. Hence, we cannot estimate the zenith angle $\theta$ through the degree of polarization $\rho$ individually like Equation~\ref{eq:zenith}. 

\subsubsection{Angle of polarization}
\begin{figure}[t]
\centering
\begin{tabular}{@{}c@{}}
\includegraphics[width=1.0\linewidth]{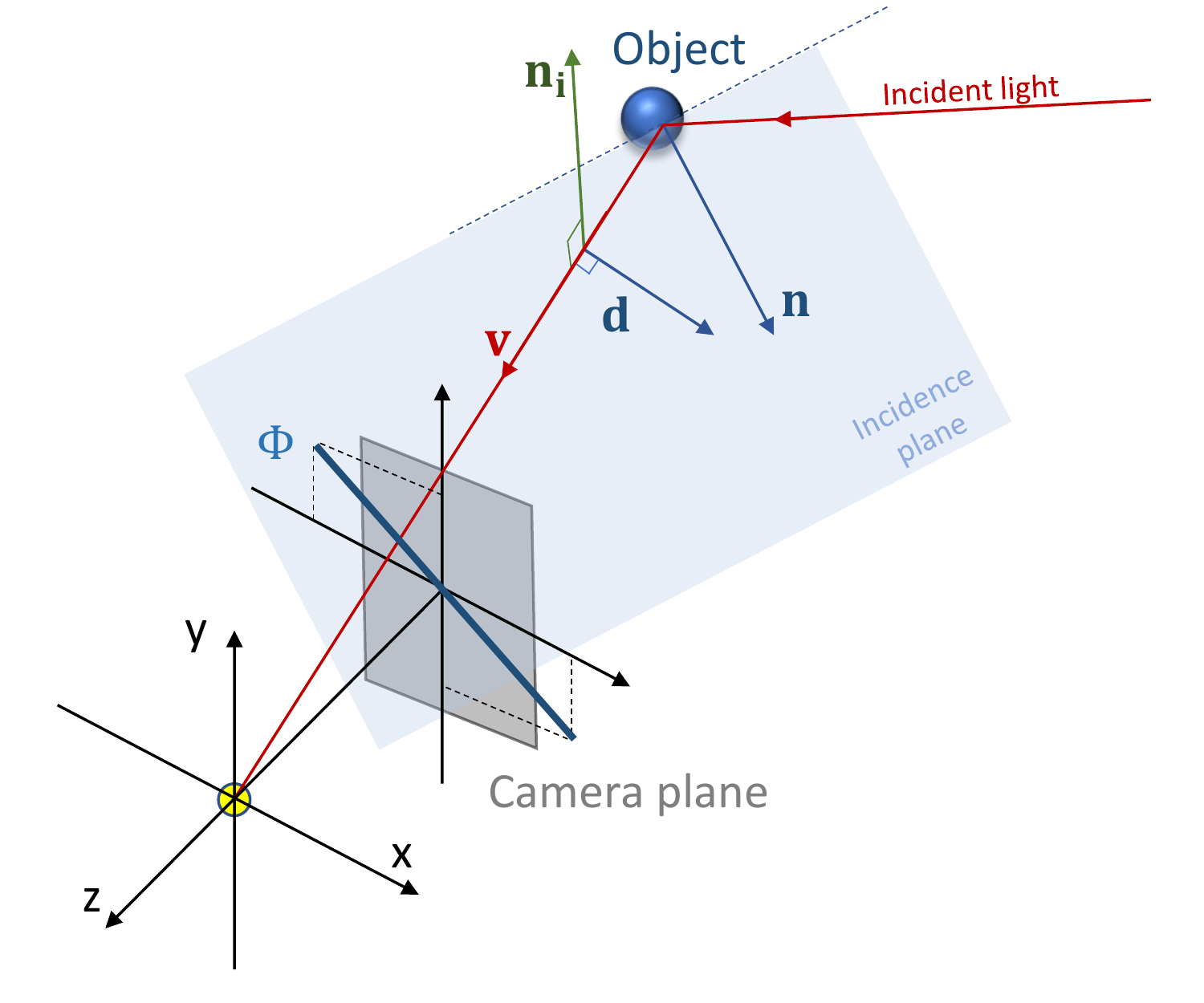}
\end{tabular}
\caption{Shape from polarization under perspective projection.}
\label{fig:perspective projection}
\end{figure}

We extend the azimuth angle formulation to the perspective case here. The incidence plane is the plane that contains the surface normal, incident light, and viewing direction. Hence, the normal of incidence plane $\mathbf{n_i}$ is perpendicular to surface normal $\mathbf{n}$ and viewing direction $\mathbf{v}$:
\begin{align}
    \mathbf{n_i} = \mathbf{n} \times \mathbf{v}.
\end{align}

In terms of their physical properties, the polarization direction $\mathbf{d}$ has no difference with its opposite direction $-\mathbf{d}$. For brevity, we only consider one direction. The polarization direction $\mathbf{d}$ is perpendicular to the propagation direction of light. In addition, $\mathbf{d}$ is always parallel or perpendicular to the incidence plane. Hence, we have:
\begin{align}
\label{eq:aop_}
    \mathbf{d} =\left\{
\begin{array}{lcl}
\mathbf{n_i} ,  \ \  &\mathbf{d} \bot \mathbf{P_i}     \\
\mathbf{n_i} \times \mathbf{v}, \ \  &\mathbf{d} \parallel  \mathbf{P_i},
\end{array} \right. 
\end{align} 
where $\mathbf{P_i}$ is the incidence plane.




When polarization direction $\mathbf{d}$ is projected on the camera plane, we have the following equation since this is an intersection between polarization direction $\mathbf{d}$ and camera plane:
\begin{align}
    \Phi &= (\mathbf{d} \times \mathbf{v}) \times \mathbf{n_c}=[\cos \phi, \sin \phi,0]^\intercal, \\
    \phi &= \arctan({\Phi}_y / {\Phi}_x),\label{eq:phi_Phi}
\end{align}
where $\mathbf{n_c}=[0,0,1]^\intercal$ is the surface normal of camera plane. At last, we can get the angle of polarization $\phi$ through Equation~\ref{eq:phi_Phi} directly.

To sum up, the $\Phi$ can be modeled as follows. When diffuse reflection dominates, we have:
\begin{align}
    \Phi =  \mathbf{d} \times \mathbf{n_c} =  \mathbf{n_i} \times \mathbf{n_c} =  \mathbf{ \mathbf{n} \times v} \times \mathbf{n_c}.
\label{eq: phi_diffuse}
\end{align}
When specular reflection dominates, we have:
\begin{align}
    \Phi =  \mathbf{d} \times \mathbf{n_c} =  \mathbf{n_i} \times \mathbf{v} \times \mathbf{n_c} =  \mathbf{ \mathbf{n} \times v} \times \mathbf{v} \times \mathbf{n_c}.
\label{eq: phi_specular}
\end{align}

We provide an example for Equation~\ref{eq: phi_diffuse} and Equation~\ref{eq: phi_specular}. From Equation~\ref{eq: phi_diffuse}, when $\mathbf{v}=[0,0,1]^\intercal$, we have:
\begin{align}
    \Phi_{[0,0,1]} &=  \mathbf{\mathbf{n} \times v} \times \mathbf{n_c} \nonumber \\
    &=[n_y v_z - n_z v_y, n_z v_x - n_x v_z, n_x v_y - n_y v_x]^\intercal \times \mathbf{n_c} \nonumber \\
    &=[n_z v_x - n_x v_z, -(n_y v_z - n_z v_y), 0]^\intercal  \nonumber \\
    &= [-n_x, -n_y, 0]^\intercal, \\
    \phi & \in \{\alpha, \alpha + \pi\}
\label{eq: phi_diffuse_001}
\end{align}
Similarly, from Equation~\ref{eq: phi_specular}, 
\begin{align}
    \Phi_{[0,0,1]} & =  \mathbf{ \mathbf{n} \times v} \times \mathbf{v} \times \mathbf{n_c} \nonumber \\
    & = [-n_x, -n_y, 0]^\intercal \times \mathbf{n_c} \nonumber \\
    &= [-n_y, n_x, 0]^\intercal, \\
    \phi & \in \{\alpha + \pi/2, \alpha - \pi/2\}
\label{eq: phi_diffuse_001}
\end{align}
Our result under $\mathbf{v}=[0,0,1]^\intercal$ is consistent with Equation~\ref{eq:aop}.

{\small
\bibliographystyle{ieee_fullname}
\bibliography{egbib}
}